%% file: neurips_2025.tex
\documentclass{article}

\usepackage[preprint]{neurips_2025}

\usepackage[utf8]{inputenc} %
\usepackage[T1]{fontenc}    %
\usepackage{hyperref}       %
\usepackage{url}            %
\usepackage{booktabs}       %
\usepackage{amsfonts}       %
\usepackage{nicefrac}       %
\usepackage{microtype}      %
\usepackage{xcolor}         %
\usepackage{amsmath}
\usepackage{amssymb}
\usepackage{mathtools}
\usepackage{amsthm}
\usepackage[ruled,linesnumbered]{algorithm2e}
\usepackage[capitalize,noabbrev]{cleveref}
\usepackage{bm}
\theoremstyle{plain}
\newtheorem{theorem}{Theorem}[section]

\theoremstyle{definition}

\theoremstyle{remark}

\title{Hierarchical Rectified Flow Matching with Mini-Batch Couplings}

\author{Yichi Zhang\textsuperscript{1},\quad Yici Yan\textsuperscript{1},\quad Alex Schwing\textsuperscript{1},\quad Zhizhen Zhao\textsuperscript{1} \\
\\
\textsuperscript{1}University of Illinois Urbana-Champaign\\
}

\begin{document}

\maketitle

\begin{abstract}
\input{00_abs}
\end{abstract}

\input{01_intro}
\input{02_prelim}

\input{03_approach}

\input{04_exp}

\input{05_rel}

\input{06_conc}

\clearpage
\bibliographystyle{abbrvnat}
\bibliography{neurips_2025}

\clearpage
\appendix
\input{10_appendix}

\input{11_appendix_res}

\input{12_appendix_imp}

\end{document}

%% file: 00_abs.tex
Flow matching has emerged as a compelling generative modeling approach that is widely used across domains. 
To generate data via a flow matching model, an ordinary differential equation (ODE) is numerically solved via forward integration of the modeled velocity field. 
To better capture the multi-modality that is inherent in typical velocity fields, hierarchical flow matching was recently introduced. It uses a hierarchy of ODEs that are numerically integrated when generating data. 
This hierarchy of ODEs captures the multi-modal velocity distribution just like vanilla flow matching is capable of modeling a multi-modal data distribution. 
While this hierarchy enables to model multi-modal velocity distributions, the complexity of the modeled distribution remains identical across levels of the hierarchy. 
In this paper, we study how to gradually adjust the complexity of the distributions across different levels of the hierarchy via mini-batch couplings. 
We show the benefits of mini-batch couplings in hierarchical rectified flow matching via compelling results on synthetic and imaging data. 
Code is available at \url{https://riccizz.github.io/HRF_coupling}. 

%% file: 01_intro.tex
\section{Introduction}
\label{sec:intro}

Flow matching~\citep{LipmanICLR2023,liu2023flow,albergo2023building} has gained significant attention across computer vision~\citep{esser2024scaling,liu2023instaflow}, robotics~\citep{zhang2024affordance}, computational biology~\citep{yim2023fast,jing2024alphafold}, and time series analysis~\citep{chen2024probabilistic,zhang2024trajectory}. This is largely due to its ability to generate high-quality data and due to its simple simulation-free learning of a data distribution. For this, it uses 1) an intermediate state, which is computed by (linearly) interpolating between a sample from a known source distribution and a randomly drawn data point, and 2) the velocity at this intermediate state. This velocity controls a neural ordinary differential equation (ODE), which governs the transformation of the samples from the source distribution to the target data distribution. Note, the distribution of velocities at an intermediate state is  multi-modal~\citep{zhang2025towards}.

The velocity field is obtained by matching velocities from the interpolated samples with a parametric deep net using a mean squared error (MSE) objective. It is known that the  MSE objective used in classic rectified flow does not permit to capture this multi-modal velocity distribution. Instead, classic rectified flow leads to a model that aims to capture the mean of the velocity distribution. Capturing the mean of the velocity distribution is sufficient for characterizing a multi-modal data distribution~\citep{liu2023flow}. However, it inevitably results in curved forward integration paths, making the sampling process inefficient. %
Recently, hierarchical rectified flow~\citep{zhang2025towards} was suggested as an approach to model the multi-modal velocity field via coupled ODEs. %

To model the multi-modal velocity field, hierarchical rectified flow essentially applies a rectified flow formulation in the velocity space by matching `acceleration.' It was also suggested to expand the idea further towards an arbitrary hierarchy level. 
While this enables to model multi-modal velocity distributions, the complexity of the modeled distribution remains identical across all levels of the hierarchy. Said differently, the velocity distribution that a hierarchical rectified flow models across  levels of its hierarchy is no easier than the original data distribution, potentially limiting  benefits. 

We hence wonder: \textit{can we gradually simplify the complexity of the ``ground-truth'' velocity distribution?} %
For simplicity, in this paper, we focus on two hierarchy levels. Interestingly, we find mini-batch couplings to provide a compelling way to control the ``ground-truth'' velocity distribution. 
Instead of computing intermediate states by interpolating between samples independently drawn from both the known source distribution and the dataset, we draw a mini-batch of samples from both the source distribution and the dataset, and subsequently couple them, e.g., via a procedure like optimal transport. %
Intuitively, considering as an extreme situation a mini-batch containing the entire dataset leads to a velocity distribution that is uni-modal. 

Empirically, we find that %
hierarchical flow matching with mini-batch coupling in the data space consistently improves the generation quality of vanilla hierarchical rectified flow matching and vanilla flow matching with optimal transport coupling. Jointly coupling mini-batch samples in data and velocity space leads to further benefits if the number of neural function evaluations (NFEs) is low. 

%% file: 02_prelim.tex
\section{Preliminaries}
\label{sec:prelim}

\noindent\textbf{Rectified Flow (RF).}
A rectified flow models an unknown target data distribution $\rho_1$ given a dataset ${\cal D} = \{x_1\}$, where we assume  data points $x_1 \sim \rho_1$. Given a known source distribution $\rho_0$ (e.g., standard Gaussian), at inference time, source samples $x_0 \sim \rho_0$ evolve from time $t = 0$ to time $t = 1$ following the ordinary differential equation (ODE)  
\begin{equation}
\label{eq:RF}
d z_t = v(z_t, t) dt, \, \text{with } z_0 \sim \rho_0, \quad t \in [0, 1].
\end{equation}  
Here, $v(z_t, t)$ is a velocity field that depends on time $t$ and the current intermediate state $z_t$. This ODE-based sampling enables 
 to capture multi-modal data distributions.

At training time,  flow matching learns the velocity field $v(z_t, t)$ by minimizing the $\ell_2$-loss between the predicted velocity $v(x_t, t)$ and a ground-truth velocity $v_{\text{gt}}(x_t, t)$.  %
To obtain the ground-truth velocity we first define an
intermediate state $x_t$ which, in a rectified flow  formulation, is obtained by linearly interpolating between a randomly drawn source sample $x_0$ and a randomly drawn data point $x_1$, i.e.,  
\begin{equation}
\label{eq:lin_int}
x_t = (1 - t)x_0 + t x_1, \quad \text{where} \, x_0 \sim \rho_0, \, x_1 \sim {\cal D}.
\end{equation}  
Interpreting the intermediate state $x_t$ as a location, we obtain the ground-truth velocity $v_{\text{gt}}(x_t, t) = \partial x_t/\partial t = x_1 - x_0$. Combined, 
the training objective is then:
\begin{equation}
\label{eq:lin_rect_flow}
\inf_{v} \mathbb{E}_{x_0 \sim \rho_0, x_1 \sim {\cal D}, t \sim U[0, 1]} \left[ \|x_1 - x_0 - v(x_t, t)\|_2^2 \right],
\end{equation}  
where the optimization is over all measurable velocity fields. In practice, $v(x_t, t)$ is parameterized by a neural network with trainable parameters $\theta$, i.e., $v(x_t, t) \approx v_\theta(x_t, t)$, and the optimization minimizes over $\theta$.

However, for a given $t$ and $x_t$, different pairs $(x_0, x_1)$ will yield different ground-truth velocities. The ground-truth velocity distribution at a given time $t$ and intermediate state $x_t$ is hence multi-modal. However, the $\ell_2$-loss averages these velocities, resulting in the `optimal' velocity field:
$
v^*(x_t, t) = \mathbb{E}_{\{(x_0, x_1, t): (1 - t)x_0 + t x_1 = x_t\}}[v_\mathrm{gt}(x_t, t)].
$
According to Theorem 3.3 by~\citet{liu2023flow}, using $v^*$ in \cref{eq:RF} ensures that the stochastic process has marginal distributions consistent with the linear interpolation in \cref{eq:lin_int}.

To capture 
multi-modal velocity distributions, %
hierarchical rectified flow~\citep{zhang2025towards} was introduced. It explicitly models the multi-modal velocity distributions at each time $t$ and intermediate state $x_t$, enabling a more expressive generative framework.

\noindent\textbf{Hierarchical Rectified Flow (HRF).}
To model the ``ground-truth'' velocity distribution more accurately, hierarchical rectified flow extends the classic rectified flow framework by focusing on velocities rather than locations. This approach effectively involves learning acceleration. In a classic rectified flow, the time-dependent location $x_t$ is computed from pairs $(x_0, x_1)$, and the ground-truth velocity $v_{\text{gt}}(x_t, t) = \partial x_t / \partial t$ is used to train a velocity model $v_\theta(x_t, t)$. 

In hierarchical rectified flow, a source velocity sample $v_0 \sim \pi_0$ is drawn from a known velocity distribution $\pi_0$, while a target velocity sample $v_1(x_t, t) \sim \pi_1(v; x_t, t)$ is defined at each time $t$ and location $x_t = (1 - t)x_0 + t x_1$. For rectified flow, $v_1(x_t, t)$ is computed via $x_1 - x_0$, and these samples follow the ground-truth velocity distribution $\pi_1(v; x_t, t)$. 

To learn acceleration, a new time axis $\tau \in [0, 1]$ is introduced, and a time-dependent velocity $v_\tau(x_t, t) = (1 - \tau)v_0 + \tau v_1(x_t, t)$ is constructed. The ground-truth acceleration is then obtained as $a(x_t, t, v_\tau, \tau) = \partial v_\tau / \partial \tau = v_1(x_t, t) - v_0 = x_1 - x_0 - v_0$. For a fixed $(x_t, t)$, this leads to the ODE:
\begin{equation}
\label{eq:udiffeq}
du_\tau(x_t, t) = a(x_t, t, u_\tau, \tau) d\tau, \quad \text{with } u_0 \sim \pi_0.
\end{equation}
Here, $a(x_t, t, u_\tau, \tau)$ is the expected acceleration vector field: $a(x_t, t, u_\tau, \tau) = \mathbb{E}_{\pi_0, \pi_1(v; x_t, t)}[V_1 - V_0 | V_\tau = u]$. The acceleration vector field is learned by minimizing the following objective:
\begin{equation}
\label{eq:opt}
\inf_a \mathbb{E}_{x_0 \sim \rho_0, x_1 \sim \mathcal{D}, t \sim U[0, 1], v_0 \sim \pi_0, \tau \sim U[0, 1]} \left[\|(x_1 - x_0 - v_0) - a(x_t, t, v_\tau, \tau)\|_2^2\right].
\end{equation}
In practice, the acceleration is parameterized via a deep net $a_\theta(x_t, t, v_\tau, \tau)$, and the model is trained by minimizing this objective over the parameters $\theta$.

During sampling, coupled ODEs are used:
\begin{equation}
\label{eq:vflow}
\begin{cases}
   du_\tau(z_t, t) = a(z_t, t, u_\tau, \tau) d\tau, & u_0 \sim \pi_0, \ \tau \in [0, 1],  \\
   dz_t = u_1(z_t, t) dt, & z_0 \sim \rho_0, \ t \in [0, 1].
\end{cases}
\end{equation}
These ODEs map $z_0 \sim \rho_0$ to $z_1 \sim \rho_1$. Sampling involves drawing $v_0 \sim \pi_0$ and $x_0 \sim \rho_0$, integrating forward to obtain $v_1(x_0, 0)$, and then performing location updates iteratively until reaching $x_1$. This procedure can be implemented using the vanilla Euler method and the trained $a_\theta$. 

Considering the training objective for acceleration matching (\cref{eq:opt}) and the coupled ODEs for sampling (\cref{eq:vflow}), both can be naturally extended to any depth. In this paper, we  focus solely on depth-two HRF (HRF2) models.

\noindent\textbf{Minibatch Optimal Transport.}
Optimal Transport (OT) seeks to find an optimal coupling of two distributions that minimizes an expected transport cost~\citep{villani2009optimal}. Suppose $\alpha$ and $\beta$ are two distributions in $\mathbb{R}^d$, and $c:\mathbb{R}^d \times \mathbb{R}^d \rightarrow \mathbb{R}$ is some distance. Then OT aims to find the solution of the following optimization problem:
\begin{equation}\label{eq:ot}
    \inf_{\gamma \in \Gamma}\int_{\mathbf{R}^d\times \mathbf{R}^d}c^2(x,y)d\gamma(x,y),
\end{equation}
where $\Gamma$ is the set of all joint distributions with marginals $\alpha$ and $\beta$. When $\alpha$ and $\beta$ are both empirical distributions,  OT  reduces to linear programming, which is computationally expensive when the data size is large~\citep{peyre2019computational}. While OT is computationally expensive for large datasets, mini-batch OT~\citep{fatras2020learning, fatras2021minibatch} was introduced as an alternative: a small batch of the data is used to calculate the coupling, obtaining an unbiased estimator of the underlying truth~\citep{fatras2020learning}. Although mini-batch OT incurs an error compared to the exact OT, it has found use in practice~\citep{DeshpandeCVPR2018,IDeshpandeCVPR2019,pooladian2023multisample,tong2023improving,ChengICCV2025}. %
\citet{tong2023improving, pooladian2023multisample} showed that  training and inference are more efficient with mini-batch OT.

%% file: 03_approach.tex
\section{Approach}
\label{sec:app}
In \cref{sec:app:simp_v}, we use a simple 1D example to illustrate how the mini-batch couplings in data space and velocity space affect the velocity distribution and the generation of velocity samples. This motivates the development of HRF with mini-batch coupling. %
In \cref{sec:app:data_coup}, we introduce the training of HRF2 with mini-batch coupled data points. %
In \cref{sec:app:v_coup}, we explain how mini-batch coupling for velocity is achieved by leveraging a pre-trained model. %
In \cref{sec:app:data_v_coup}, we introduce a two-stage approach that combines mini-batch data coupling and velocity coupling. %

\begin{figure*}[t]
    \centering
    \setlength{\tabcolsep}{0pt}
    {\small
    \begin{tabular}{ccc}
    \includegraphics[width=0.33\linewidth]{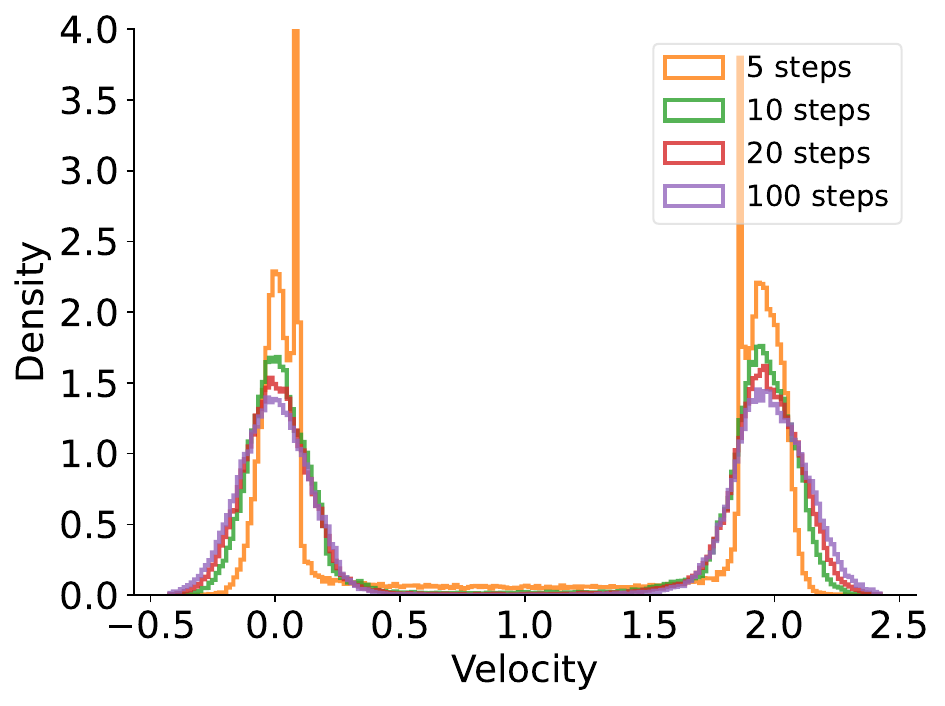}&
    \includegraphics[width=0.33\linewidth]{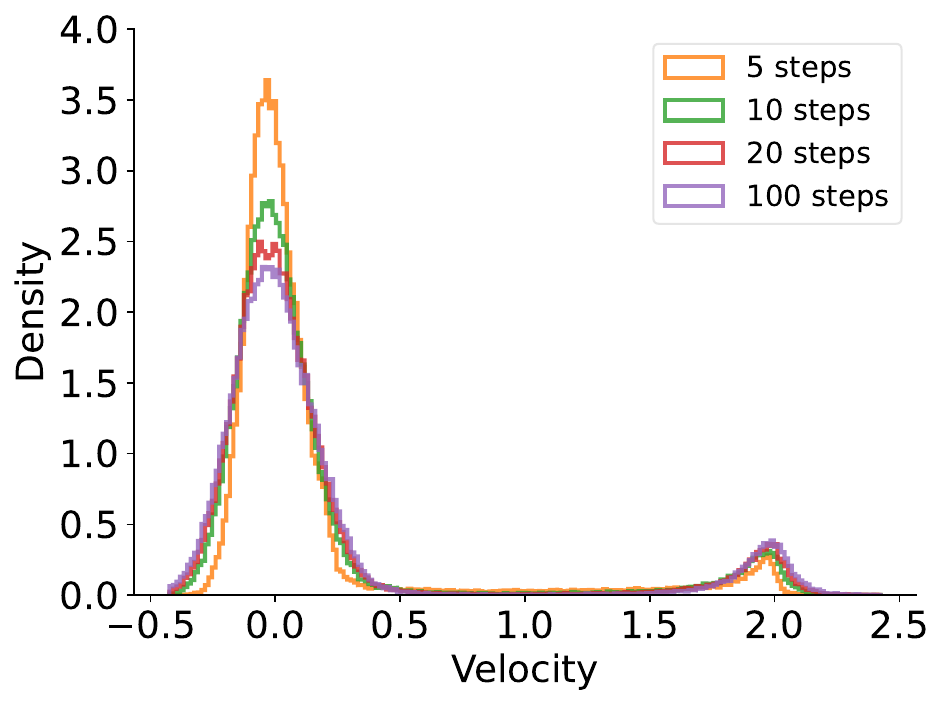}&
    \includegraphics[width=0.33\linewidth]{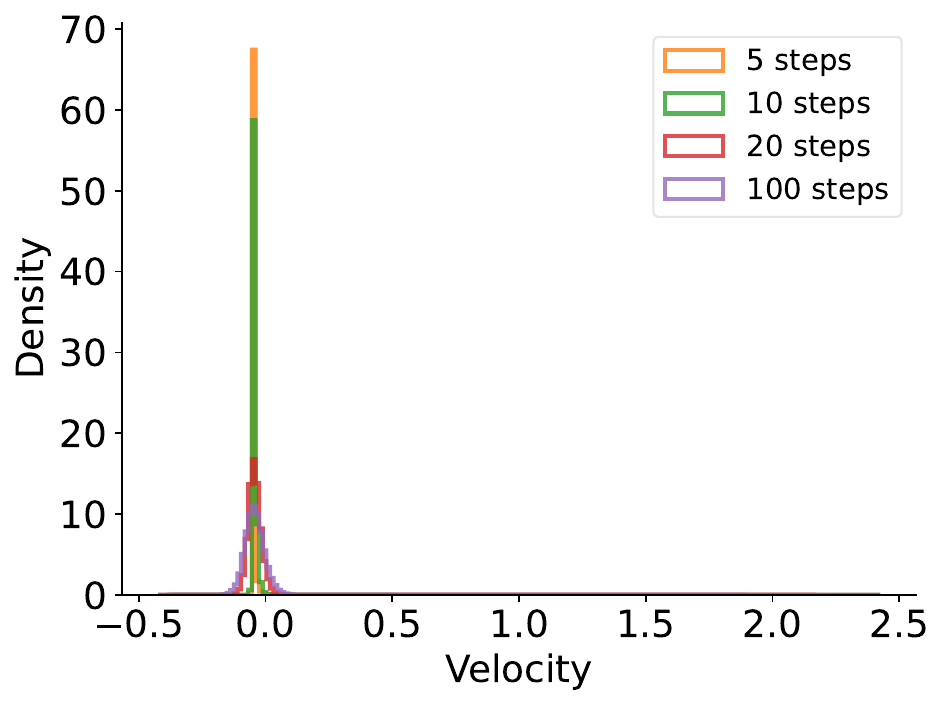}\\
    \vspace{5pt}
    (a) original & (b) data coupling w/ bs 5 & (c) data coupling w/ bs 100 \\
    \includegraphics[width=0.33\linewidth]{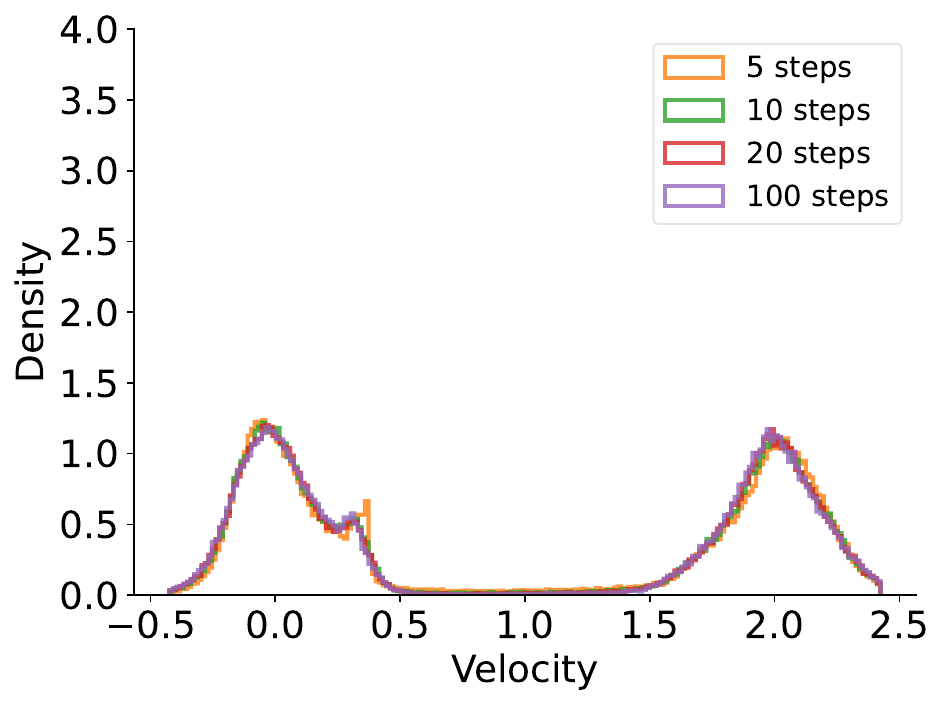}&
    \includegraphics[width=0.33\linewidth]{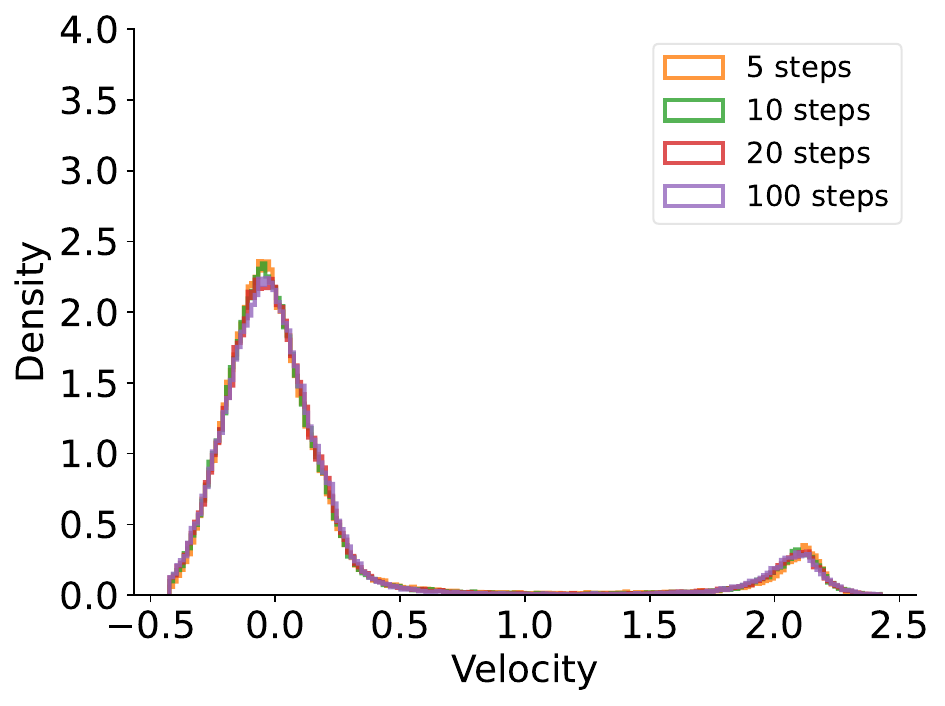}&
    \includegraphics[width=0.33\linewidth]{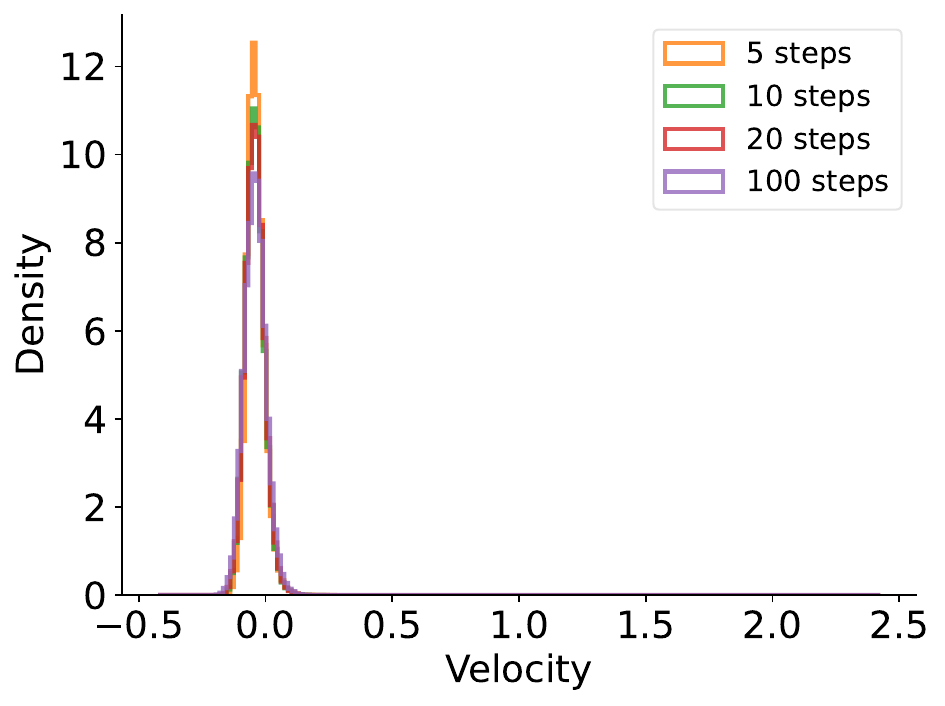}\\
    (d) velocity coupling & (e) data \& velocity coupling w/ bs 5 & (f) data \& velocity coupling w/ bs 100 
    \end{tabular}}
    \caption{The generated velocity distributions at $(x_t, t) = (-1, 0)$ for the dataset 1D $\mathcal{N} \to 2\mathcal{N}$ using HRF2, (a) without couplings, (b)-(c) with data coupling (batch sizes: 5 and 100), (d) with velocity coupling (batch size: 100), (e)-(f) with velocity coupling (batch size: 100) and data coupling (batch sizes: 5 and 100). Data coupling simplifies the velocity distribution (cf. (a)-(c)), while velocity coupling reduces the number of sampling steps. }
    \label{fig:v_dist_1to2}
\end{figure*} 

\subsection{Velocity Distribution }
\label{sec:app:simp_v}
HRF was designed with independently sampled $x_0$ and $x_1$. However, in this paper, we first show that the underlying theory can be generalized to an arbitrary joint distribution over $x_0$ and $x_1$, i.e., $\gamma(x_0, x_1)$, which has the correct marginal distributions, i.e.,  
\begin{equation}
    \int \gamma(x_0, x_1) d x_1 = \rho_0 (x_0) \, \text{and}\,\int \gamma(x_0, x_1) dx_0= \rho_1 (x_1).
\end{equation}
The following theorem characterizes the distribution of the velocity at a specific space-time location $(x_t,t)$ if an arbitrary joint distribution $\gamma$ is used instead of a product of two independent distributions.
\begin{theorem}
\label{the:pv}
The velocity distribution $\pi_1(v; x_t, t)$ at the space-time location $(x_t, t)$ induced by the linear interpolation in~\cref{eq:lin_int} for $(x_0, x_1) \sim \gamma(x_0, x_1)$ is
\begin{align} 
\label{eq:vgxt}
 \pi_1(v; x_t, t) = \frac{\gamma(x_t - tv,x_t + (1-t)v)}{\rho_t (x_t)},
\end{align}
where 
\begin{equation}
\label{eq:rhot}
\rho_t(x_t) = \int \gamma\left(x_0, \frac{x_t - (1-t)x_0}{t}\right) d x_0 = \int \gamma\left(\frac{x_t - tx_1}{1-t}, x_1 \right) d x_1,
\end{equation}
and $\rho_t(x_t) \neq 0$. 
The distribution $ \pi_1(v; x_t, t)$ is undefined if $\rho_t(x_t) =0$. 
\end{theorem}
The proof of \cref{the:pv} is deferred to \cref{sec:proofpvgivenxt}. 

For vanilla HRF2, the source and target distributions are independent, meaning $ \gamma(x_0, x_1) = \rho_0(x_0) \rho_1(x_1) $. Consequently, as derived by \citet{zhang2025towards} for a special case of \cref{the:pv}, at time $ t = 0 $, the velocity distribution becomes $ \pi_1(v; x_t, t) = \rho_1(x_t + v) $, making it a shifted version of the data distribution. Hence, learning this distribution is as challenging as directly modeling the data distribution. %

To progressively simplify the velocity distribution across the levels of the HRF hierarchy, we introduce data coupling and velocity coupling. %
This makes the distributions easier to learn and improves overall model performance. Notably, coupling at one hierarchy level simplifies the distributions at all lower levels, thereby facilitating the matching process at the current level. Importantly, the complexity of the learned distribution can be controlled by adjusting the batch size in the mini-batch coupling process. 

To illustrate the aforementioned distribution simplification we provide an example with 1D data. In this example, the source distribution is a standard Gaussian, while the target distribution is a mixture of two Gaussians with means located at $-1$ and $1$. As shown in \cref{fig:v_dist_1to2}(a-c), after applying data coupling (depth 1), the velocity distribution (depth 2) collapses into a single-mode Gaussian as the coupling batch size (bs) increases, effectively simplifying the velocity layer's distribution. The number given in the legend refers to the number of used velocity ODE integration steps.

From \cref{fig:v_dist_1to2}(d), we observe that  velocity coupling on its own does not simplify the velocity distribution. %
Instead, it simplifies the distribution at the next level (acceleration) and straightens the  paths for velocity samples, reducing the number of integration steps needed to model the velocity distribution. %
\cref{fig:v_dist_1to2}(e,f) demonstrates that data coupling and velocity coupling are not mutually exclusive. They can be applied simultaneously to complement each other. 

Finally,  comparing the columns in \cref{fig:v_dist_1to2} shows that a larger batch size for mini-batch coupling results in a uni-modal distribution. Therefore, in practice, it is important to choose an appropriate batch size to control the multi-modality at each level. Next, we detail how data coupling and velocity coupling can be achieved. 

\subsection{HRF2 with Data Coupling}
\label{sec:app:data_coup}
To simplify the velocity distribution by reducing its multi-modality, it is crucial to first understand its root cause. During training, if the source data $x_0$ and target data $x_1$ are sampled independently, the multi-modality inherent in the data is preserved in the velocity distribution at $t = 0$. We discussed this in the sequel of \cref{the:pv}. Therefore, breaking this independence is key to simplifying the velocity distribution. 
We find that data couplings that restrict flexibility, e.g., mini-batch OT, provide an opportunity to do this. Intuitively, using mini-batch OT results in a coupling of source and target data that isn't entirely arbitrary, which inherently simplifies the velocity distribution. 

\begin{figure}
\centering
\begin{minipage}[t]{0.48\textwidth}
\begin{algorithm}[H]
\SetKwComment{Comment}{//}{}
\SetKwInOut{input}{Input}
\SetKwInOut{output}{Output}
\caption{HRF2 with Data Coupling}\label{alg:data_coup}
\input{The source distributions $\rho_0$ and $\pi_0$, the dataset $\mathcal{D}$, and the batch size $B$. \\}
\While{stopping conditions not satisfied}{
 Sample $\{x_0^{(i)} \}_{i=1}^B  \sim \rho_0$, $\{x^{(i)}_1\}_{j=1}^B \sim {\mathcal{D}}$, and $\{v^{(i)}_0\}_{k = 1}^B \sim \pi_0$\;
 Sample $\{t^{(i)}\}_{i=1}^B \!\sim \!U[0,1]$ and $ \{ \tau^{(i)}\}_{i\! = \!1}^B \!\sim \!U[0,1]$\;
 Use optimal transport to construct a set of coupled source and target pairs $\{(x^{(i)}_0, x_1^{(\sigma(i))})\}_{i = 1}^B$\;
 Compute loss following Eq.~\eqref{eq:x_coup_obj}\;
 Perform gradient update on $\theta$\;
}
\output{$\theta$}
\end{algorithm}

\end{minipage}\hfill
\begin{minipage}[t]{0.48\textwidth}
\begin{algorithm}[H]
\SetKwComment{Comment}{//}{}
\SetKwInOut{input}{Input}
\SetKwInOut{output}{Output}
\caption{HRF2 with Velocity Coupling}\label{alg:hrf_v_coup}
\input{The source distributions $\rho_0$ and $\pi_0$, and the dataset $\mathcal{D}$ } 
\While{stopping conditions not satisfied}{\vspace{5pt}
 Sample $x_0 \sim \rho_0, x_1 \sim {\mathcal{D}}$, and $v_0 \sim \pi_0$\; \vspace{17pt}
 Sample $t\sim U[0,1]$ and $\tau \sim U[0,1]$\; \vspace{13pt}
 Compute coupled $v_0$ and $v_1(x_t, t)$ using \cref{alg:v_coup_ot}\; \vspace{12pt}
 Compute loss using Eq.~\eqref{eq:v_coup_obj}\;
 Perform gradient update on $\theta$\;
}
\output{$\theta$}
\end{algorithm}
\end{minipage}
\vspace{-3mm}
\end{figure}

Following \citet{tong2023improving,pooladian2023multisample}, we apply mini-batch OT on the data used for HRF2 training. Let $\{x_0^{(i)}\}_{i = 1}^B \sim \rho_0$ and $\{x_1^{(i)}\}_{i = 1}^B \sim \mathcal{D}$. The OT problem in~\cref{eq:ot} can be solved exactly and efficiently on a small batch size using standard solvers, e.g., \texttt{POT}~\citep{flamary2021pot}. The resulting coupling from the algorithm gives us a permutation matrix that pairs $x_0^{(i)}$ with $x_1^{(\sigma(i))}$ for $i \in\{ 1, \dots, B\}$. 
Instead of sampling $x_0$ and $x_1$ independently from $\rho_0$ and the dataset $\mathcal{D}$ during training, we jointly sample $(x_0, x_1)$ from the  joint distribution $\gamma(x_0, x_1)$ characterized by the mini-batch OT result. Using these samples, the training objective  reads as follows: 
\begin{equation}
\label{eq:x_coup_obj}
\min_\theta \mathbb{E}_{(x_0, x_1) \sim \gamma, t \sim U[0, 1], v_0 \sim \pi_0, \tau \sim U[0, 1]} \left[\|(x_1 - x_0 - v_0) - a_\theta(x_t, t, v_\tau, \tau)\|_2^2\right].
\end{equation}
The optimization procedure is detailed in \cref{alg:data_coup}.

\subsection{HRF2 with Velocity Coupling}
\label{sec:app:v_coup}
Similar to data coupling, velocity coupling also aims to eliminate the independence between $v_0$ and $v_1(x_t, t)$.  %
With mini-batch coupled velocity samples that are drawn from an underlying joint distribution $\kappa_{x_t, t}(v_0, v_1(x_t, t))$, the corresponding objective function is defined as follows:
\begin{equation}
\label{eq:v_coup_obj}
\min_\theta \mathbb{E}_{x_0 \sim \rho_0, x_1 \sim \mathcal{D}, (v_0, v_1) \sim \kappa_{x_t, t}, t \sim U[0, 1], \tau \sim U[0, 1]} \left[\|(v_1(x_t, t) - v_0) - a_\theta (x_t, t, v_\tau, \tau)\|_2^2\right].
\end{equation}
\cref{alg:hrf_v_coup} summarizes the optimization procedure, 
for which we study the following coupling. %
\begin{algorithm}[t]
\SetKwComment{Comment}{//}{}
\SetKwInOut{input}{Input}
\SetKwInOut{output}{Output}
\caption{Velocity Coupling via Mini-Batch OT}\label{alg:v_coup_ot}
\input{Location $(x_t, t)$, source distribution $\pi_0$, batch size $B$, and a pre-trained HRF2 model $a_\theta$.}
Sample $\{v^{(i)}_0 \}_{i=1}^B \sim \pi_0$ \;
Generate $v_1^{(i)}(x_t, t)$ from $v^{(i)}_0$ via numerically solving Eq.~\eqref{eq:udiffeq} with a pre-trained $a_\theta$ for $i\!=\! 1,\! \dots\!, B$\;
Use OT to couple source and target points\;
\output{Coupled samples $\{ (v^{(i)}_0, v_1^{(\sigma(i))}(x_t, t) )\}_{i\!=\!1}^B $.}
\end{algorithm}

\noindent\textbf{Velocity Coupling via Batch OT.}
Different from the data coupling, where the target data samples are readily available, obtaining velocity samples for a fixed $(x_t, t)$ requires simulation. Note that the target velocity samples $v_1(x_t, t) \sim \pi_1(v; x_t, t)$. To correctly couple $v_0$ and $v_1(x_t, t)$, it is essential to fix the space-time location $(x_t, t)$, ensuring that the samples $v_1(x_t, t)$ are drawn from the same velocity distribution. Thus, the first step of velocity coupling is to obtain a batch of $v_1(x_t, t)$ at a fixed $(x_t, t)$. We achieve this by using a pre-trained HRF2 model $a_\theta(x_t, t, v_\tau, \tau)$, which transport $v_0$ to $v_1(x_t, t)$ according to~\cref{eq:udiffeq}. %
We then use the 2-Wasserstein optimal transport to sample coupled pairs $(v_0, v_1(x_t, t))$. This is detailed in \cref{alg:v_coup_ot}. %

\subsection{HRF2 with Hierarchical Data \& Velocity Couplings}
\label{sec:app:data_v_coup}
As shown before, data and velocity coupling complement each other. 
To use both couplings we need $(x_0, x_1) \sim \gamma$ and $(v_0, v_1(x_t, t)) \sim \kappa_{x_t, t}$. To achieve this, we apply a two-stage training. First, we use data coupling to train the model $a_\theta$ according to \cref{alg:data_coup}. In the second stage, using this pre-trained model, we generate paired samples $(v_0, v_1(x_t, t))$ according to \cref{alg:v_coup_ot}. We then train our network using the following objective:  
\begin{equation}
\label{eq:dv_coup_obj}
\min_\theta \mathbb{E}_{(x_0, x_1)\sim \gamma, (v_0, v_1) \sim \kappa_{x_t, t}, t \sim U[0, 1], \tau \sim U[0, 1]} \left[\|(v_1(x_t, t) - v_0) - a_\theta (x_t, t, v_\tau, \tau)\|_2^2\right].
\end{equation}
Importantly, note that the coupling of the velocities depends on the coupling of the data.

\subsection{Marginal Preserving Property}
The consistency of the velocity distribution with mini-batch velocity coupling  directly follows prior works that use mini-batch coupling and reflow for data generation~\citep{tong2023improving,pooladian2023multisample}.

In addition, we can prove that the generation process according to \cref{eq:vflow} with trained $a_\theta$ using mini-batch data coupling preserves the target data distribution and leads to correct marginals for all times $t \in [0, 1]$.
\begin{theorem}
\label{thm:marginal}
The time-differentiable stochastic process $\bm{Z} = \{Z_t: t \in [0, 1] \}$ generated by \cref{eq:vflow} has the same marginal distribution as the time-differentiable stochastic process $\bm{X} = \{X_t: t \in [0, 1] \}$ generated by the linear interpolation in \cref{eq:lin_int} with the joint distribution $\gamma$ induced by mini-batch coupling. 
\end{theorem}
The proof of \cref{thm:marginal} is deferred to \cref{sec:proof_thm_marginal}.

%% file: 04_exp.tex
\section{Experiments}
\label{sec:exp}
In this section, we explore how data coupling and velocity coupling influence the velocity distribution and assess whether simplifying it leads to performance improvements. For all experiments, we report the total neural function evaluations (NFEs), which represents the product of the number of integration steps across all HRF levels.

\subsection{Synthetic Data}
\label{sec:exp:syn}
We conduct experiments on four synthetic datasets used by \citet{zhang2025towards} to ensure a fair comparison with HRF2. %
These datasets include two 1D cases ($\mathcal{N} \to 2\mathcal{N}$, $\mathcal{N} \to 5\mathcal{N}$), and two 2D cases ($\mathcal{N} \to 6\mathcal{N}$ and $8\mathcal{N} \to$ moon-shaped data). We use the Wasserstein and sliced 2-Wasserstein distances to evaluate 1D and 2D experiments, respectively. A complete description of the model architecture, parameter settings, and training details is provided in \cref{app:imp:syn}. 

\begin{figure*}[t!]
\vspace{3mm}
    \centering
    \setlength{\tabcolsep}{0pt}
    {\small
    \begin{tabular}{cccc}
    \includegraphics[width=0.25\linewidth]{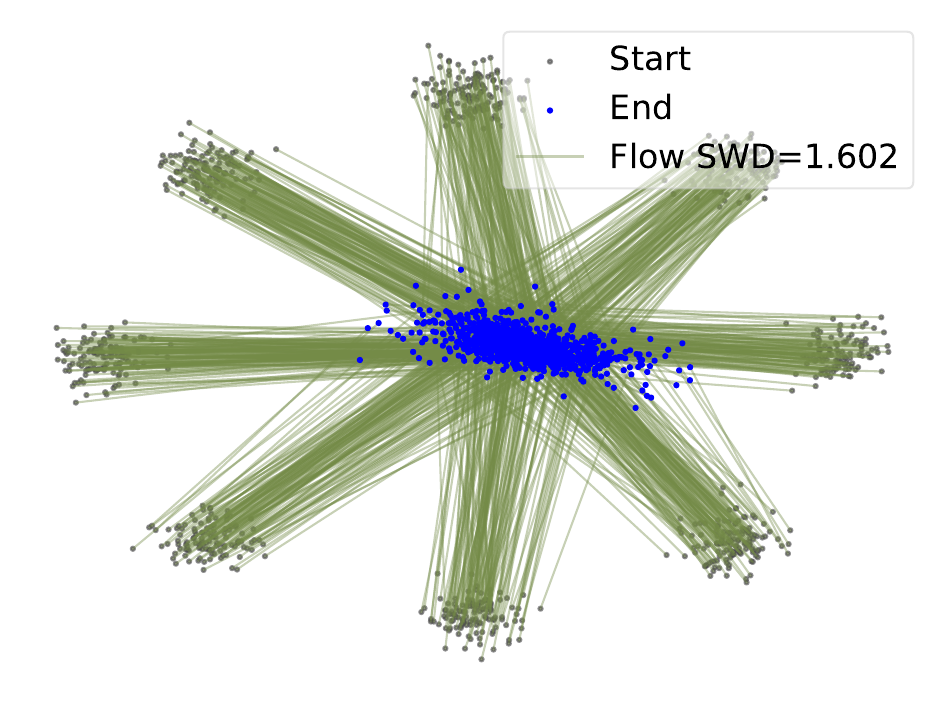}&
    \includegraphics[width=0.25\linewidth]{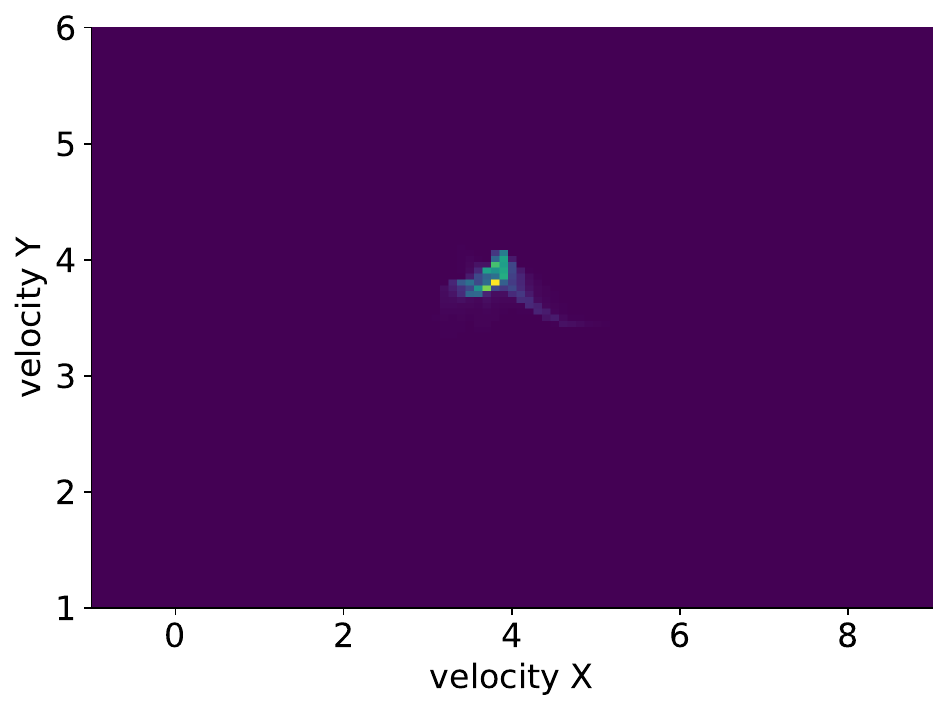}&
    \includegraphics[width=0.25\linewidth]{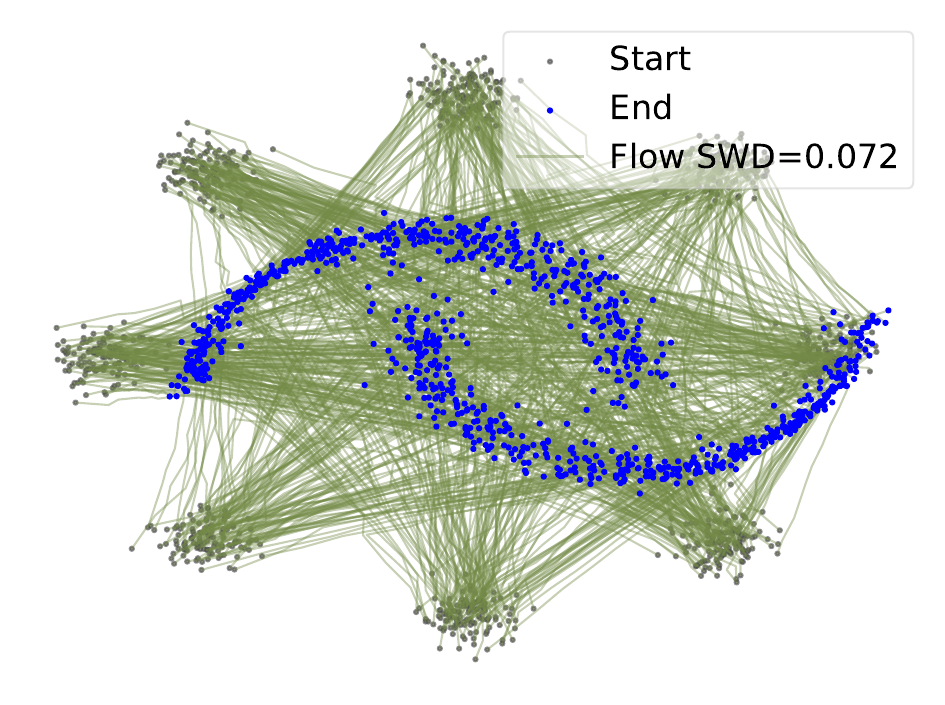}&
    \includegraphics[width=0.25\linewidth]{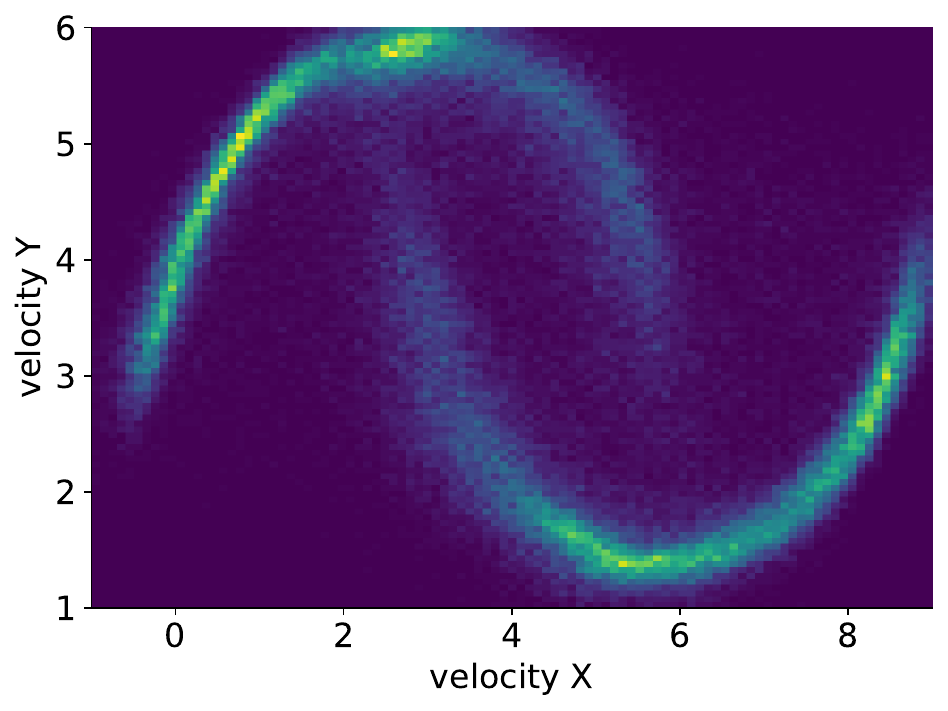}\\
    \includegraphics[width=0.25\linewidth]{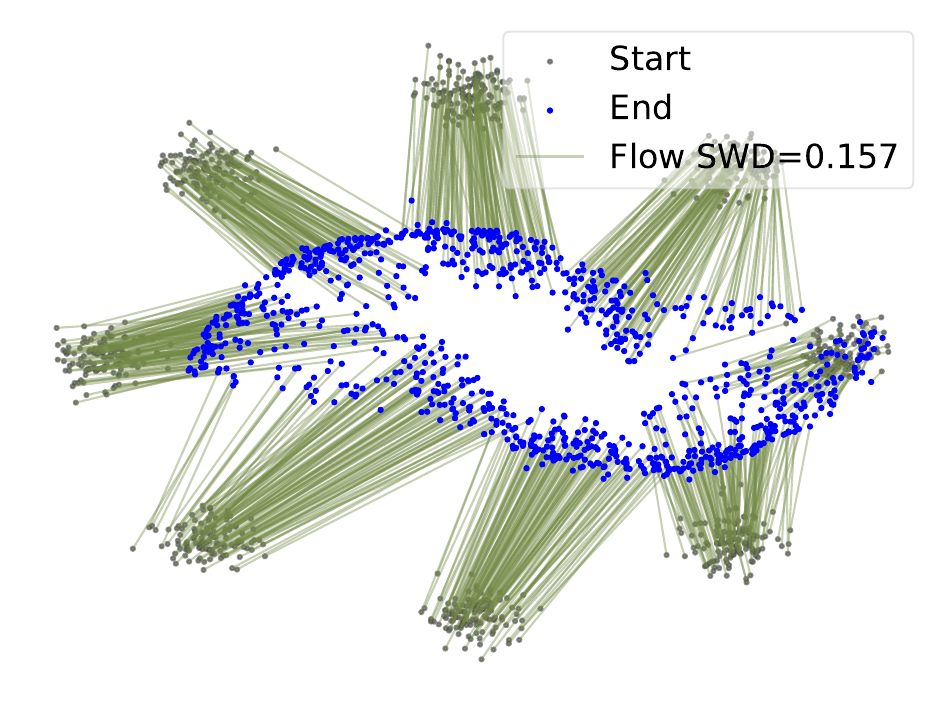}&
    \includegraphics[width=0.25\linewidth]{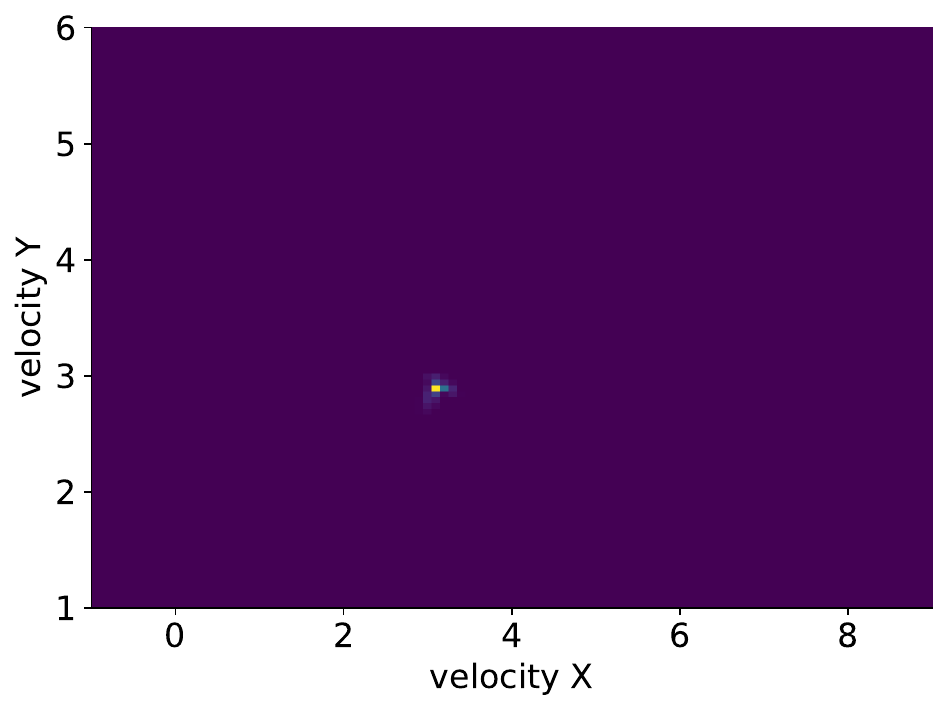}&
    \includegraphics[width=0.25\linewidth]{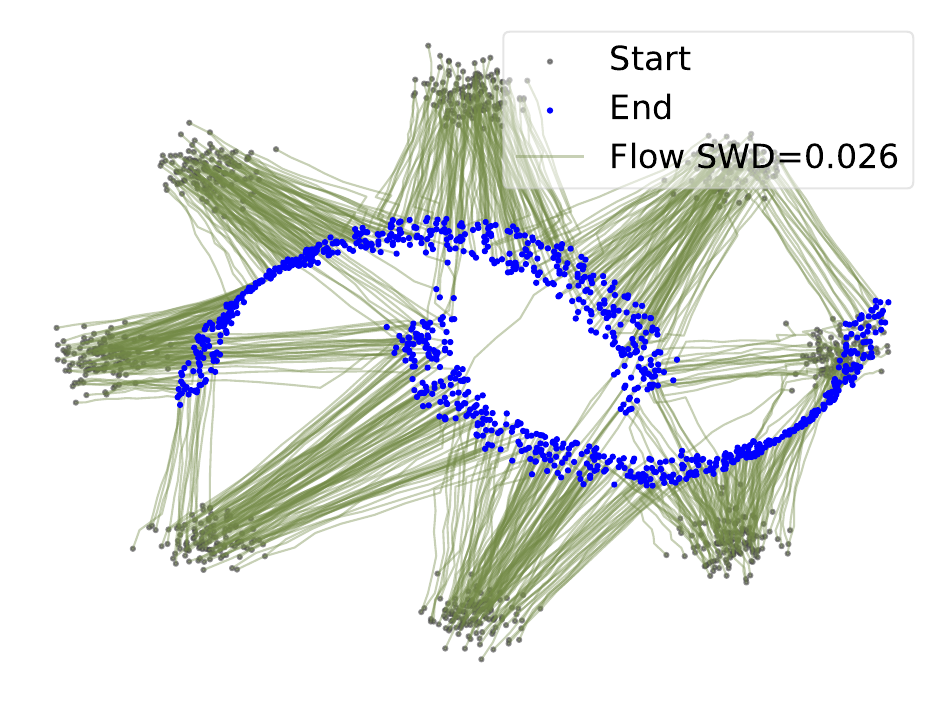}&
    \includegraphics[width=0.25\linewidth]{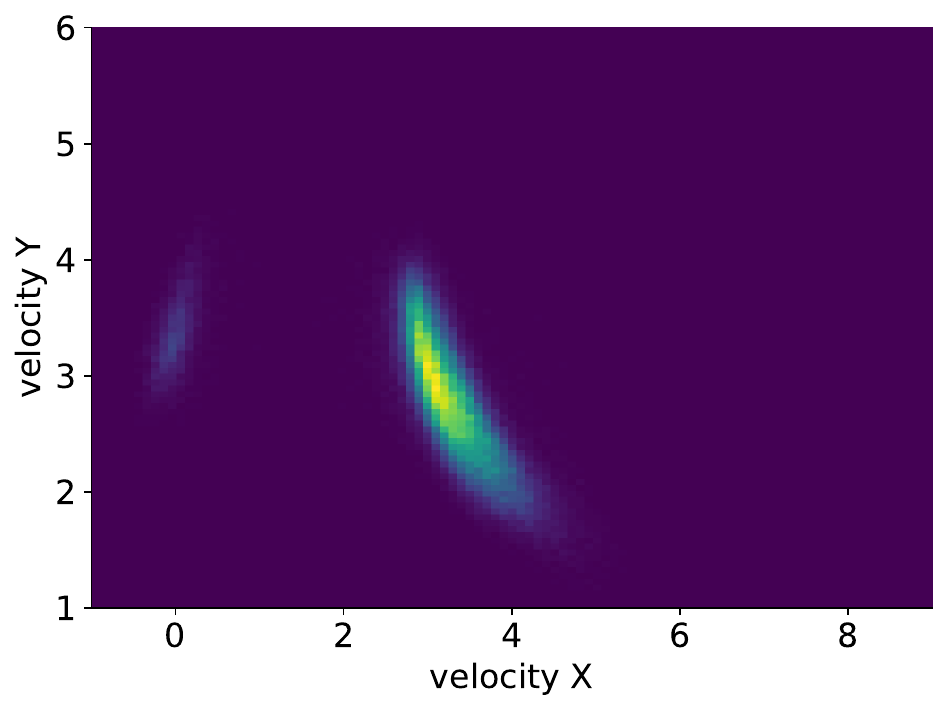}\\
    \includegraphics[width=0.25\linewidth]{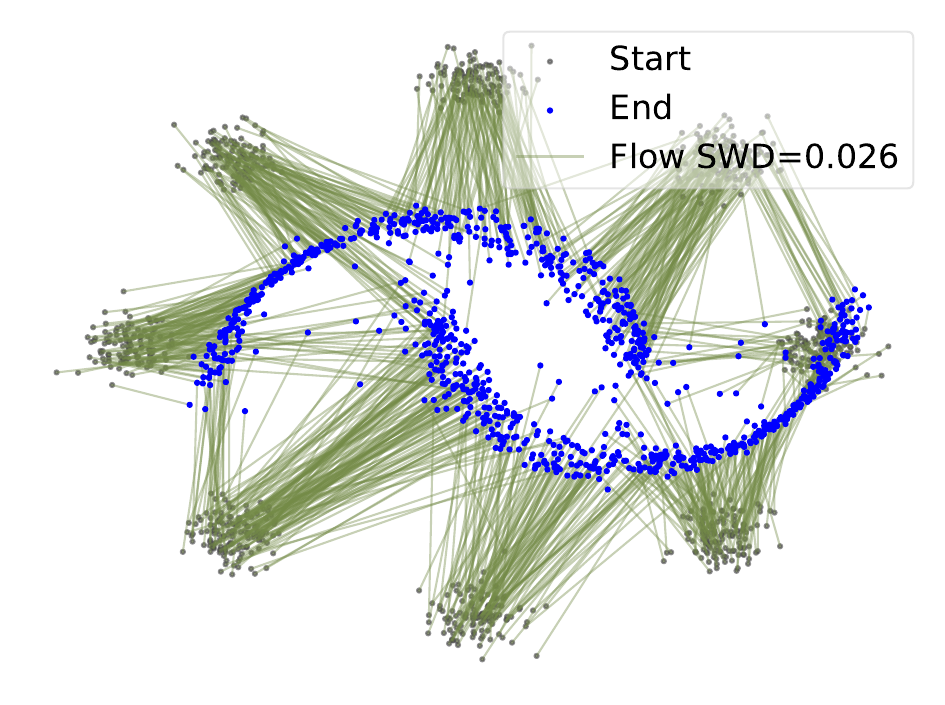}&
    \includegraphics[width=0.25\linewidth]{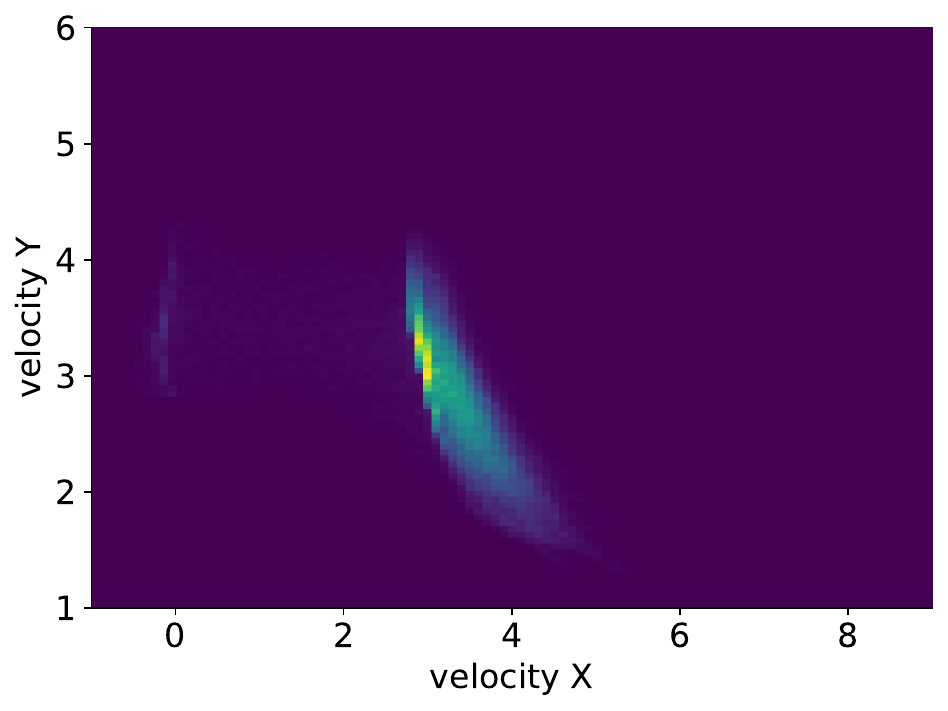}&
    \includegraphics[width=0.25\linewidth]{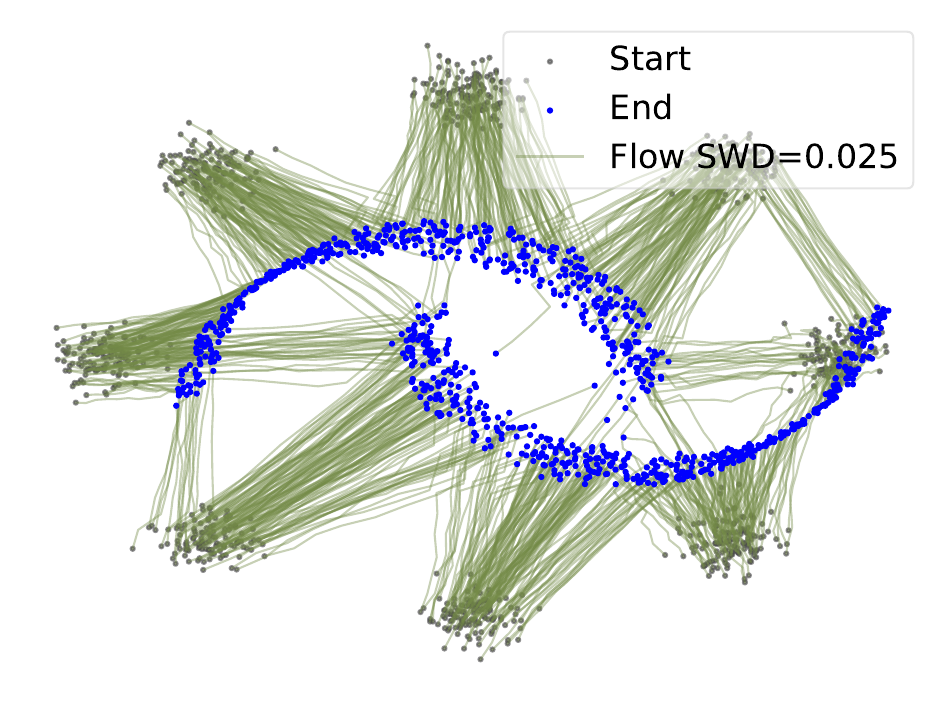}&
    \includegraphics[width=0.25\linewidth]{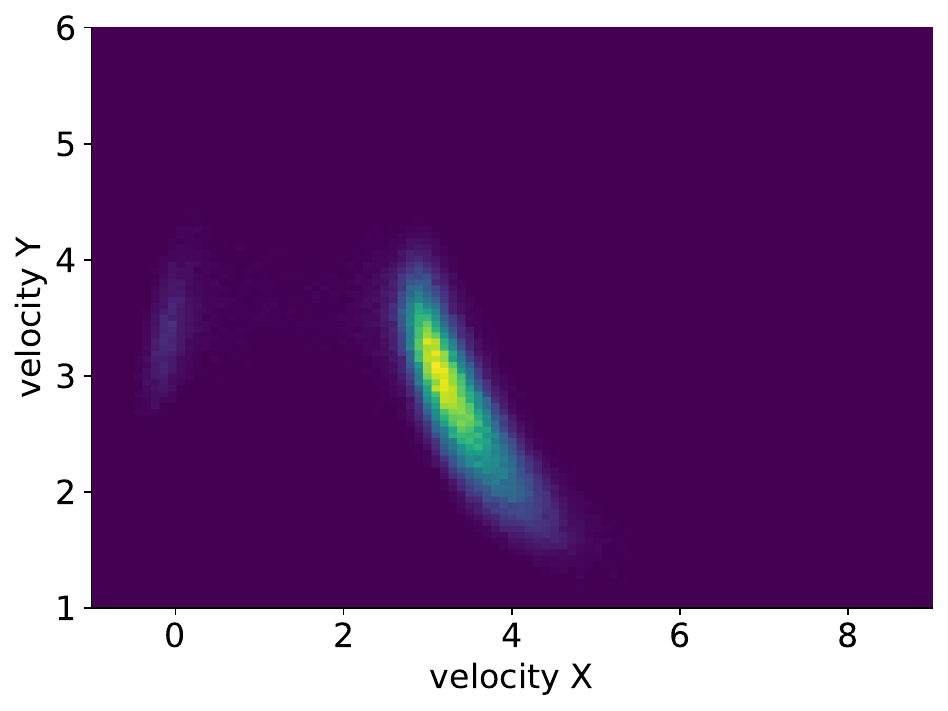}\\
    (a) Data distribution & (b) Velocity distribution & (c) Data distribution & (d) Velocity distribution \\
    total NFE=1 & 1 step integration & total NFEs=100 & 100 step integration
    \end{tabular}}
    \vspace{1mm}
    \caption{Results on $8\mathcal{N}\to$ moon dataset. Three rows are HRF2, HRF2 with data coupling, HRF2 with data \& velocity coupling. (a) and (c) are trajectories (green) of sample particles flowing from source distribution (grey) to target distribution (blue) with total NFEs 1 and 100. (b) and (d) are velocity distributions at the center of the bottom left Gaussian mode at $t=0$. Data coupling simplifies the velocity distribution and velocity coupling reduces the required sampling steps. }
    \label{fig:moon}
\end{figure*}

Recall that in \cref{fig:v_dist_1to2}, we have observed that for the 1D $\mathcal{N} \to 2\mathcal{N}$ dataset, data coupling simplifies the velocity distribution. Now, we extend this analysis to 2D datasets. We denote HRF2 with data coupling as HRF2-D and HRF2 with joint data and velocity coupling as HRF2-D\&V. As shown in \cref{fig:moon}, the 2D results corroborate the 1D findings. For the original HRF2, the velocity distribution at $t=0$ is simply a shifted version of the data distribution. After applying data coupling, the velocity distribution at a given space-time location $(x_t, t)$ becomes more unimodal, effectively aligning with a portion of the target distribution. For example, in \cref{fig:moon}(d), we observe that the velocity distribution primarily consists of the region of the target distribution closest to $x_t$. 

\begin{figure*}[t!]
    \centering
    \setlength{\tabcolsep}{0pt}
    {\small
    \begin{tabular}{cccc}
    \includegraphics[width=0.25\linewidth]{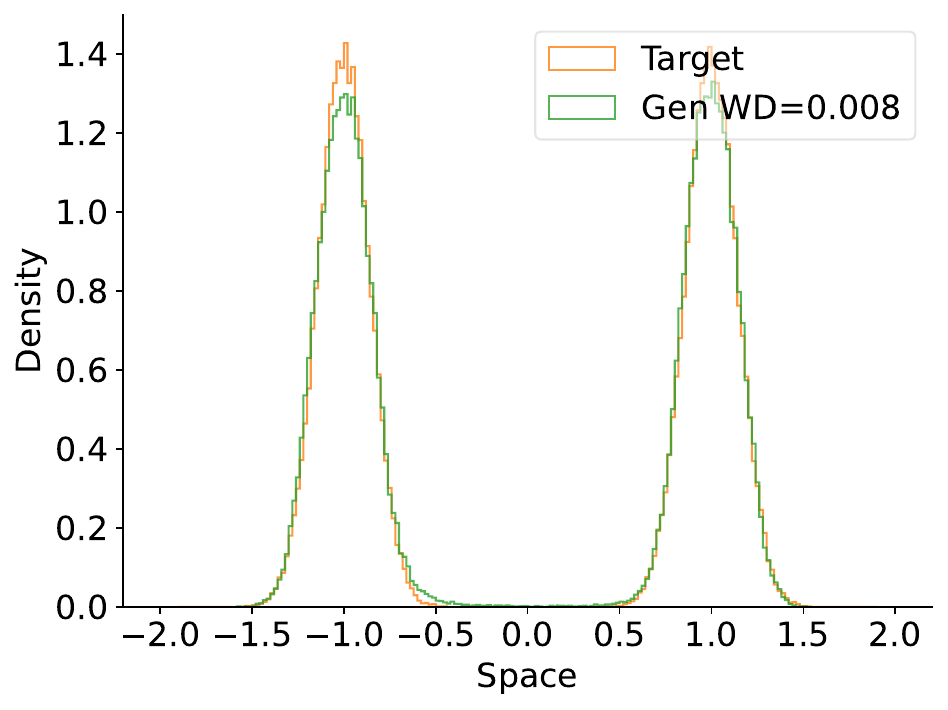}&
    \includegraphics[width=0.25\linewidth]{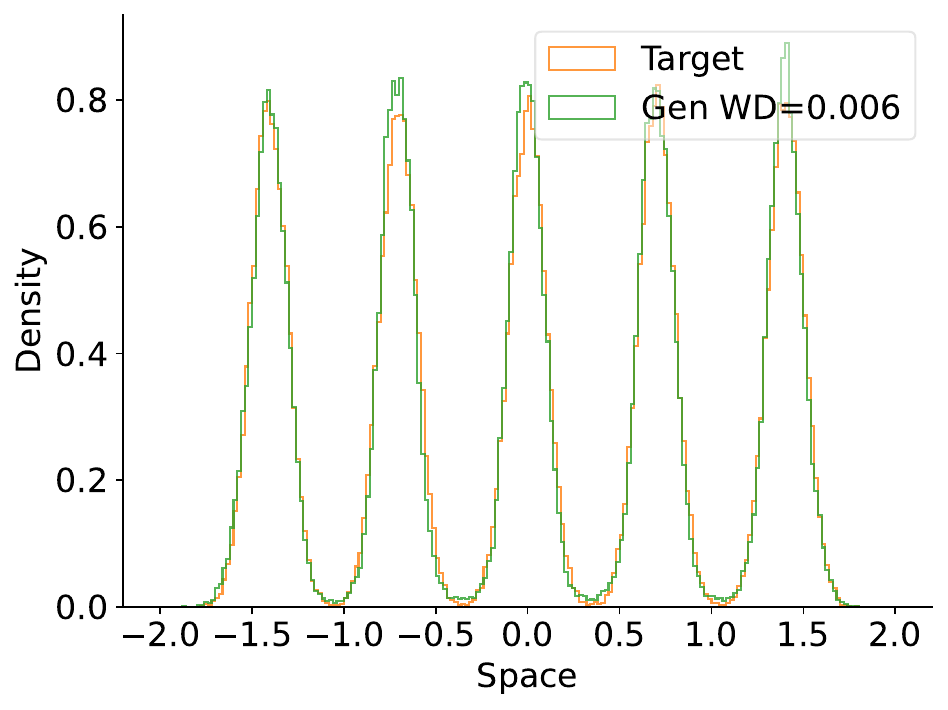}&
    \includegraphics[width=0.25\linewidth]{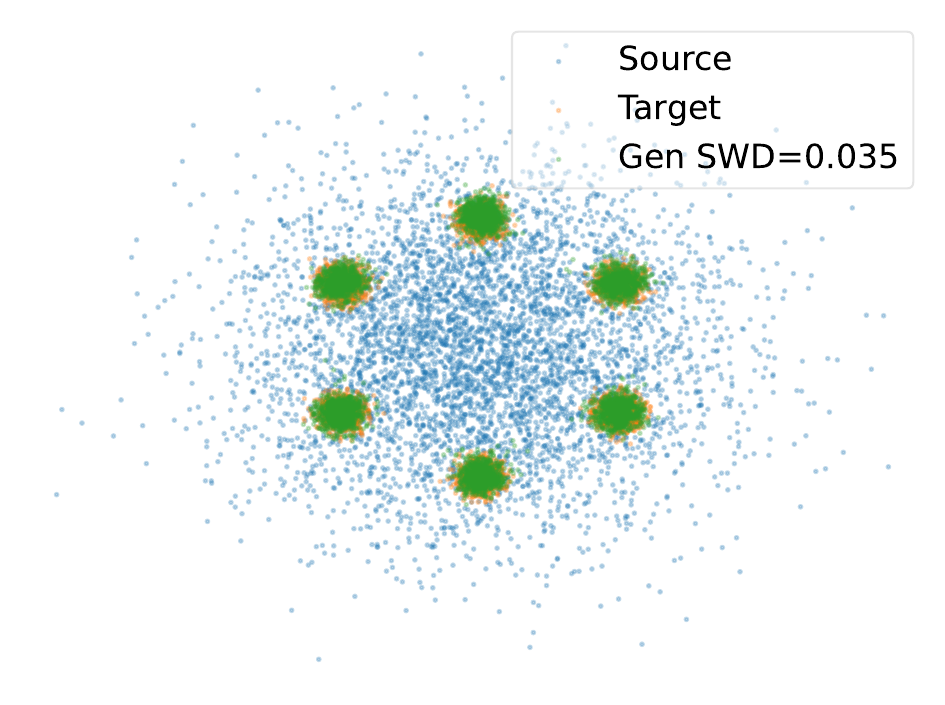}&
    \includegraphics[width=0.25\linewidth]{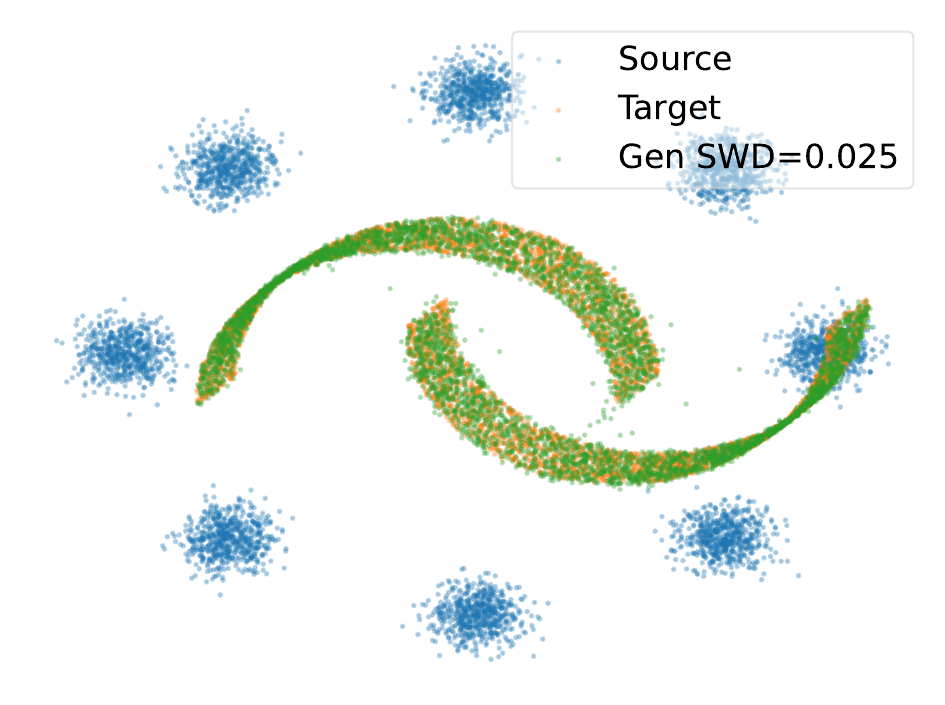}\\
    \includegraphics[width=0.25\linewidth]{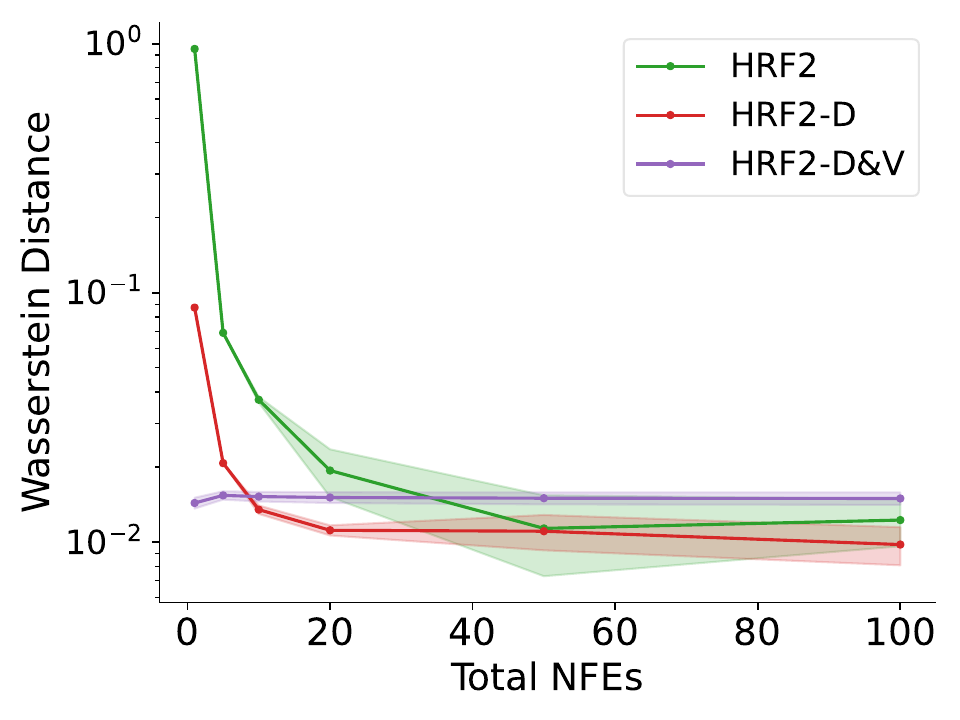}&
    \includegraphics[width=0.25\linewidth]{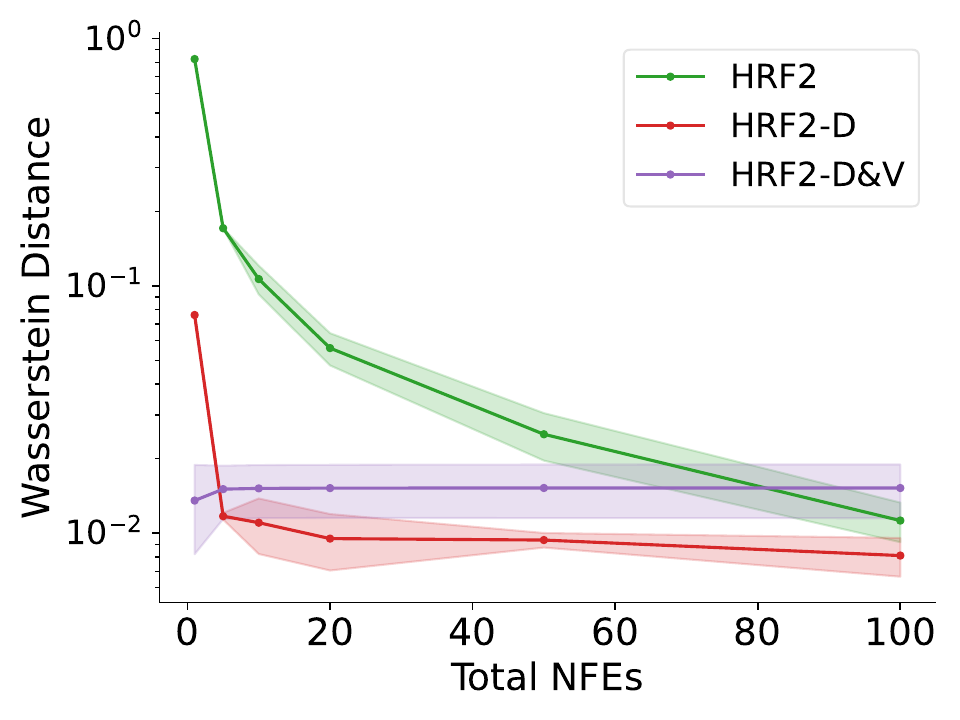}&
    \includegraphics[width=0.25\linewidth]{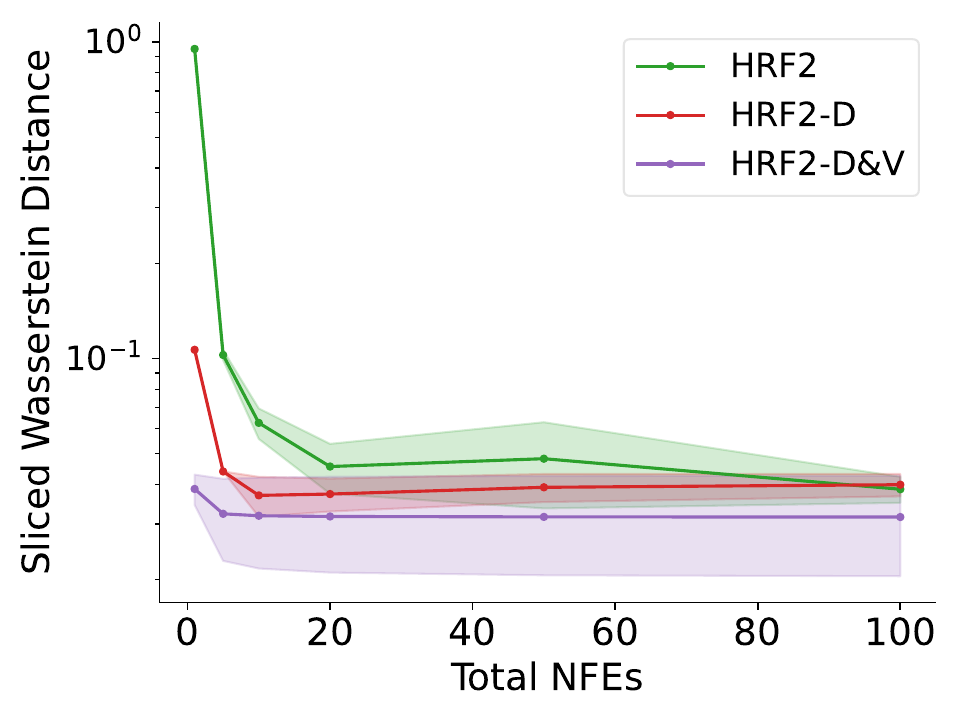}&
    \includegraphics[width=0.25\linewidth]{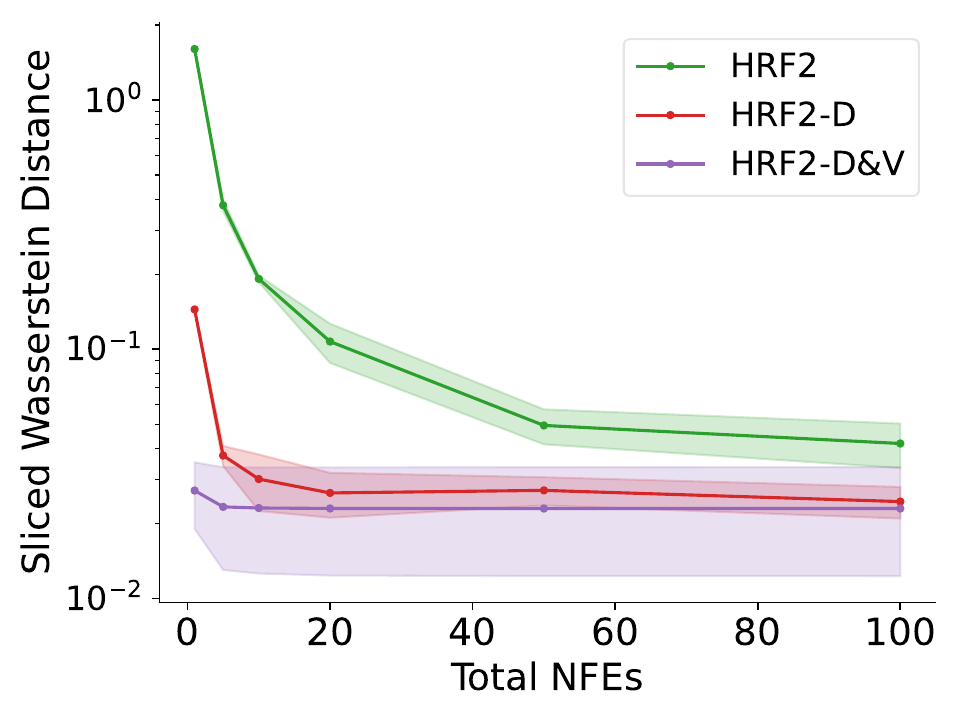}\\
    (a) 1D $\mathcal{N}\to 2\mathcal{N}$ & (b) 1D $\mathcal{N}\to 5\mathcal{N}$ & (c) 2D $\mathcal{N}\to 6\mathcal{N}$ & (d) 2D $8\mathcal{N}\to$ moon \\
    \end{tabular}}
    \caption{Results on synthetic datasets: (a) 1D $\mathcal{N}\to 2\mathcal{N}$ (b) 1D $\mathcal{N}\to 5\mathcal{N}$ (c) 2D $\mathcal{N}\to 6\mathcal{N}$ (d) 2D $8\mathcal{N}\to$ moon. Top row: HRF2-D\&V generated data distributions. Bottom row: performance vs.\ total NFEs. We use Wasserstein and sliced 2-Wasserstein distances for 1D and 2D data, respectively. }
    \label{fig:perf}
\end{figure*}

\begin{figure*}[t]
    \centering
    \setlength{\tabcolsep}{0.5pt}
    {\small
    \begin{tabular}{cccc}
    \includegraphics[width=0.24\linewidth]{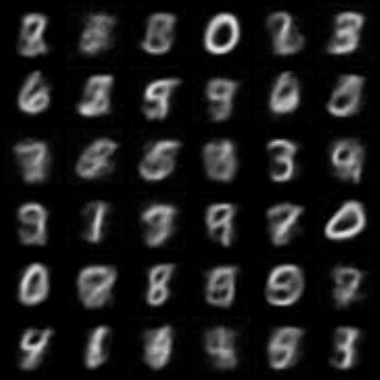}&
    \includegraphics[width=0.24\linewidth]{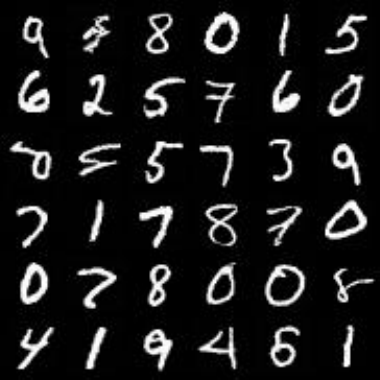}&
    \includegraphics[width=0.24\linewidth]{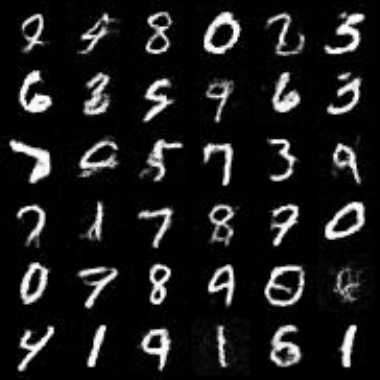}&
    \includegraphics[width=0.25\linewidth]{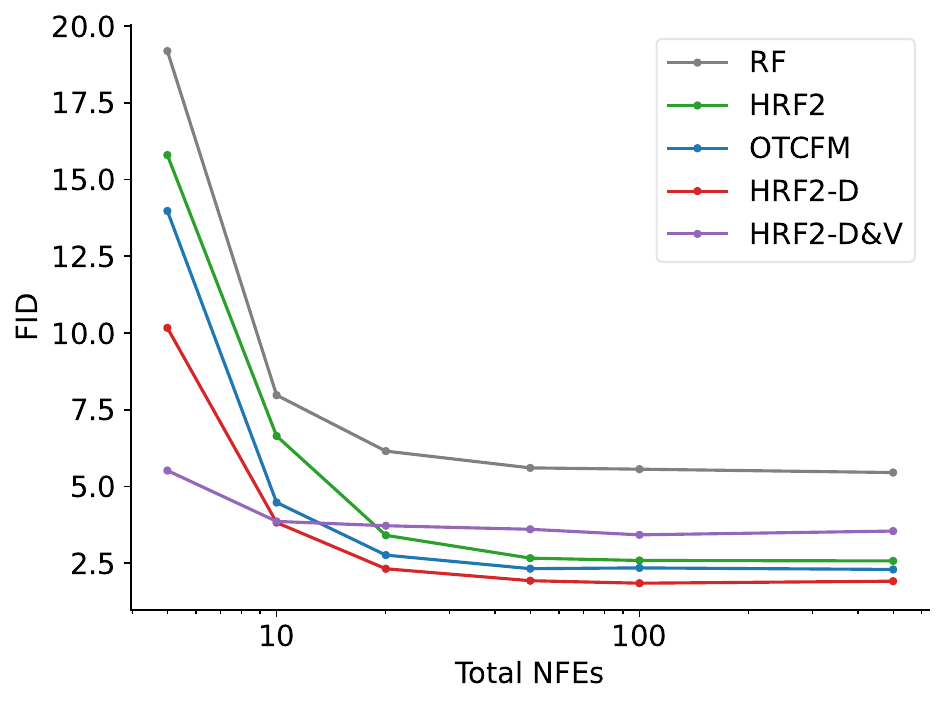}\\
    \includegraphics[width=0.24\linewidth]{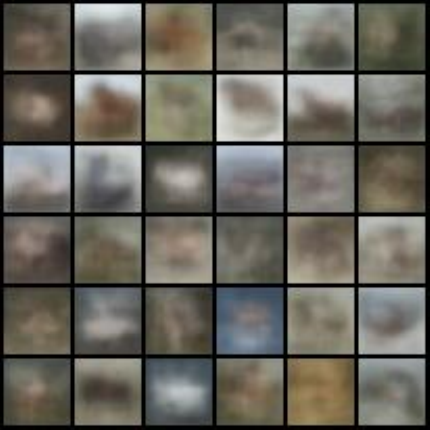}&
    \includegraphics[width=0.24\linewidth]{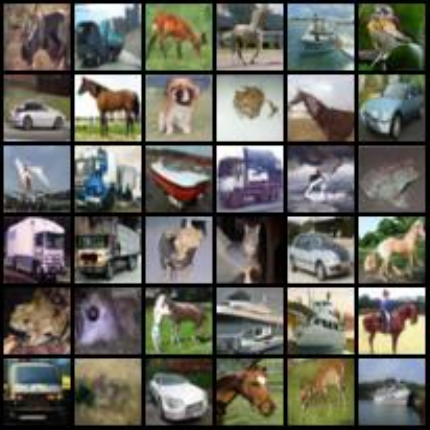}&
    \includegraphics[width=0.24\linewidth]{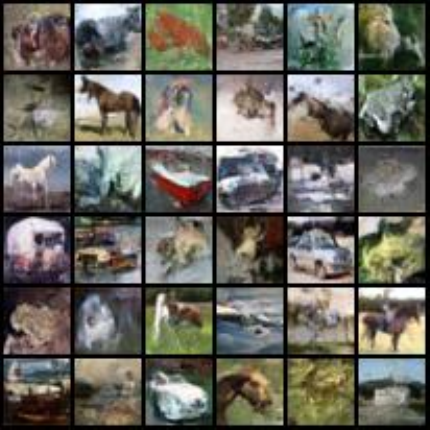}&
    \includegraphics[width=0.25\linewidth]{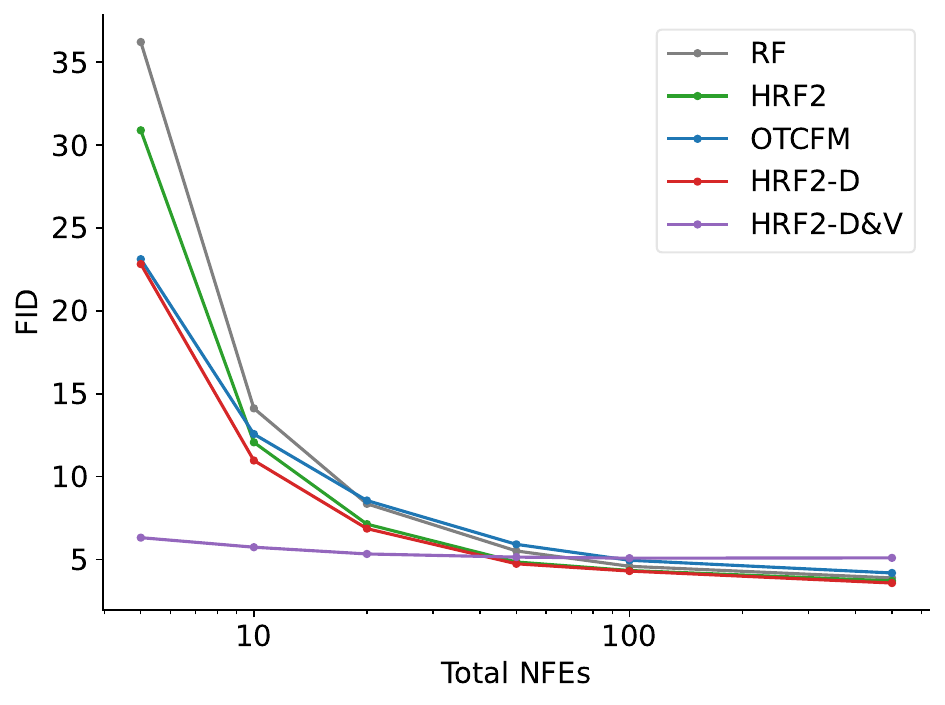}\\
    \includegraphics[width=0.24\linewidth]{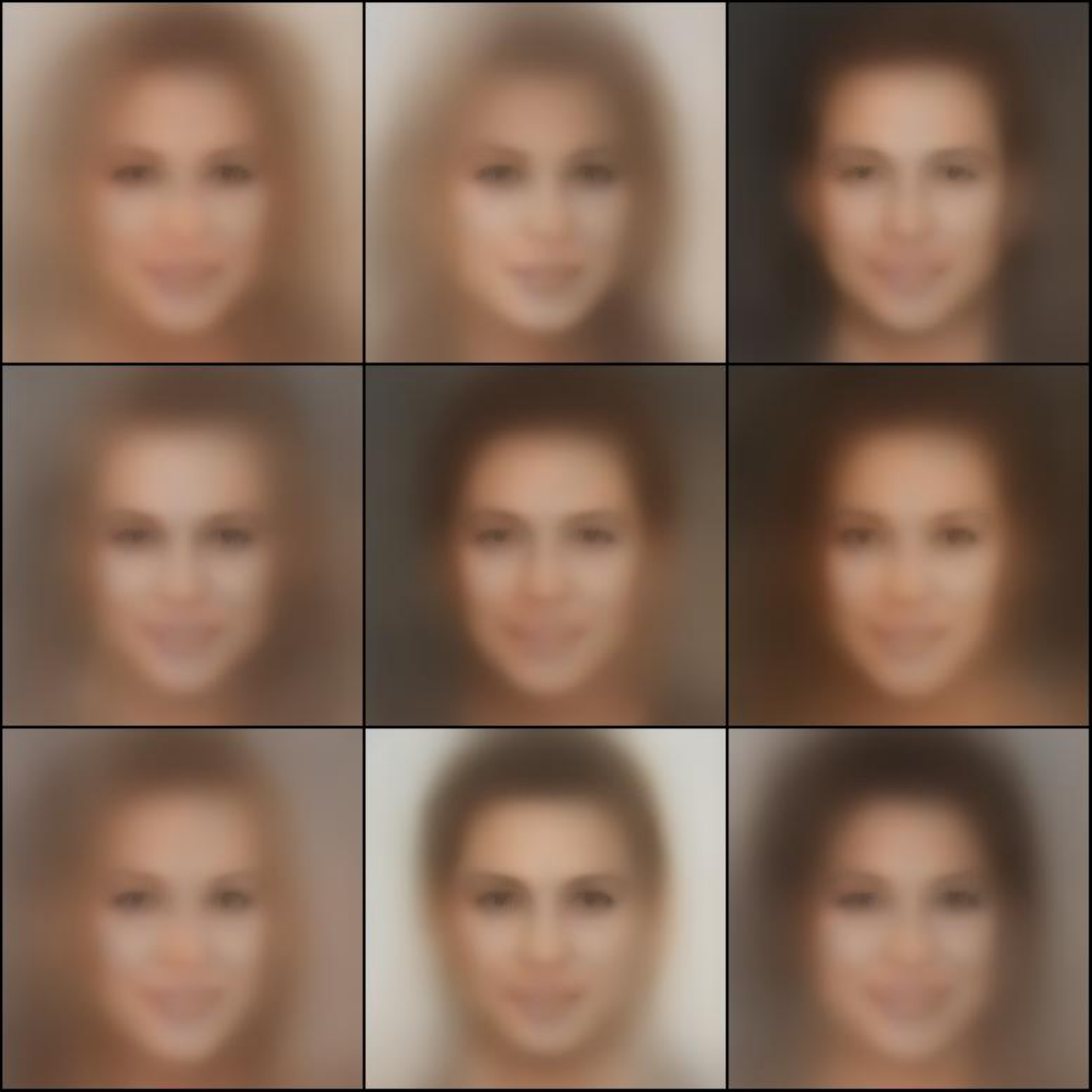}&
    \includegraphics[width=0.24\linewidth]{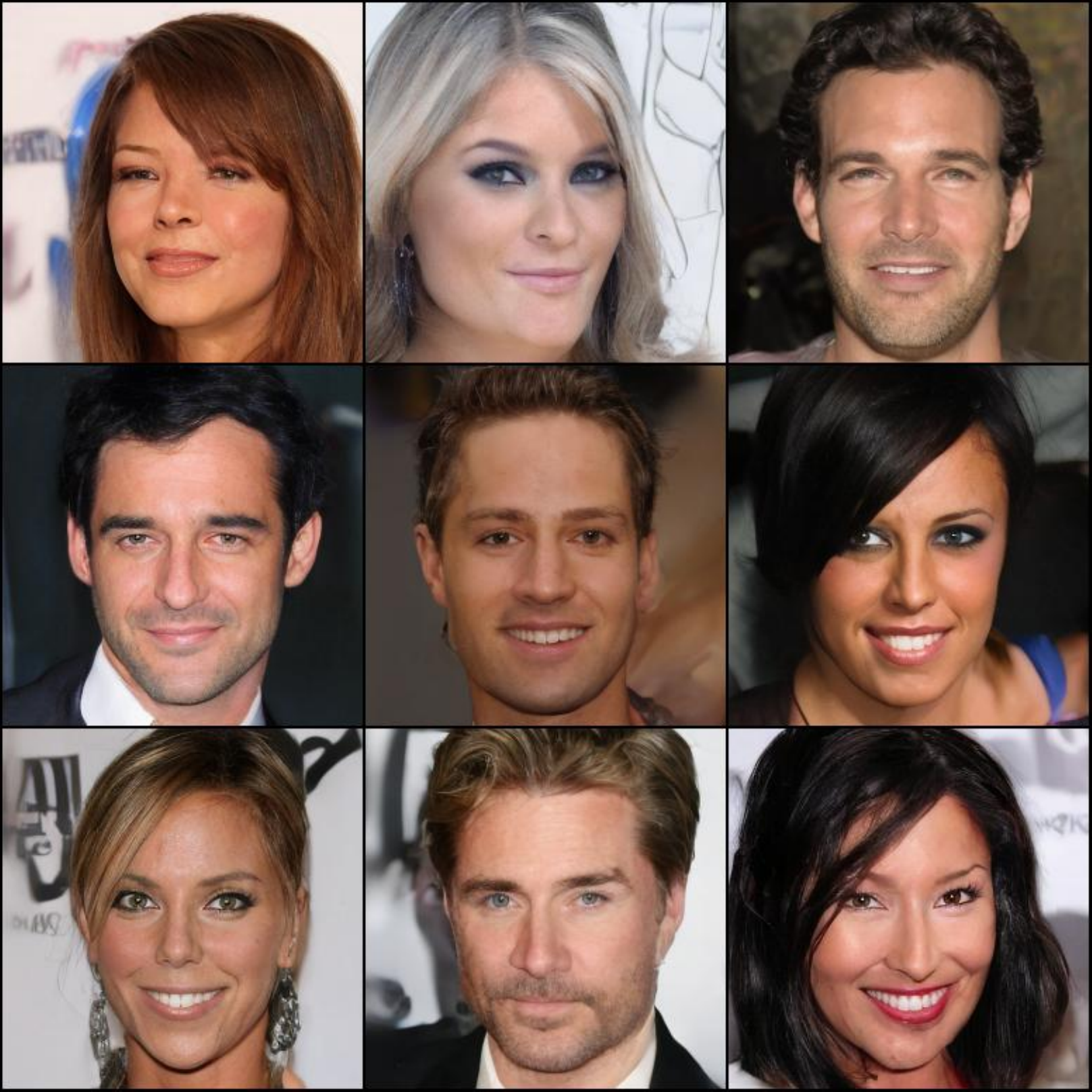}&
    \includegraphics[width=0.24\linewidth]{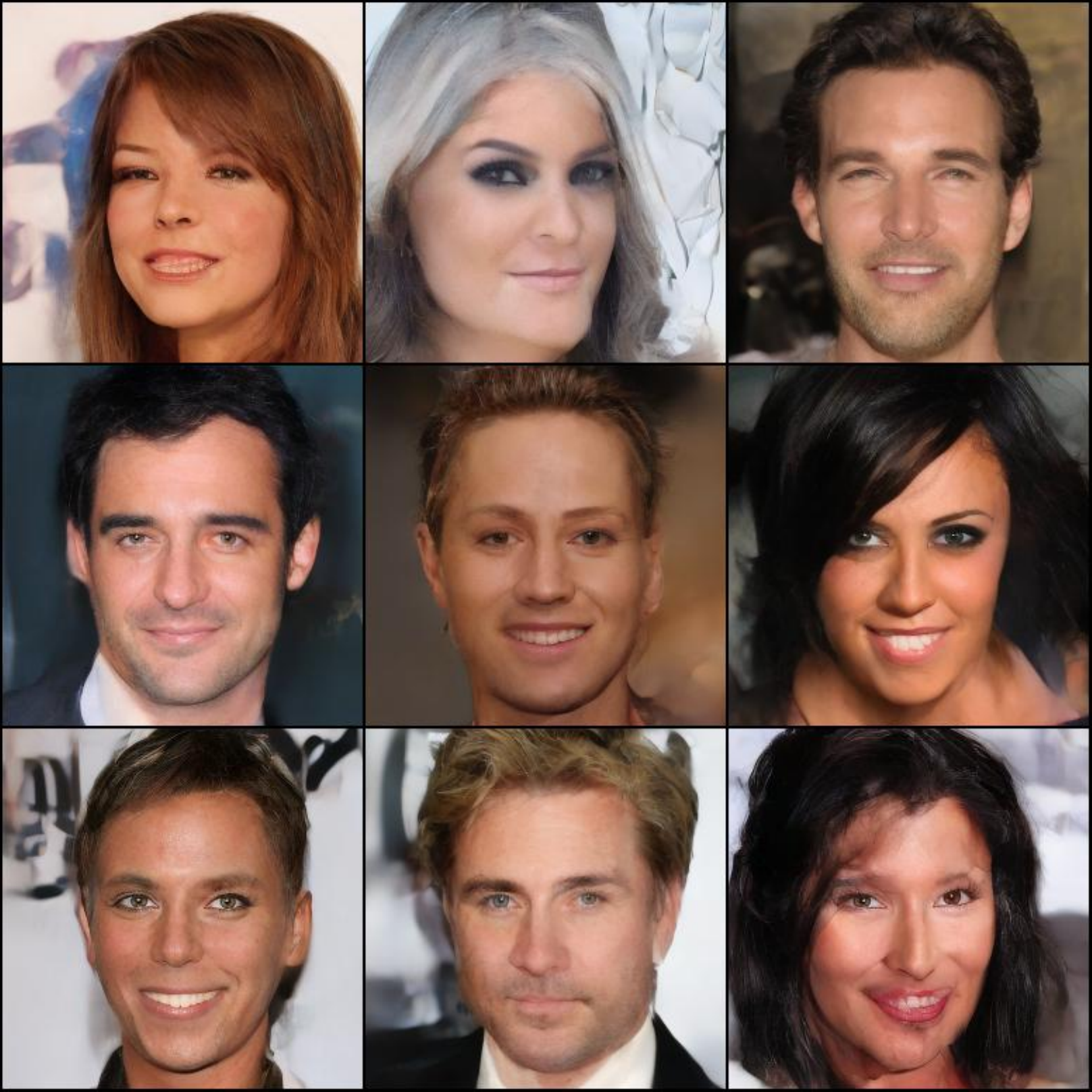}&
    \includegraphics[width=0.25\linewidth]{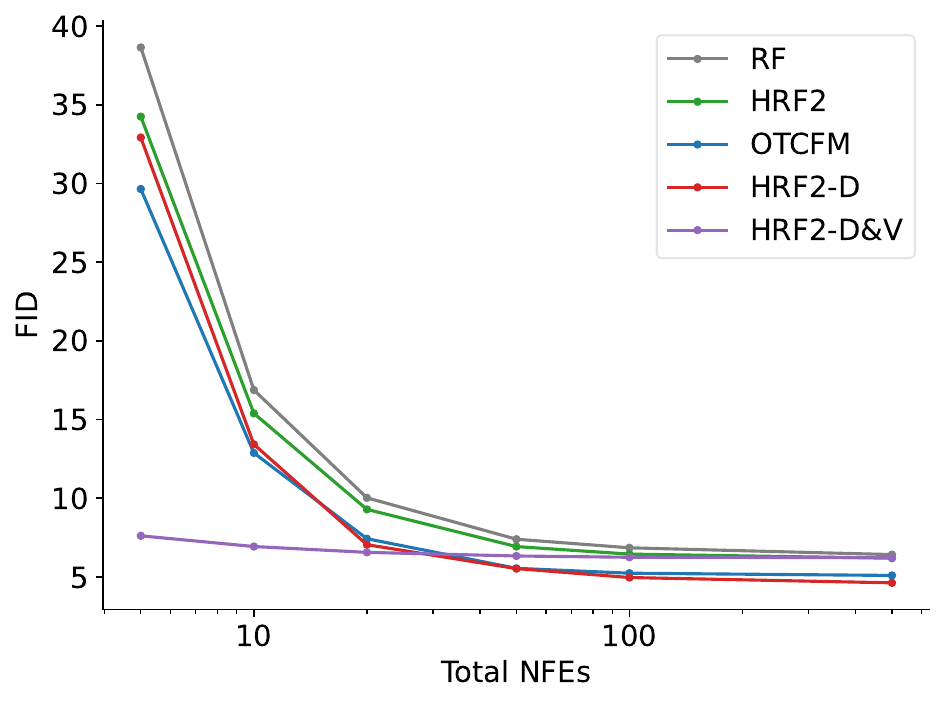}\\
    (a) HRF2-D total NFE=1 & (b) HRF2-D total NFE=500 & (c) HRF2-D\&V total NFE=1 & (d) Performance
    \end{tabular}}
    \caption{Results on MNIST, CIFAR-10 and CelebA-HQ 256 datasets. (a) HRF2-D with total NFE=1. (b) HRF2-D with total NFE=500. (c) HRF2-D\&V with total NFE=1. Here we report the results with HRF2-D\&V-OT. (d) FID scores with respect to total NFEs. With joint data coupling and velocity coupling, HRF2-D\&V can generate reasonably good results with only 1 step. }
    \label{fig:img_data}
\end{figure*}

In contrast, velocity coupling does not modify the velocity distribution itself but significantly reduces the number of required sampling steps. As shown in \cref{fig:moon}(b), a single integration step already produces a reasonable velocity distribution with joint data and velocity couplings. %
The results in \cref{fig:perf} demonstrate that joint data and velocity couplings effectively enhance the model performance, particularly when NFEs are low. Additional results on synthetic data are provided in \cref{app:res:syn}.%

\subsection{Image Data}
\label{sec:exp:img}
For high-dimensional image data, we conduct experiments on MNIST \citep{lecun1998gradient}, CIFAR-10 \citep{krizhevsky2009learning}, and CelebA-HQ 256 \citep{karras2018progressive}, using Fr\'{e}chet Inception Distance (FID) as the evaluation metric. For MNIST and CIFAR-10, we directly operate in the pixel space, with input dimensions of $1\times28\times28$ and $3\times32\times32$, respectively. For CelebA-HQ, we first encode the original $3\times256\times256$ images into a $4\times32\times32$ latent space using a pretrained VAE encoder, and conduct training and inference in the latent space. The experiments on CelebA-HQ demonstrate that our methods scale well on higher-dimensional data. 

We compare our method to RF \citep{liu2022rectified}, HRF2 \citep{zhang2025towards} and OT-CFM \citep{tong2023improving} baselines. HRF2 is the base pre-trained model for our HRF2-D and HRF2-D\&V. OT-CFM is essentially equivalent to HRF1-D. As shown in \cref{fig:img_data}(d), data coupling significantly improves performance across both low and high total NFEs. However, applying velocity coupling on top of data coupling only yields substantial improvements in the low-NFE regime. From \cref{fig:img_data}(a)-(c), we observe that data coupling alone enhances performance at low NFEs, but HRF2-D still struggles in extreme cases. Notably, incorporating velocity coupling enables the model to generate compelling results even under the extreme condition of total NFE = 1. More results are presented in \cref{app:res:img}. 

The details of the model architectures are deferred to~\cref{app:imp:img}.
Since the model architecture remains unchanged, HRF2-D and HRF2-D\&V have the same memory usage and inference time as HRF2. For training, we apply data coupling and velocity coupling following \cref{alg:data_coup,alg:hrf_v_coup}. To obtain accurate velocity pairs $(v_0, v_1)$ for velocity coupling, more integration steps are required here than for synthetic data. 
More details are provided in \cref{app:imp:img}.

%% file: 05_rel.tex
\section{Related Work}
\label{sec:rel}
\textbf{Flow Matching:}
Concurrently,~\citet{liu2023flow,LipmanICLR2023,albergo2023building} presented learning of the ODE velocity that governs the generation of new data through a time-differentiable stochastic process defined by interpolating between samples from the source and data distributions. This provides flexibility by enabling precise connections between any two densities over finite time intervals. \citet{liu2023flow} concentrated on a linear interpolation, which provides straight paths connecting points from the source and the target distributions.~\citet{LipmanICLR2023} introduced the interpolation through the lens of conditional probability paths leading to a Gaussian. %
\citet{albergo2023building,albergo2023stochastic} introduced stochastic interpolants with more general forms. They all learn the expected velocity field, which leads to curved sampling paths for data generation. Flow matching has been extended to handle discrete data~\citep{gat2024discrete,stark2024dirichlet} and manifold data~\citep{chen2024flow}. 

\textbf{Straightening Flows:}  
\citet{liu2023flow} proposed an iterative method called reflow, which connects points from the source and target distributions using a trained rectified flow model to smooth the transport path. They demonstrated that repeating this process results in an optimal transport map. However, in practice, errors in the learned velocity field can introduce bias. Other related studies address this by adjusting how noise and data are sampled during training, rather than using iterations. For instance, \citet{pooladian2023multisample,tong2023improving} computed mini-batch optimal transport couplings between the source and data distributions to reduce transport costs and flow variance. \citet{park2024constant} address the curved paths in flow matching by learning both initial velocity and acceleration, such that the sampling paths can cross. However, it requires a pre-trained diffusion model to acquire noise-data pairs. \citet{ChengICCV2025} study conditional data.

\textbf{Distribution of flow fields:} ~\citet{zhang2025towards} capture the distribution of the random flow fields induced by the linear interpolation of source and target data. The sampling process is governed by coupled ODEs, which allows sampling paths to cross.~\citet{schwing2025variational} model the flow field distribution using a variational autoencoder.   

Building upon work by \citet{zhang2025towards}, we show that modeling the velocity distributions after the mini-batch coupling in data space improves the performance of  HRF2 and OT-CFM. Hierarchically coupling the data and velocity leads to significantly improved results at low NFEs.

%% file: 06_conc.tex
\section{Conclusion}
\label{sec:conc}
We study ways to control the complexity of the multi-modal velocity distribution and their impact on capturing this distribution with hierarchical flow matching. We find hierarchical flow matching with mini-batch coupling in the data space consistently improves the generation quality compared to vanilla hierarchical rectified flow matching and vanilla flow matching with mini-batch optimal transport. Joint coupling in the data space and the velocity space leads to further improvements if few function evaluations are used. %

\textbf{Limitations and broader impacts: } Our proposed methods offer faster and more accurate data generation. It can help advance scientific modeling and simulations, contributing to advances in areas like physics, healthcare, and drug discovery. For the limitations: the current velocity coupling approach requires simulated target velocity samples during training. Simulation-free velocity coupling is an interesting direction for future research.

\textbf{Acknowledgments:} Work supported in part by NSF grants 1934757, 2008387, 2045586, 2106825, MRI 1725729, NIFA award 2020-67021-32799, and the Alfred P.\ Sloan Foundation.

%% file: 10_appendix.tex
\newpage
\section*{Appendix: Hierarchical Rectified Flow Matching with Mini-Batch Couplings}

This appendix is structured as follows: 
in \cref{sec:proofpvgivenxt} we provide the proof of \cref{the:pv}; %
in \cref{sec:proof_thm_marginal} we provide the proof of \cref{thm:marginal};
in \cref{app:res} we provide additional results for both synthetic and image data; and in \cref{app:imp} we discuss implementation details.%

\section{Proof of Theorem~\ref{the:pv}}
\label{sec:proofpvgivenxt}
\textbf{Proof of~\cref{the:pv}:}
For simplicity, we show the proof for 1D random variables $x_0$ and $x_1$ drawn from a joint distribution $\gamma(x_0, x_1)$. The joint distribution of $v(x_t, t)$ and $x_t$ is $\pi_1(v; x_t, t) \rho_t(x_t)$, since $\pi_1(v; x_t, t)$ corresponds to the conditional distribution of the velocity given location $x_t$. According to the linear interpolation in~\cref{eq:lin_int}, we have,
\begin{align}
\begin{bmatrix}
v(x_t, t) \\
x_t 
\end{bmatrix} = \begin{bmatrix}
1 & -1 \\
t & (1-t) 
\end{bmatrix} \begin{bmatrix}
x_1 \\
x_0  
\end{bmatrix} = A \begin{bmatrix}
x_1 \\
x_0  
\end{bmatrix}, 
\end{align}
where the matrix $A$ has determinant 1. Since $[v(x_t, t), x_t]^T$ is a linear transformation of $[x_1, x_0]^T$, we have the following expression for the joint distribution of $v(x_t, t)$ and $x_t$,  
\begin{equation}
\label{eq:joint}
 \pi_1(v; x_t, t) \rho_t(x_t)  = \frac{1}{\det(A)} \gamma\left(A^{-1}\begin{bmatrix}
v(x_t, t) \\
x_t 
\end{bmatrix} \right) = \gamma(x_t - tv, x_t + (1-t) v).
\end{equation}
After rearranging we get $\pi_1(v; x_t, t)= \frac{\gamma(x_t - tv, x_t + (1-t) v)}{\rho_t(x_t)}$. For the higher dimensional case, the relation in \cref{eq:joint} still holds. This completes the proof.
\hfill$\blacksquare$

\section{Proof of Theorem~\ref{thm:marginal}}
\label{sec:proof_thm_marginal}
\textbf{Proof of~\cref{thm:marginal}:}
We consider the characteristic function of $Z_{t + \Delta t} = Z_t + V \Delta t$ for $t \in [0, 1]$ and $\Delta t \in [0, 1-t]$, assuming that $Z_t$ has the same distribution as $X_t$. If the characteristic functions of $Z_{t + \Delta t}$ and $X_{t + \Delta t}$ agree, then $Z_{t + \Delta t}$ and $X_{t + \Delta t}$ have the same distribution.
 
To show this, we evaluate the characteristic function of $Z_{t + \Delta t}$,
\begin{align}
 \mathbb{E} \left[e^{\imath \langle k, Z_{t + \Delta t} \rangle} \right] &= \mathbb{E}_{\rho_t, \pi_1 } \left[ e^{\imath \langle k, X_t + V \Delta t \rangle}  \right] \nonumber \\
& = \int \int e^{\imath \langle k, x_t + v \Delta t \rangle } \pi_1(v; x_t, t) \rho_t(x_t)  dv dx_t \nonumber \\
& \stackrel{a}{=} \int \int e^{\imath \langle k, x_t + v \Delta t \rangle }  \frac{ \gamma(x_t - vt, x_t + (1-t) v) } {\rho_t (x_t)} \rho_t (x_t) dv dx_t \nonumber \\
& = \int \int e^{\imath \langle k, (x_t + v \Delta t) \rangle } \gamma(x_t-t v, x_t+(1-t)v)  dv dx_t \nonumber \\
& \stackrel{b}{=} \int \int e^{\imath \langle k, (1-t-\Delta t)x_0 + (t + \Delta t) x_1 \rangle }  \gamma(x_0, x_1)  dx_0 dx_1 \nonumber \\
& = \mathbb{E}_{\rho_{t + \Delta t}} \left[e^{\imath \langle k,  X_{t + \Delta t} \rangle }  \right]. 
\end{align}
We use the notation $\langle \cdot, \cdot \rangle$ to denote the inner product. The equality $a$ is valid due to~\cref{thm:marginal}.  The equality $b$ holds because $x_0 = x_t - tv$ and $x_1 = x_t + (1-t) v$ with the linear interpolation. Therefore, we find that $Z_{t + \Delta t}$ and $X_{t + \Delta t}$ follow the same distribution.  In addition, since $Z_0$ and $X_0$ follow the same distribution $\rho_0$, we can conclude that $Z_t$ and $X_t$ follow the same marginal distribution at $t$ for $t \in [0, 1]$. This completes the proof. \hfill$\blacksquare$

%% file: 11_appendix_res.tex
\section{Additional Experimental Results}
\label{app:res}

\subsection{Synthetic Data Results}
\label{app:res:syn}

\begin{figure*}[t]
    \centering
    \setlength{\tabcolsep}{0pt}
    {\small
    \begin{tabular}{ccc}
    \includegraphics[width=0.25\linewidth]{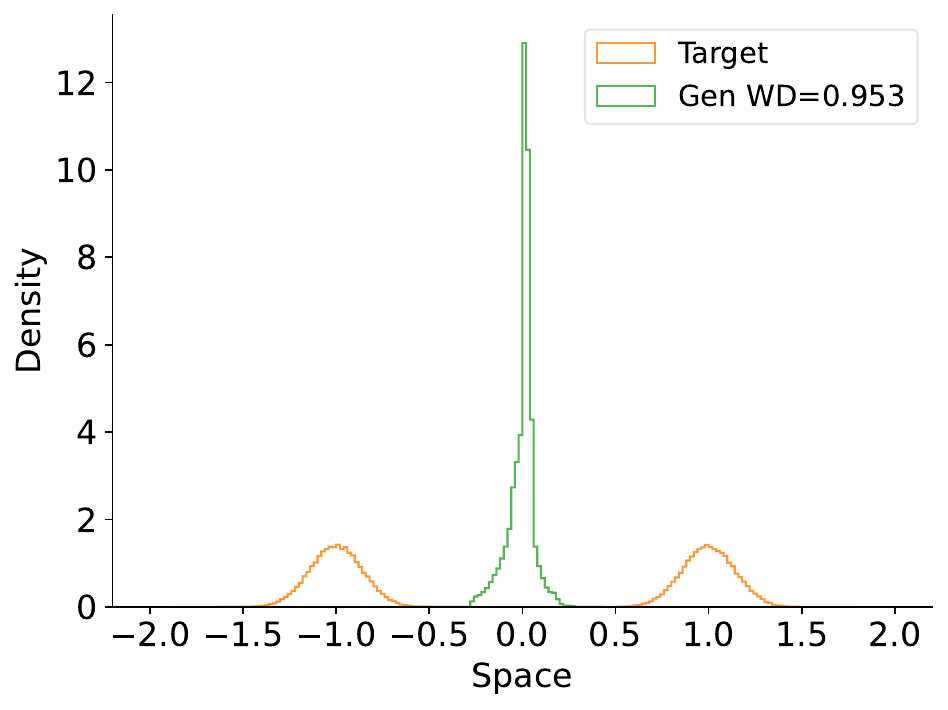}&
    \includegraphics[width=0.25\linewidth]{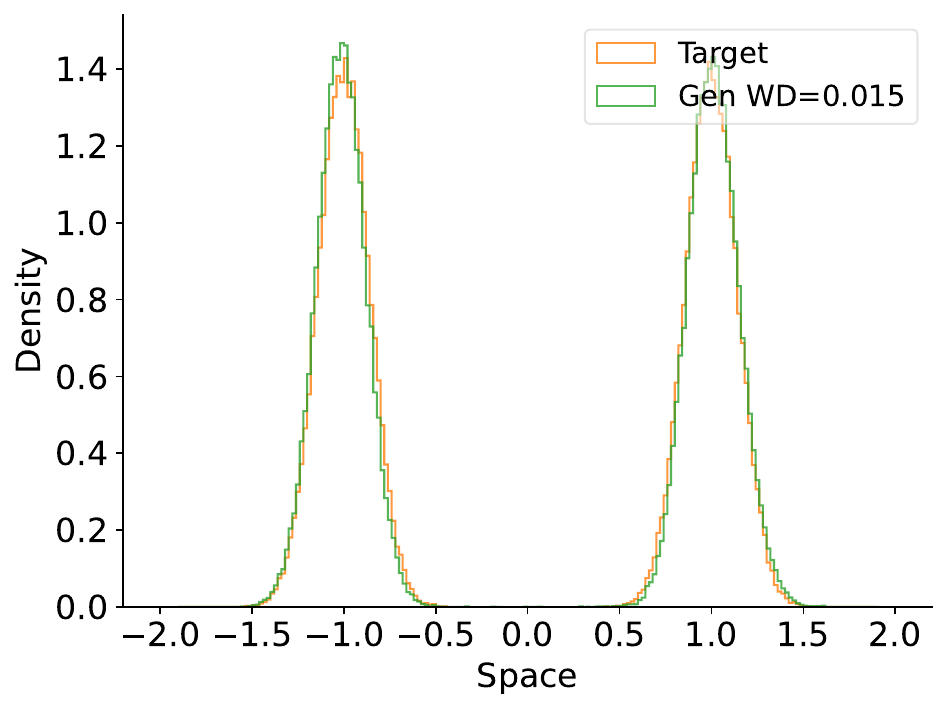}&
    \includegraphics[width=0.25\linewidth]{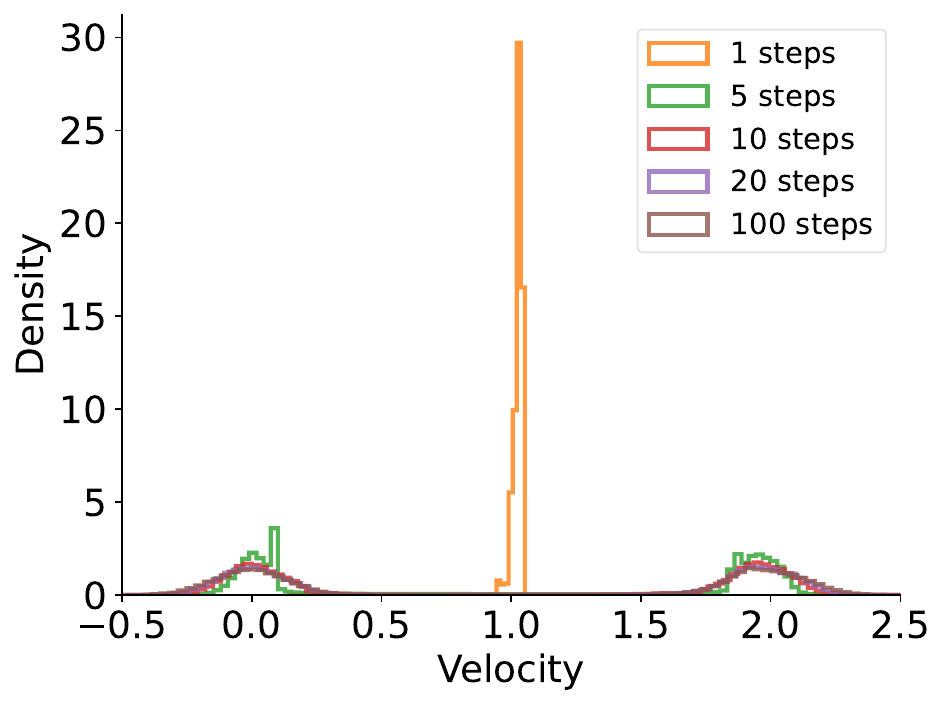}\\
    \includegraphics[width=0.25\linewidth]{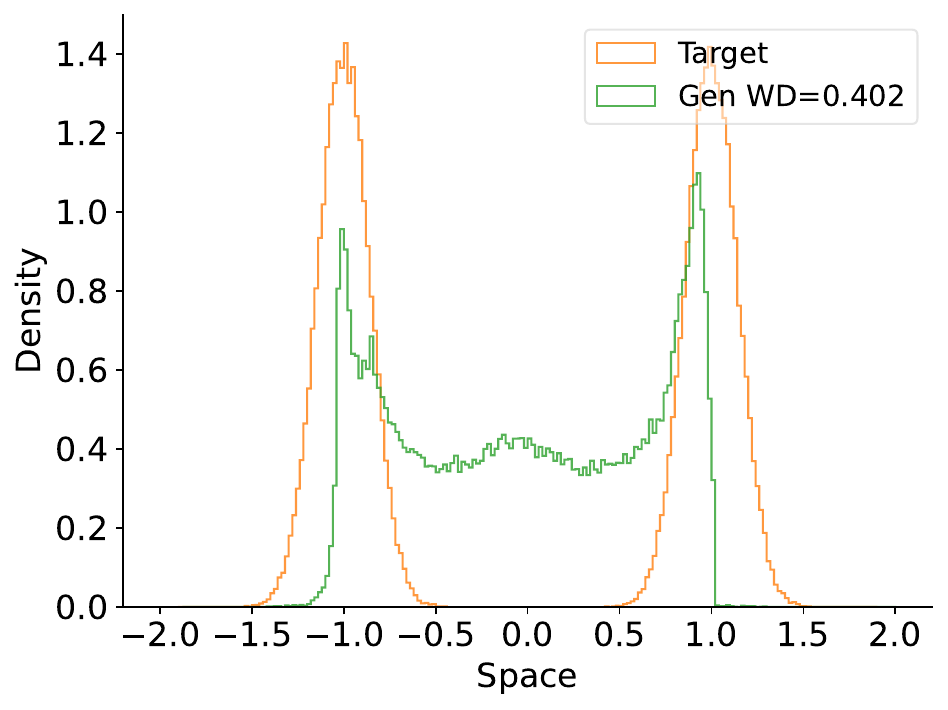}&
    \includegraphics[width=0.25\linewidth]{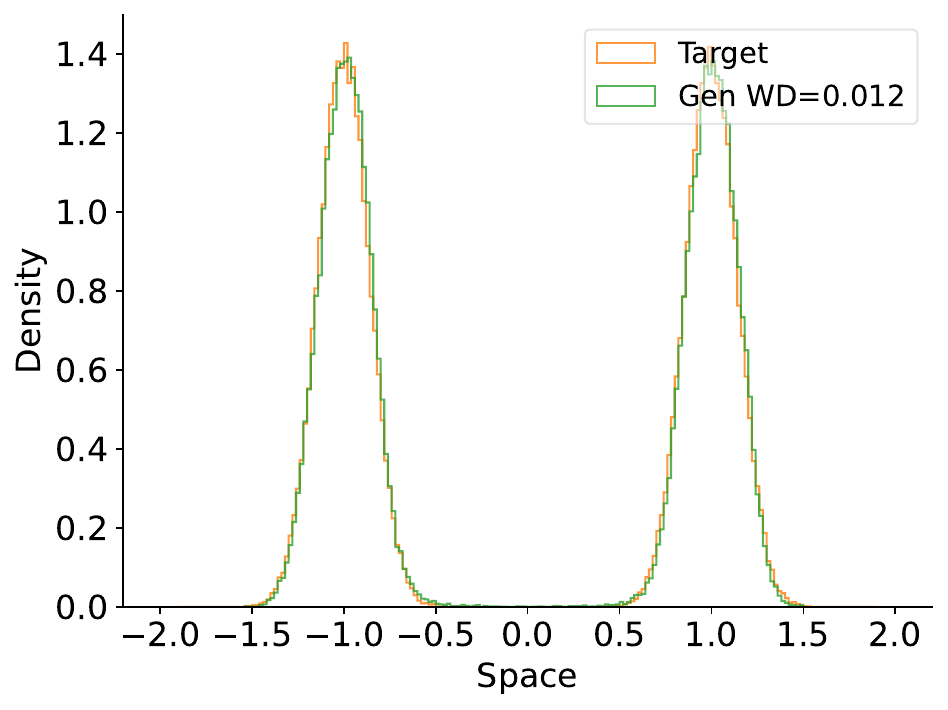}&
    \includegraphics[width=0.25\linewidth]{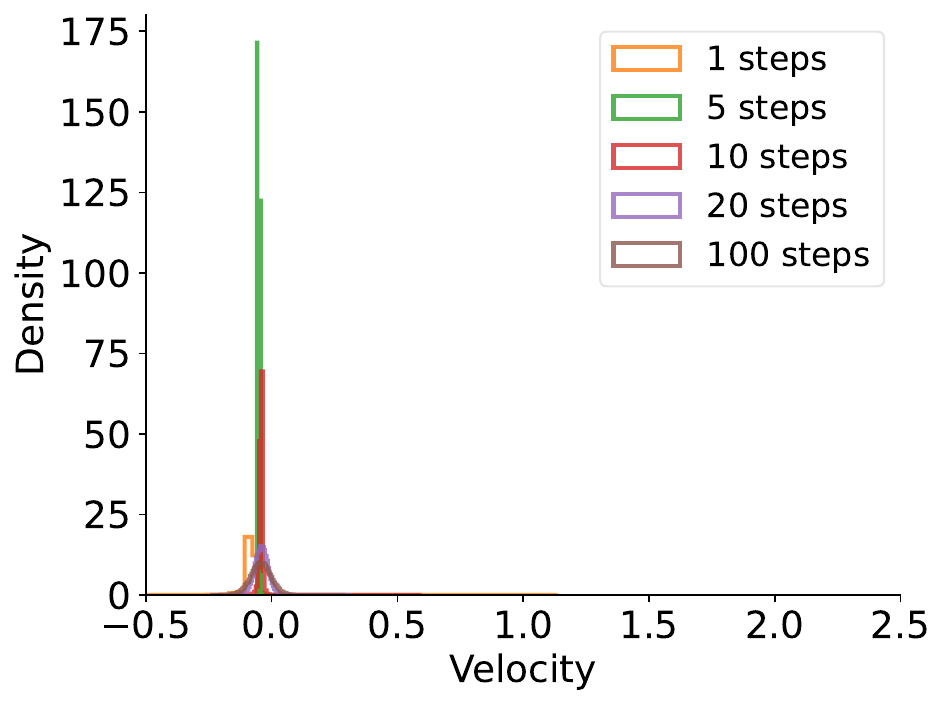}\\
    \includegraphics[width=0.25\linewidth]{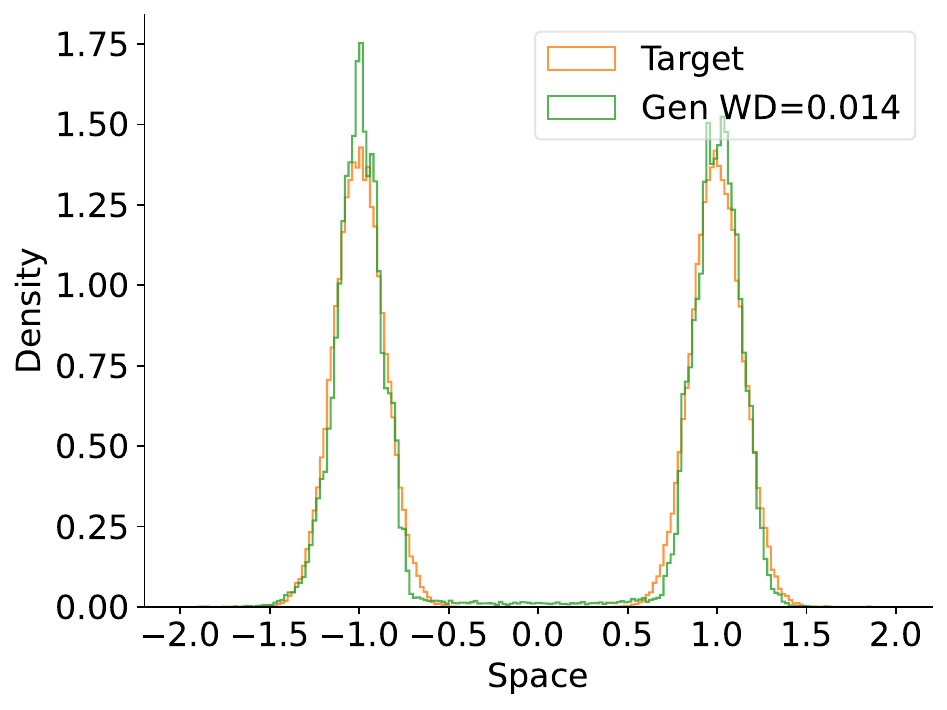}&
    \includegraphics[width=0.25\linewidth]{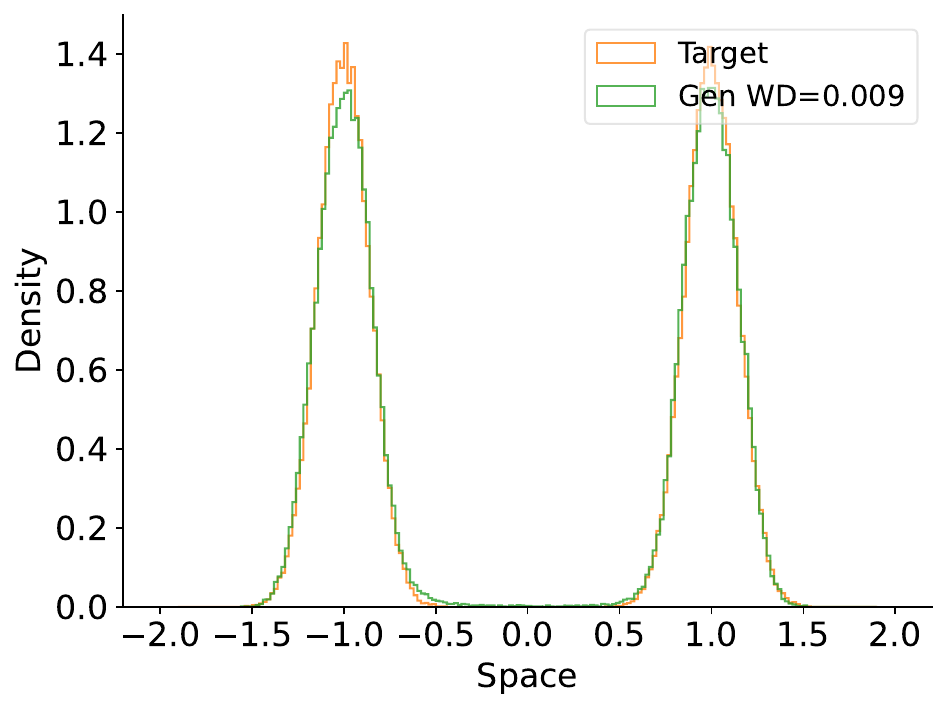}&
    \includegraphics[width=0.25\linewidth]{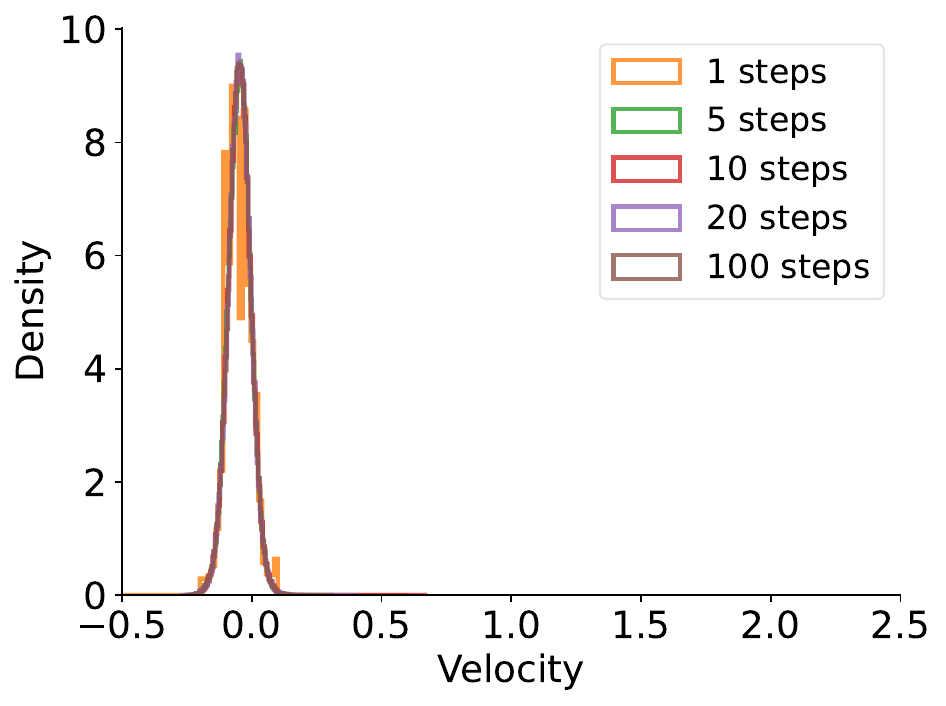}\\
    (a) Data distribution & (b) Data distribution & (c) Velocity distribution \\
    total NFEs 1 & total NFEs 100 &
    \end{tabular}}
    \caption{Results on 1D $\mathcal{N}\to2\mathcal{N}$ data. The three rows correspond to HRF2, HRF2 with data coupling, HRF2 with data \& velocity coupling. (a) and (b) are generated data distribution with total NFEs 1 and 100. (c) is velocity distribution at $(x_t,t)=(-1,0)$. }
    \label{fig:app:1to2}
    \vspace{2mm}
\end{figure*}

\begin{figure*}[t]
    \centering
    \setlength{\tabcolsep}{0pt}
    {\small
    \begin{tabular}{ccc}
    \includegraphics[width=0.25\linewidth]{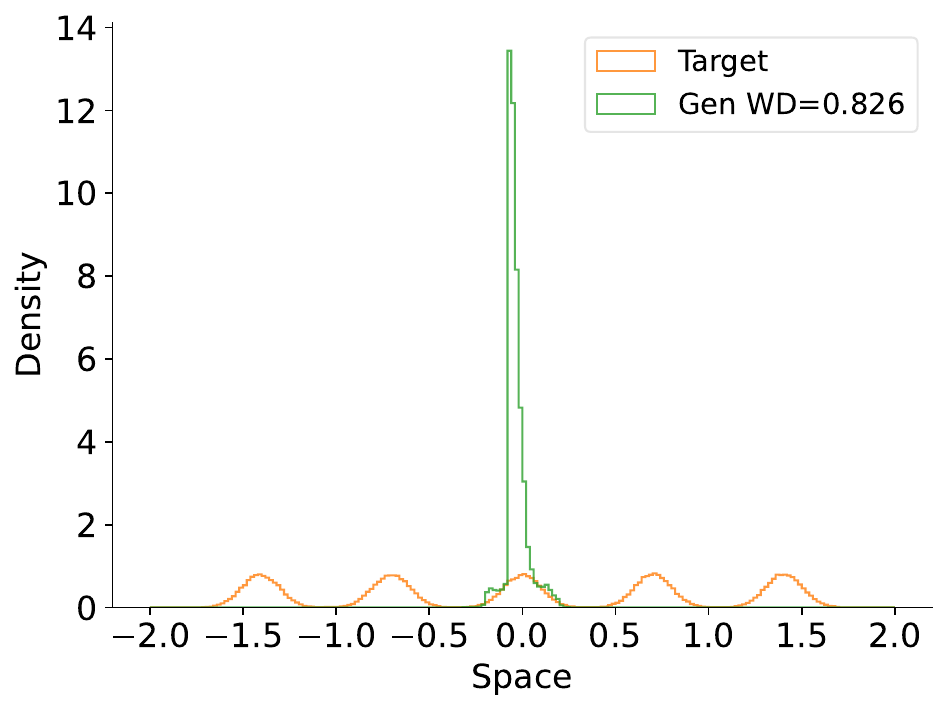}&
    \includegraphics[width=0.25\linewidth]{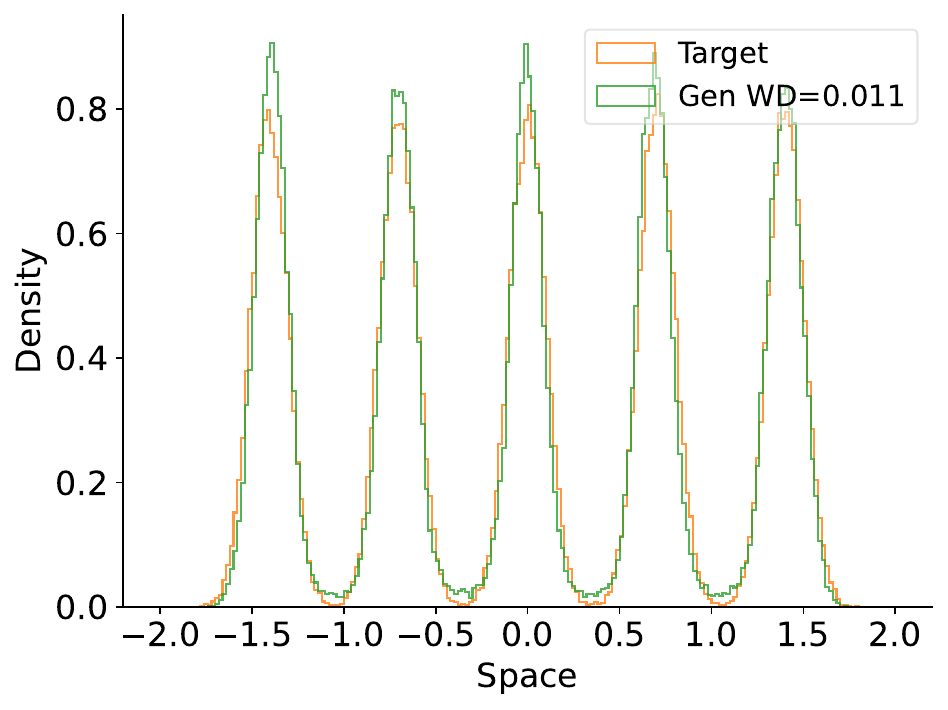}&
    \includegraphics[width=0.25\linewidth]{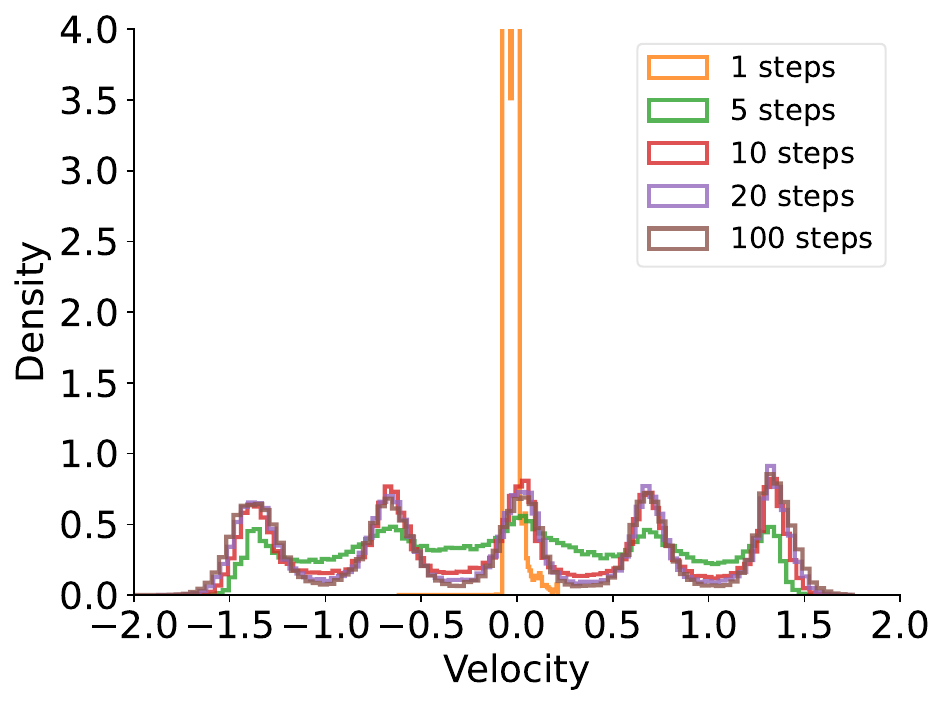}\\
    \includegraphics[width=0.25\linewidth]{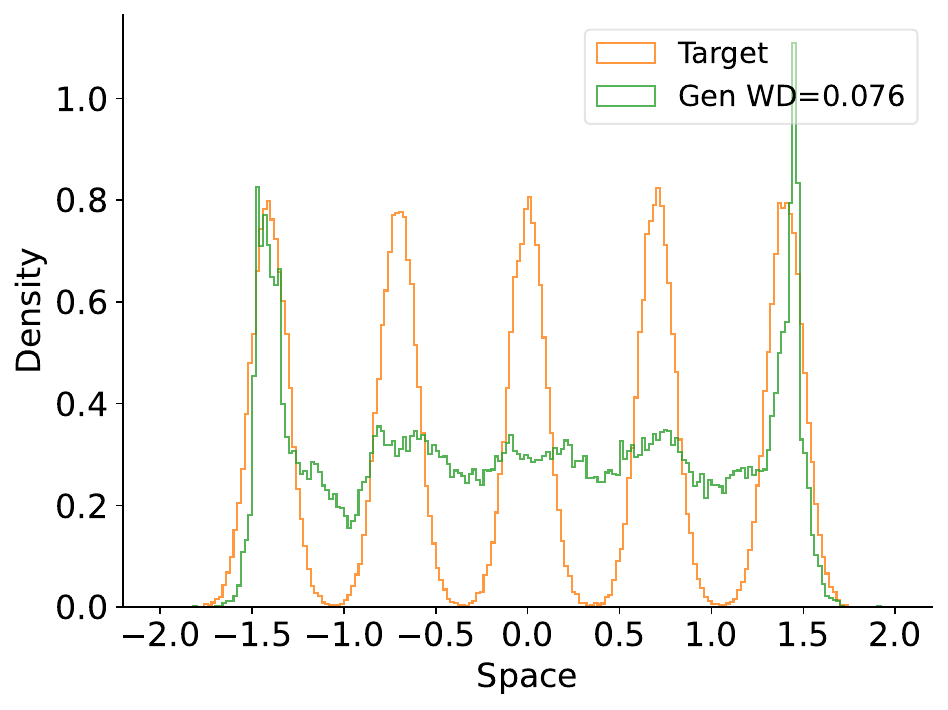}&
    \includegraphics[width=0.25\linewidth]{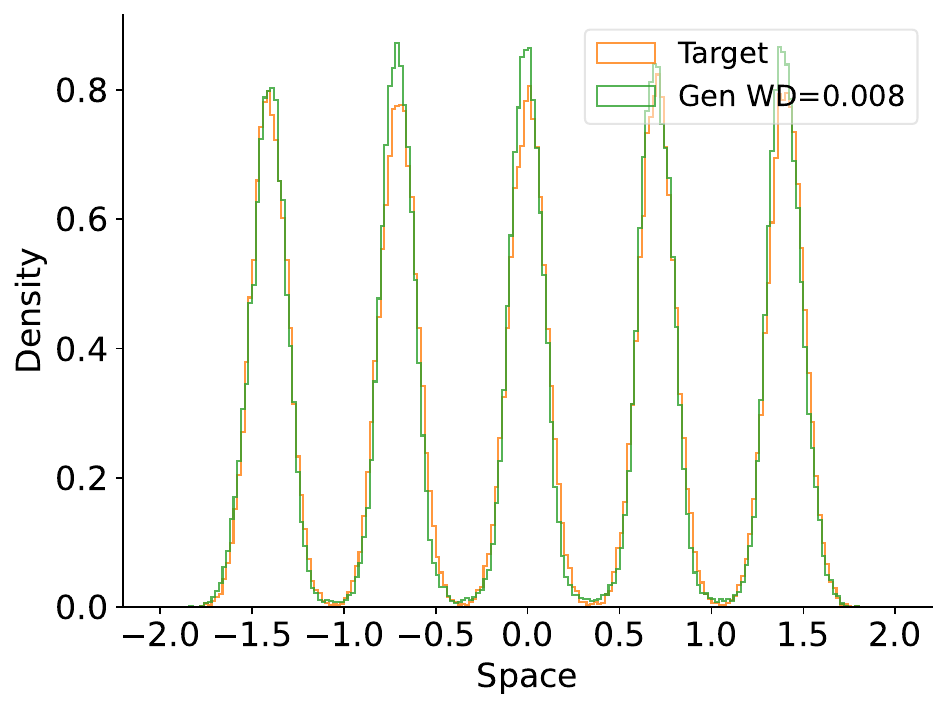}&
    \includegraphics[width=0.25\linewidth]{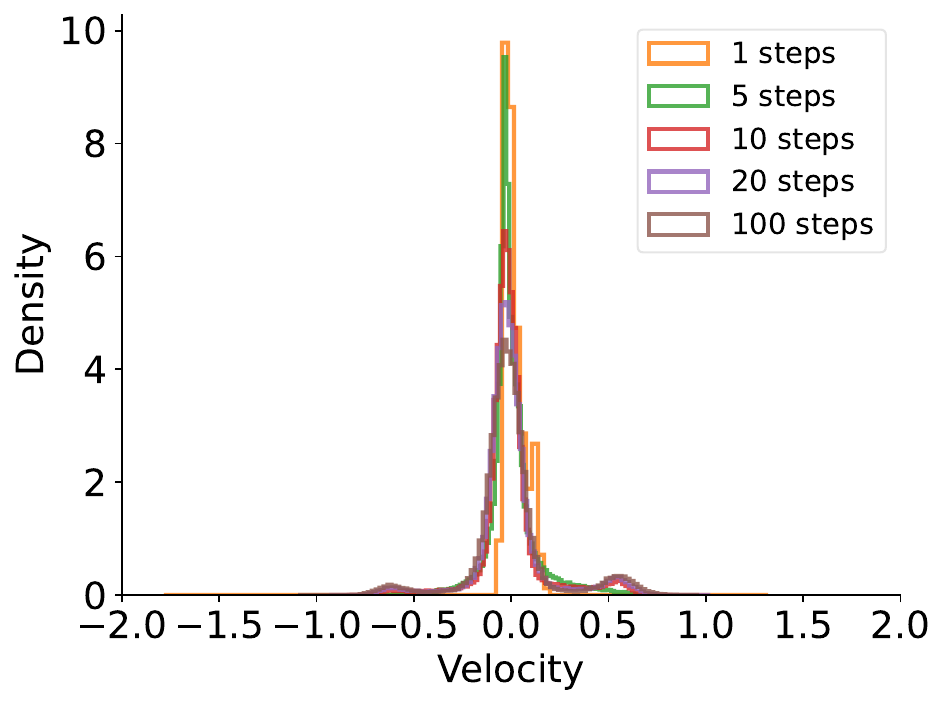}\\
    \includegraphics[width=0.25\linewidth]{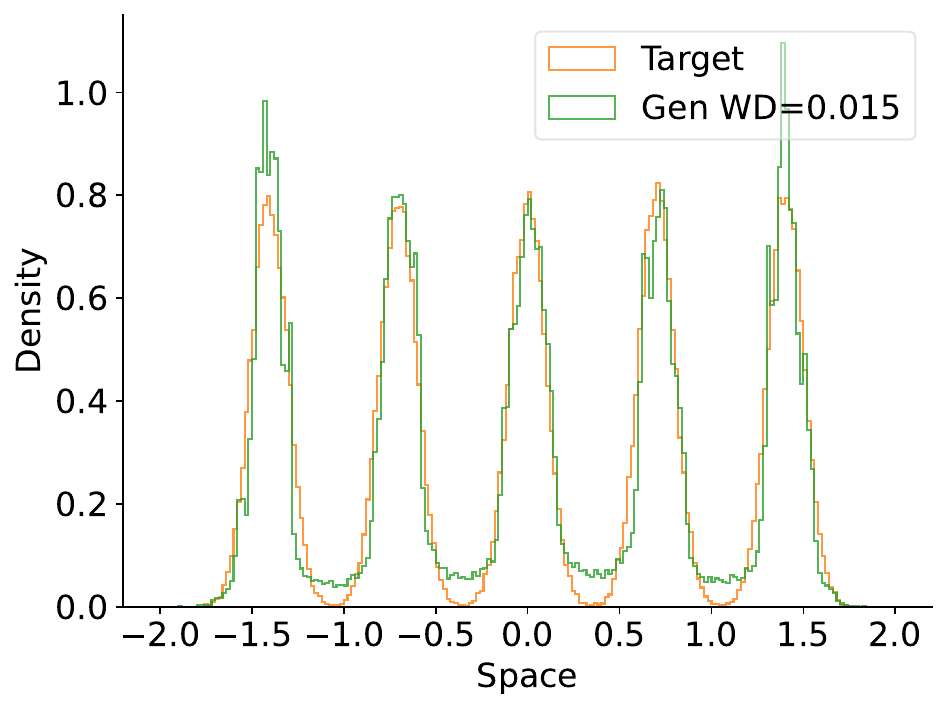}&
    \includegraphics[width=0.25\linewidth]{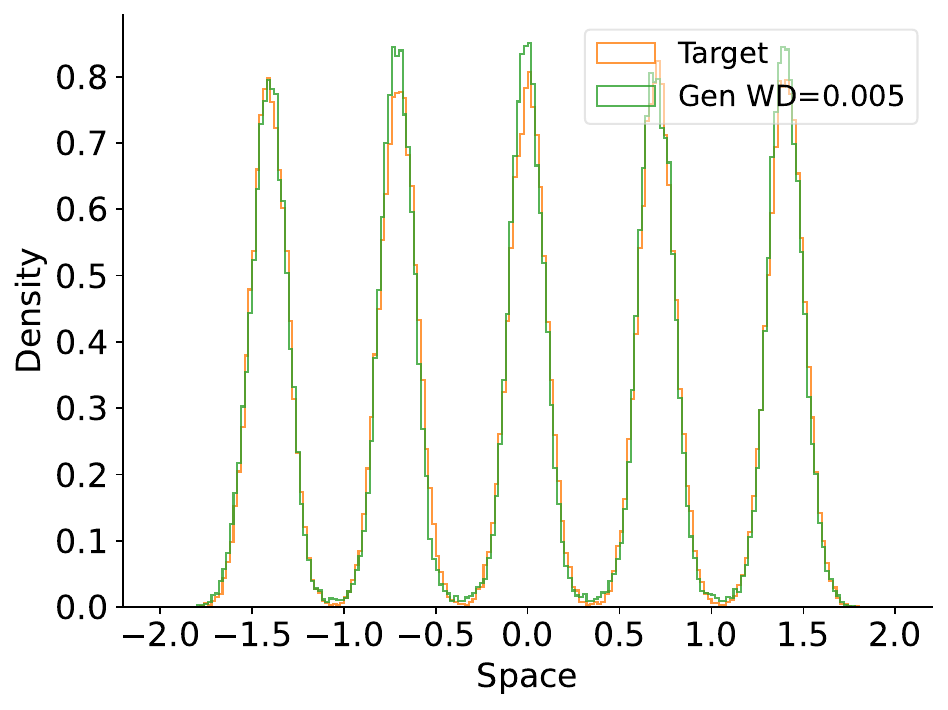}&
    \includegraphics[width=0.25\linewidth]{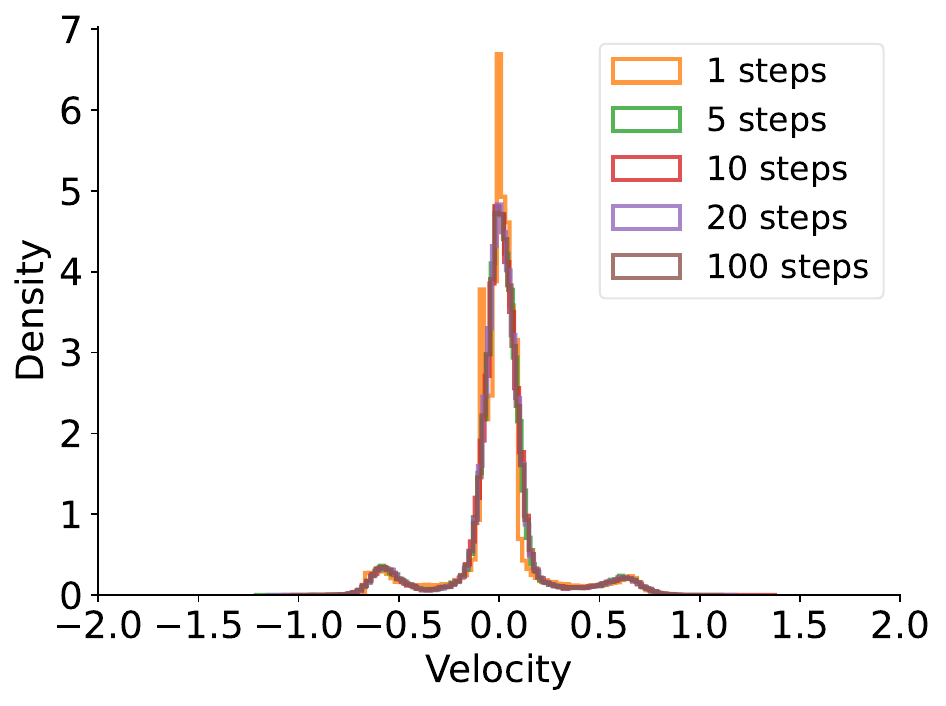}\\
    (a) Data distribution & (b) Data distribution & (c) Velocity distribution \\
    total NFEs 1 & total NFEs 100 &
    \end{tabular}}
    \caption{Results on 1D $\mathcal{N}\to5\mathcal{N}$ data. The three rows correspond to HRF2, HRF2 with data coupling, HRF2 with data \& velocity coupling. (a) and (b) are generated data distribution with total NFEs 1 and 100. (c) is velocity distribution at $(x_t,t)=(0,0)$. }
    \label{fig:app:1to5}
    \vspace{2mm}
\end{figure*}

\begin{figure*}[t]
    \centering
    \setlength{\tabcolsep}{0pt}
    {\small
    \begin{tabular}{cccc}
    \includegraphics[width=0.25\linewidth]{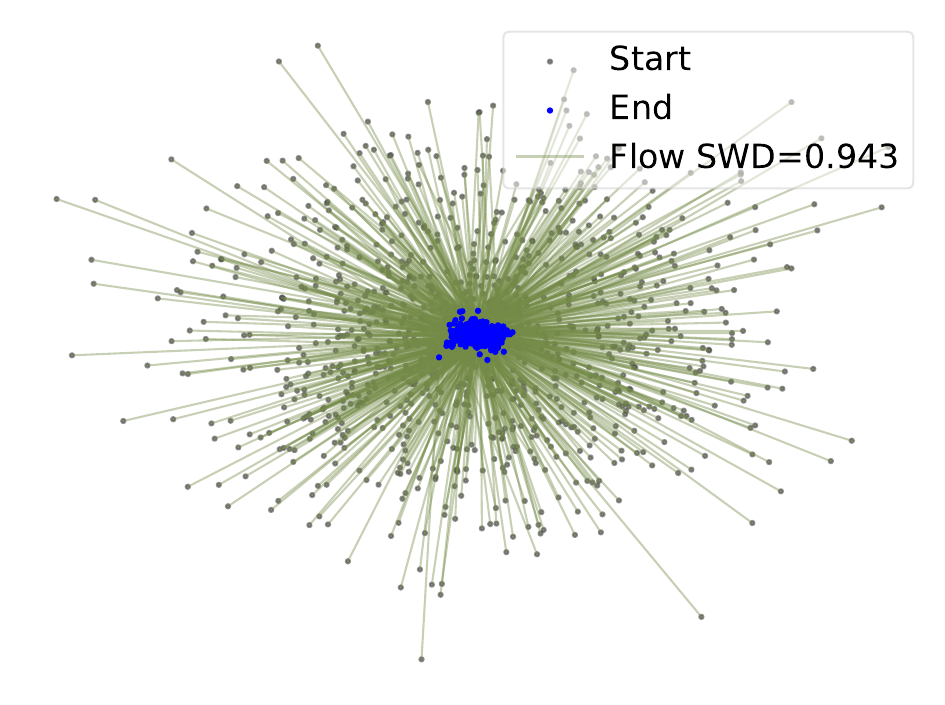}&
    \includegraphics[width=0.25\linewidth]{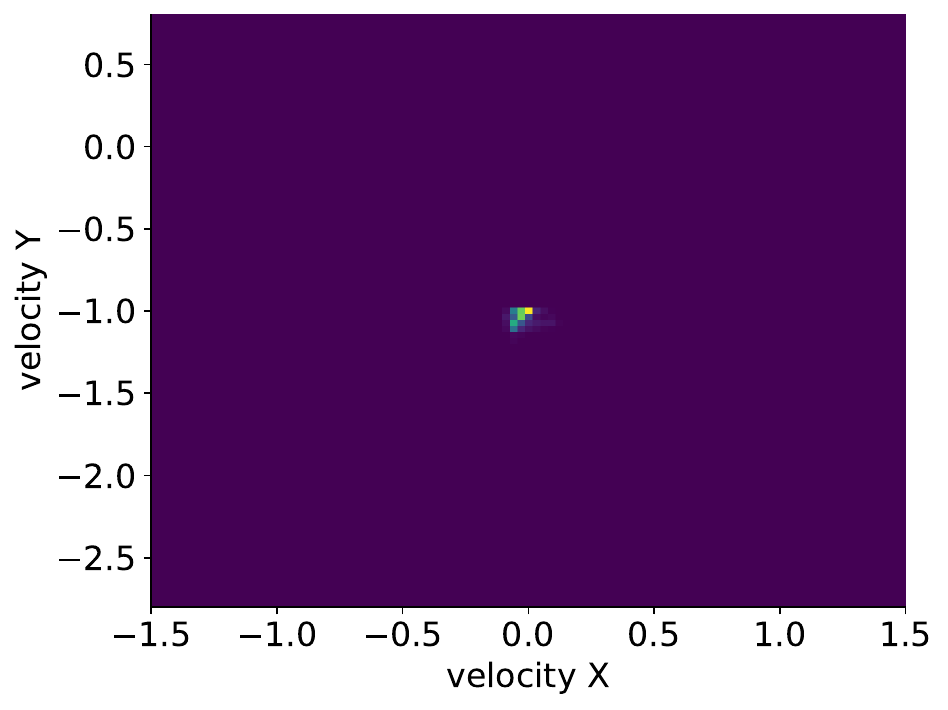}&
    \includegraphics[width=0.25\linewidth]{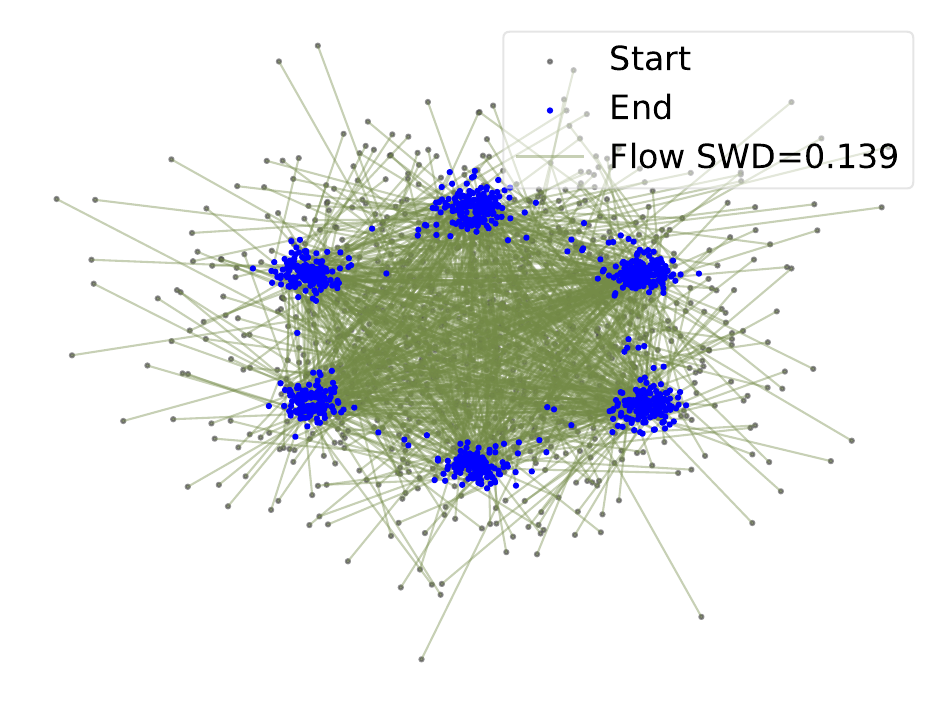}&
    \includegraphics[width=0.25\linewidth]{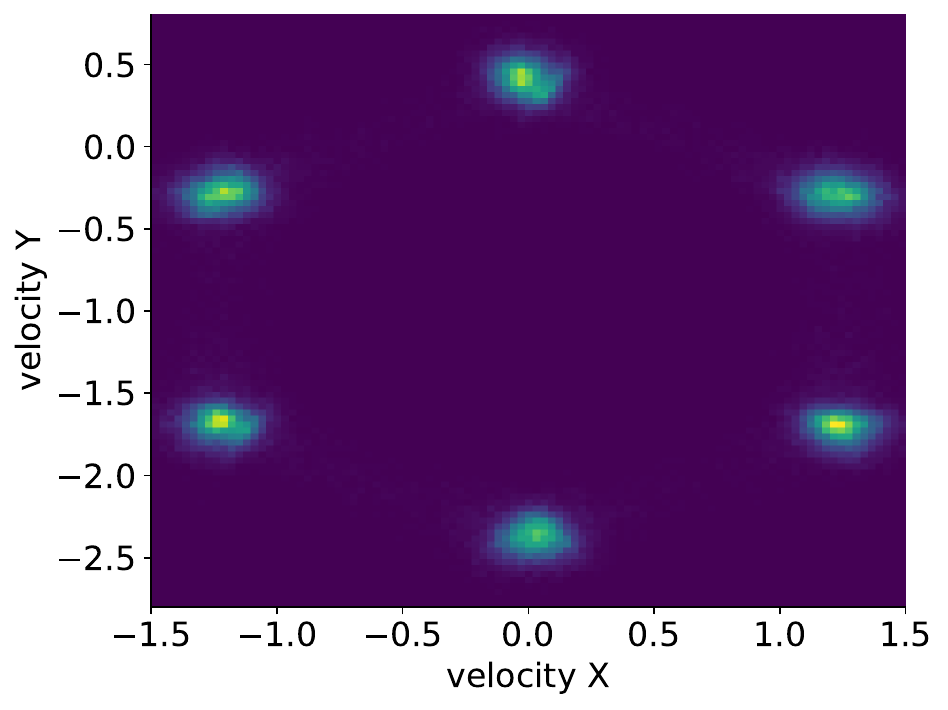}\\
    \includegraphics[width=0.25\linewidth]{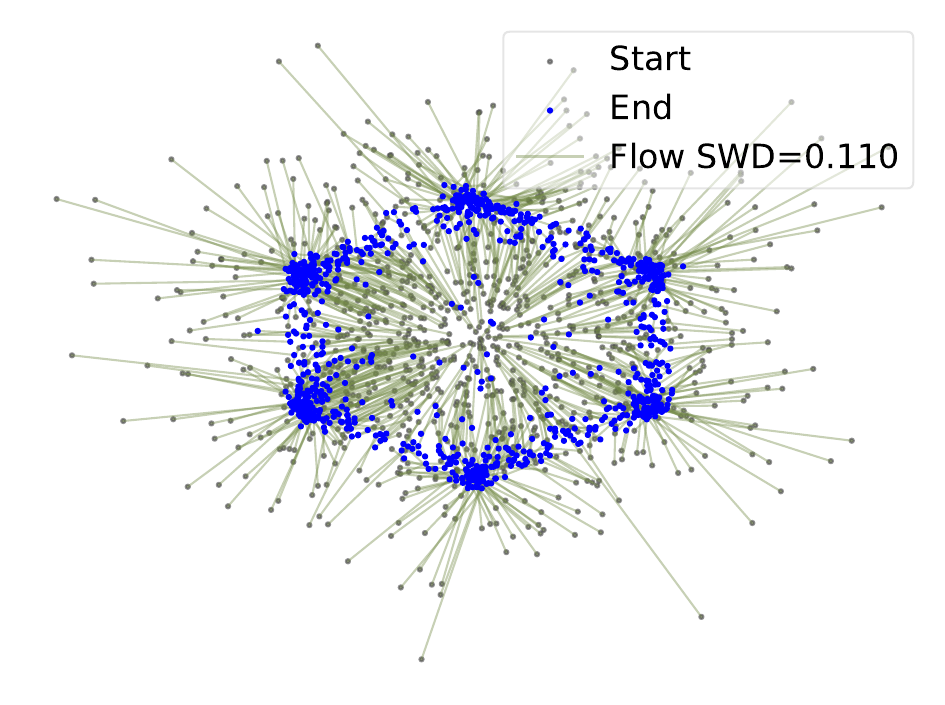}&
    \includegraphics[width=0.25\linewidth]{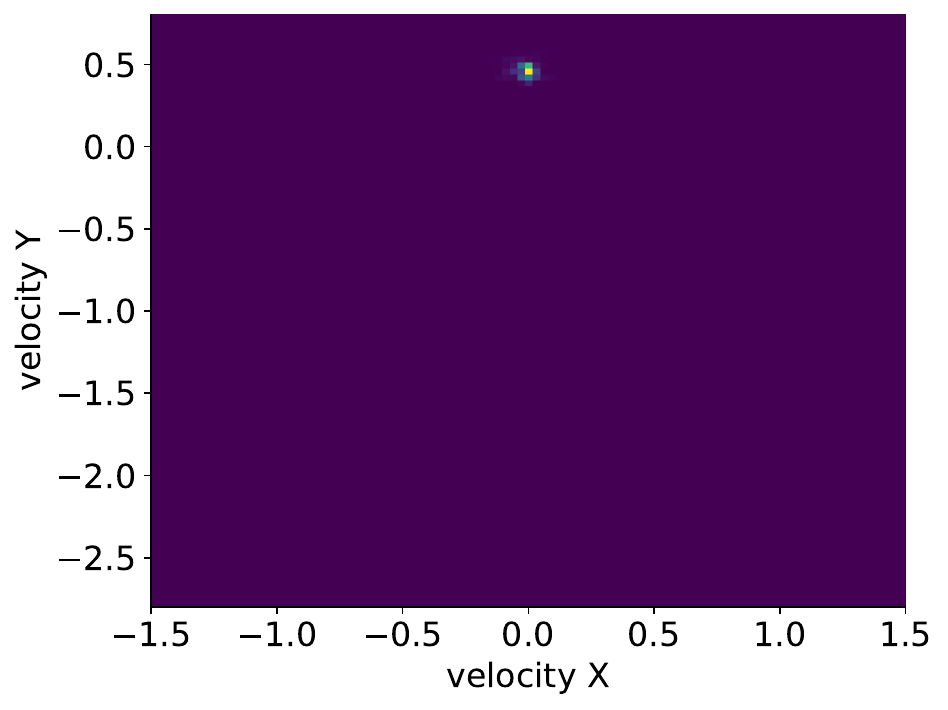}&
    \includegraphics[width=0.25\linewidth]{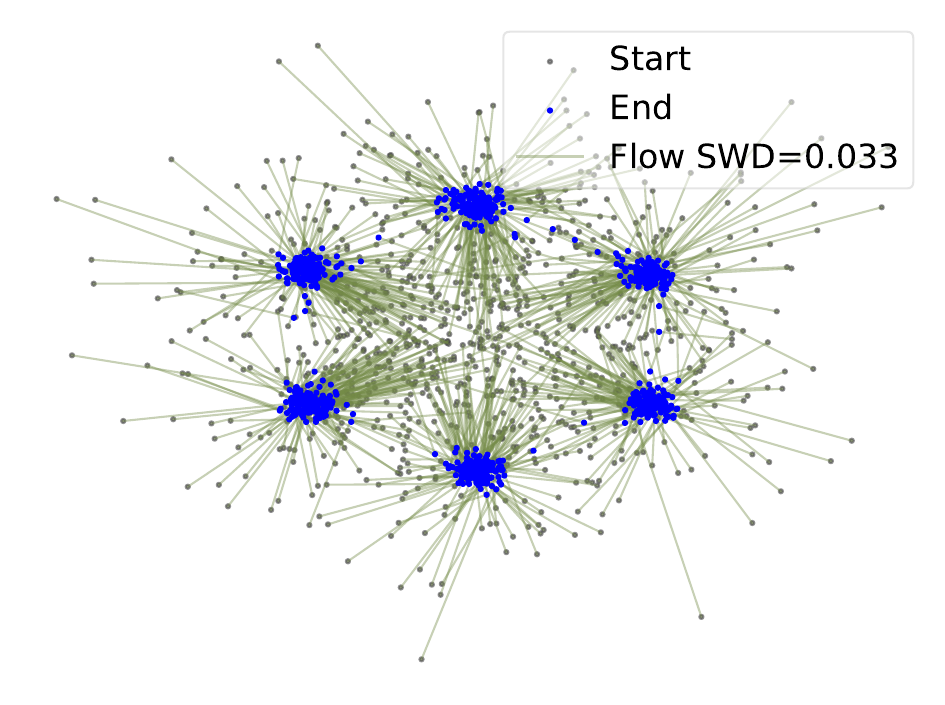}&
    \includegraphics[width=0.25\linewidth]{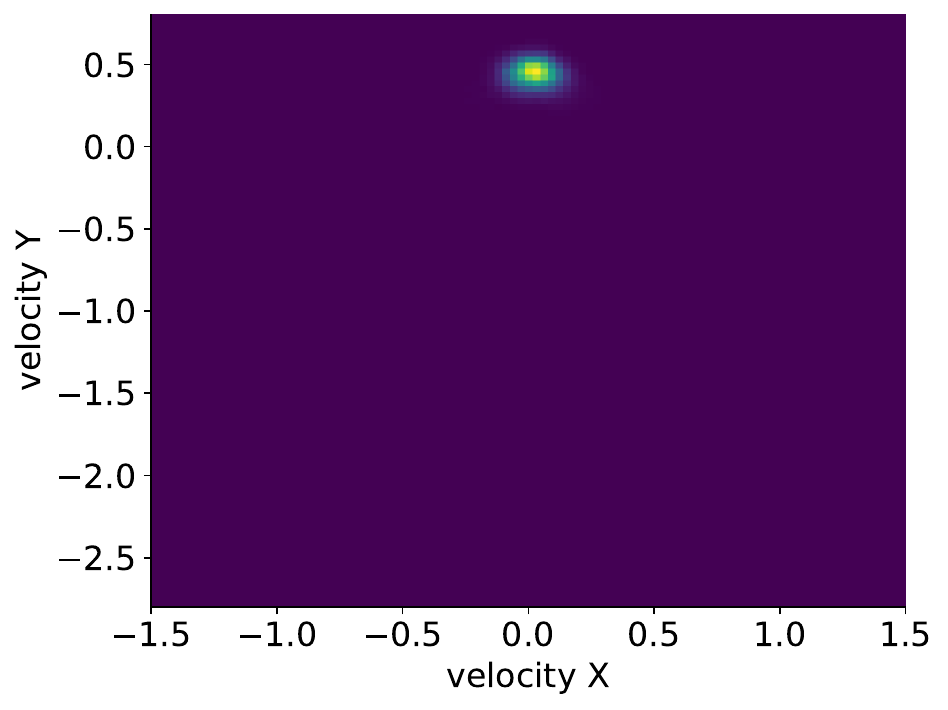}\\
    \includegraphics[width=0.25\linewidth]{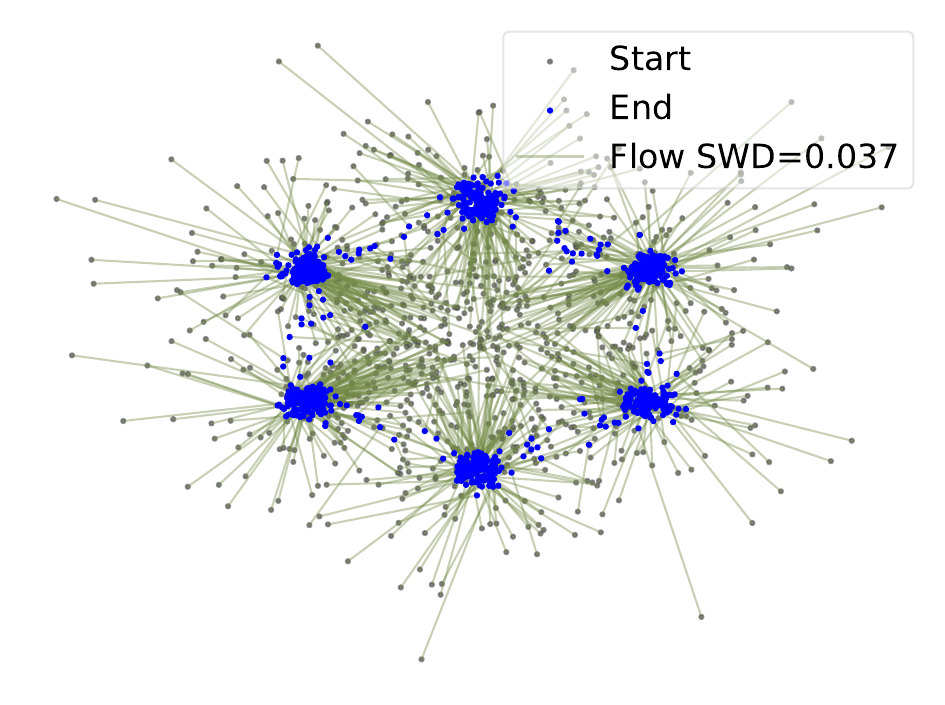}&
    \includegraphics[width=0.25\linewidth]{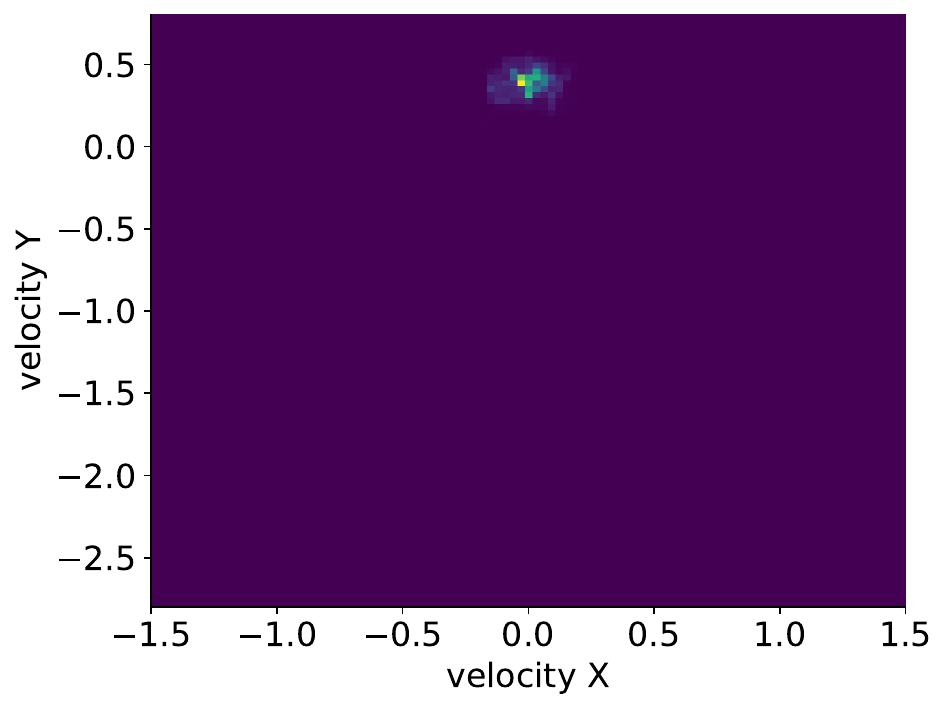}&
    \includegraphics[width=0.25\linewidth]{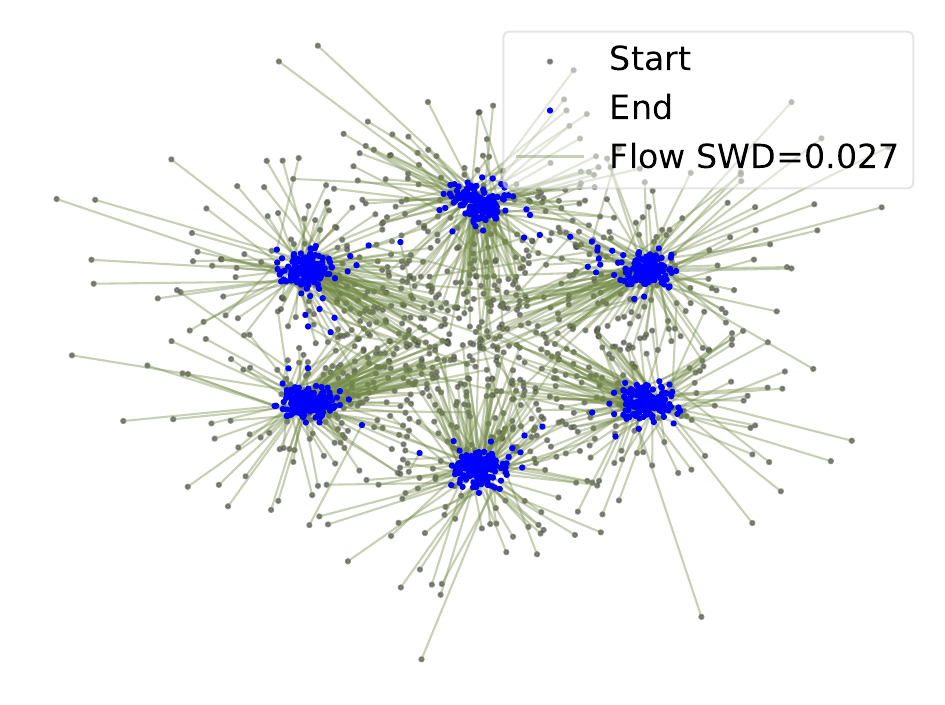}&
    \includegraphics[width=0.25\linewidth]{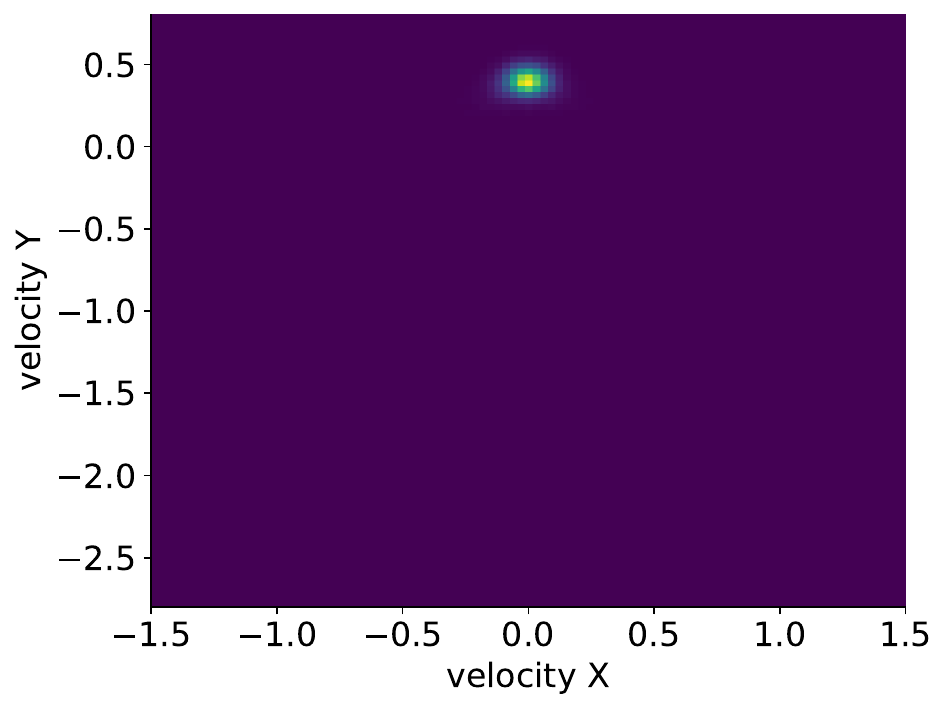}\\
    (a) Data distribution & (b) Velocity distribution & (c) Data distribution & (d) Velocity distribution \\
    total NFEs 1 & 1 step integration & total NFEs 100 & 100 step integration
    \end{tabular}}
    \caption{Results on 2D $\mathcal{N}\to6\mathcal{N}$ data. The three rows correspond to HRF2, HRF2 with data coupling, HRF2 with data \& velocity coupling. (a) and (c) are trajectories (green) of sample particles flowing from source distribution (grey) to target distribution (blue) with total NFEs 1 and 100. (b) and (d) are velocity distributions at $(0,1)$ at $t=0$. }
    \label{fig:app:2D1to6}
    \vspace{2mm}
\end{figure*}

We present more results on synthetic data: \cref{fig:app:1to2} for 1D $\mathcal{N}\to2\mathcal{N}$ data, \cref{fig:app:1to5} for 1D $\mathcal{N}\to5\mathcal{N}$ data, and \cref{fig:app:2D1to6} for 2D $\mathcal{N}\to6\mathcal{N}$ data. Across all these experiments, we consistently observe that data coupling simplifies the velocity distribution, while velocity coupling significantly reduces the required sampling steps.

\subsection{Image Data Results}
\label{app:res:img}

\begin{figure*}[t]
    \centering
    \setlength{\tabcolsep}{0pt}
    {\small
    \begin{tabular}{ccccc}
    \includegraphics[width=0.2\linewidth]{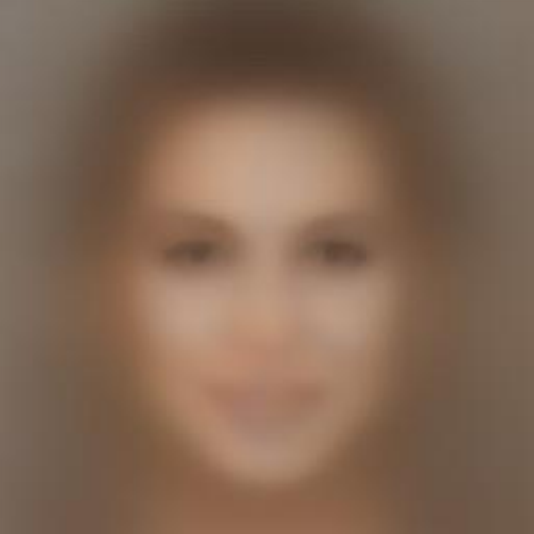}&
    \includegraphics[width=0.2\linewidth]{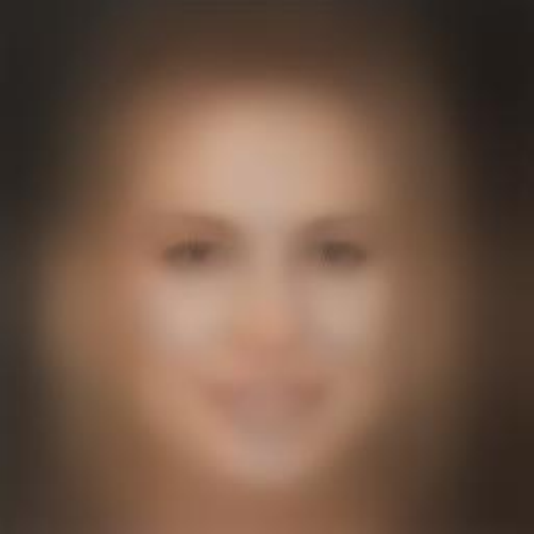}&
    \includegraphics[width=0.2\linewidth]{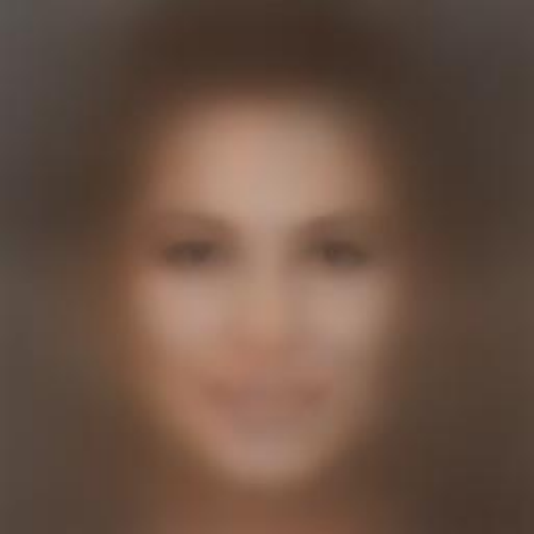}&
    \includegraphics[width=0.2\linewidth]{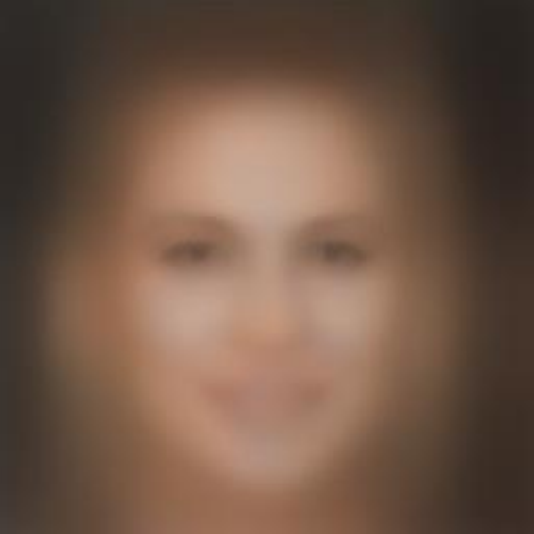}&
    \includegraphics[width=0.2\linewidth]{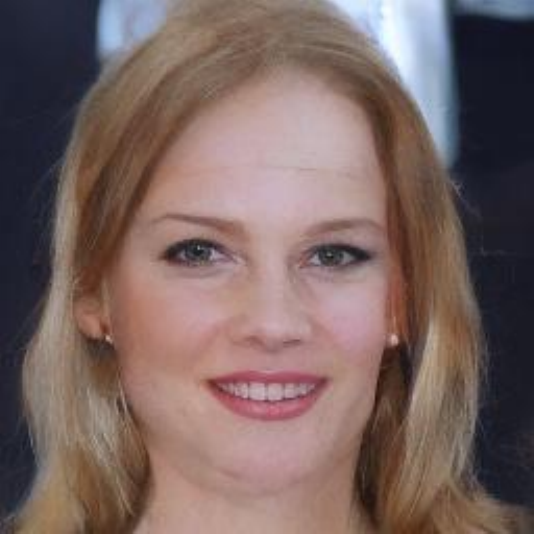}
    \\[-1mm]
    \includegraphics[width=0.2\linewidth]{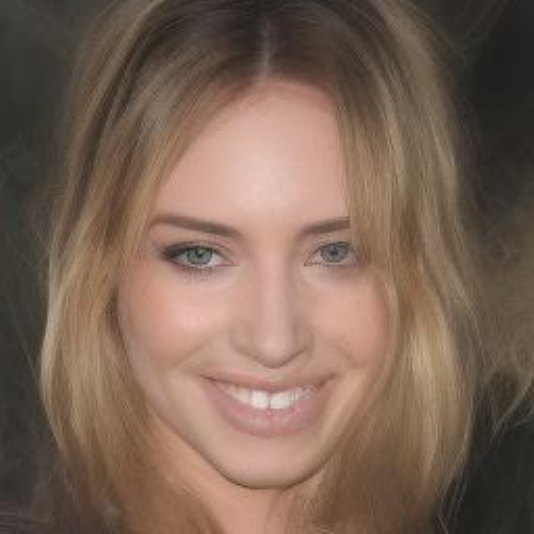}&
    \includegraphics[width=0.2\linewidth]{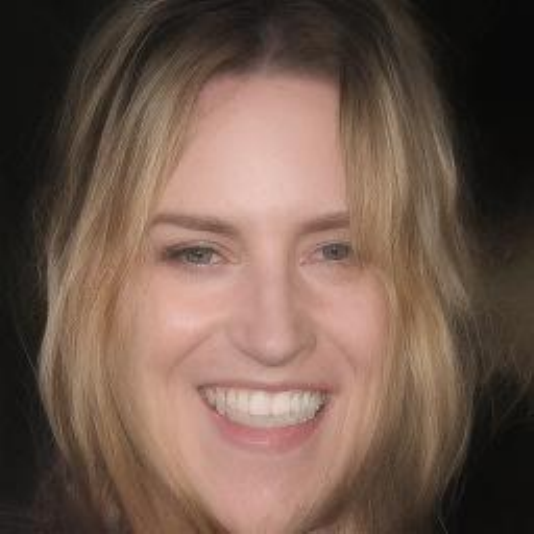}&
    \includegraphics[width=0.2\linewidth]{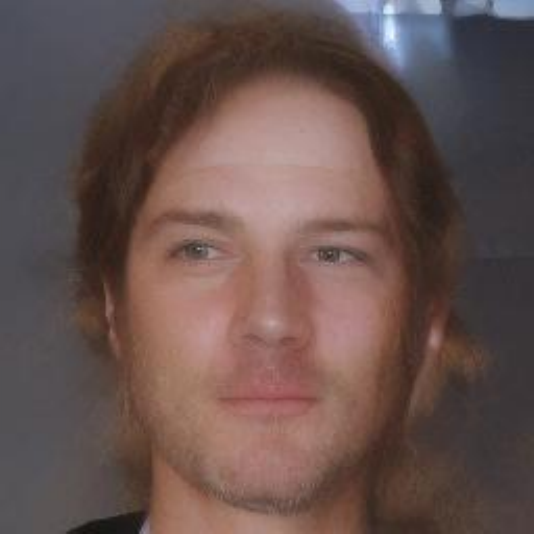}&
    \includegraphics[width=0.2\linewidth]{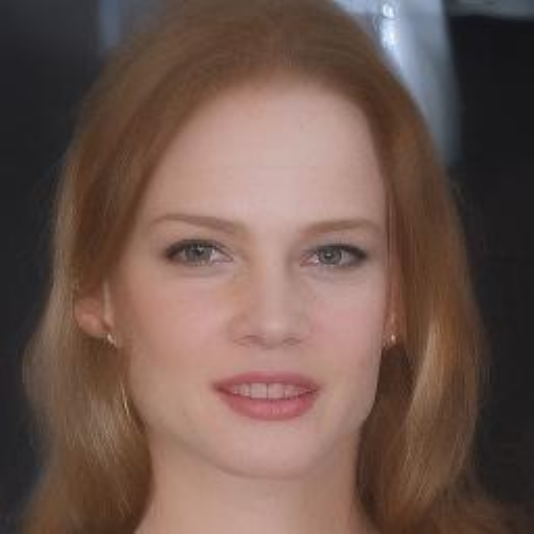}&
    \includegraphics[width=0.2\linewidth]{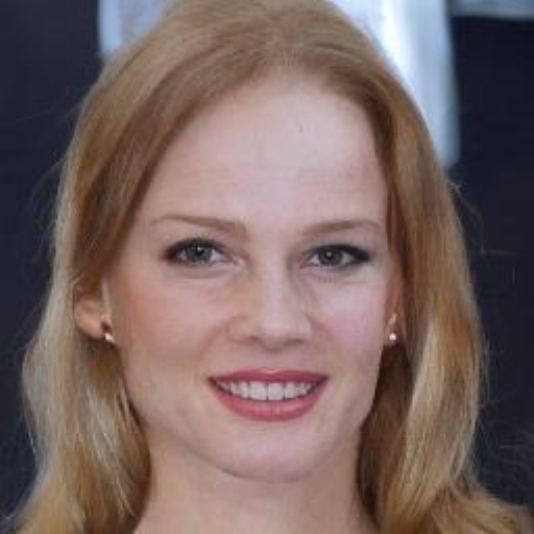}
    \\[-1mm]
    \includegraphics[width=0.2\linewidth]{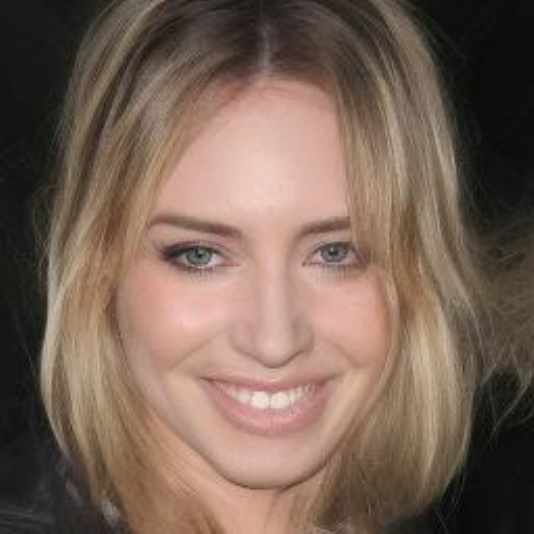}&
    \includegraphics[width=0.2\linewidth]{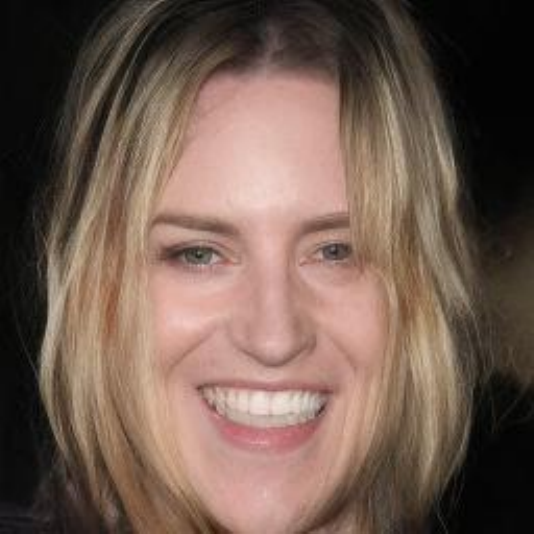}&
    \includegraphics[width=0.2\linewidth]{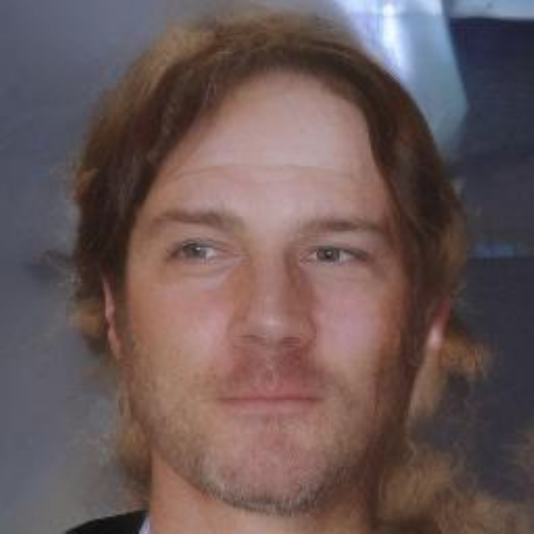}&
    \includegraphics[width=0.2\linewidth]{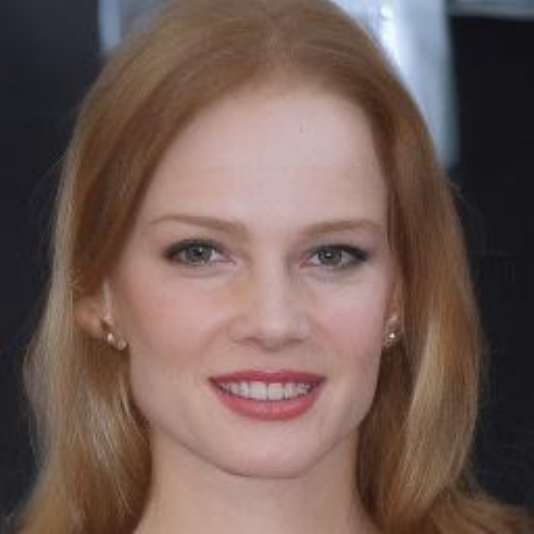}&
    \includegraphics[width=0.2\linewidth]{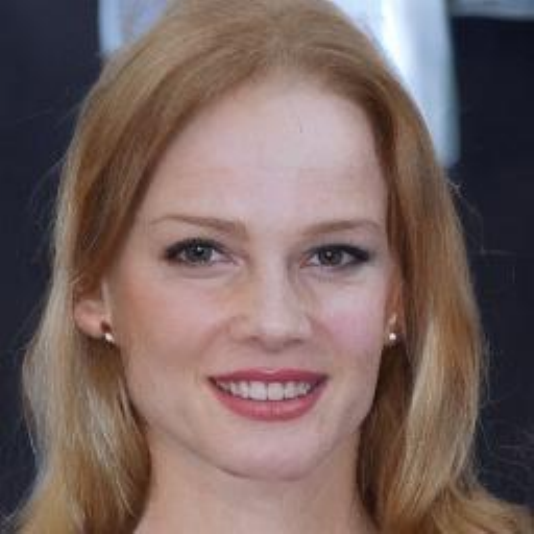}
    \\[-1mm]
    \includegraphics[width=0.2\linewidth]{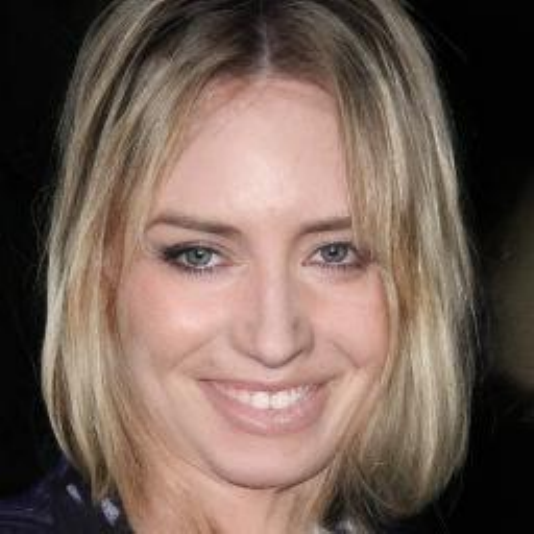}&
    \includegraphics[width=0.2\linewidth]{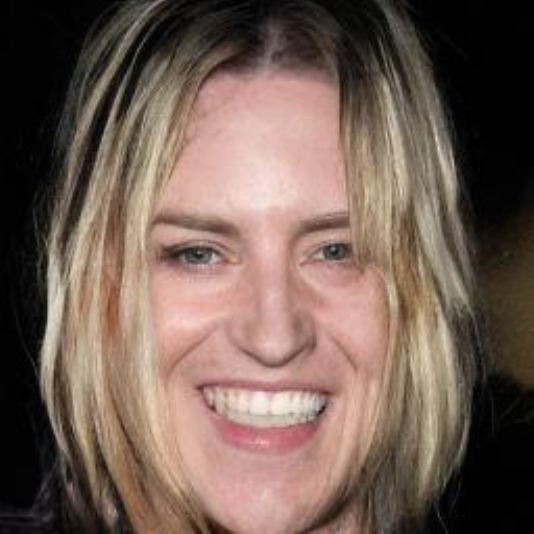}&
    \includegraphics[width=0.2\linewidth]{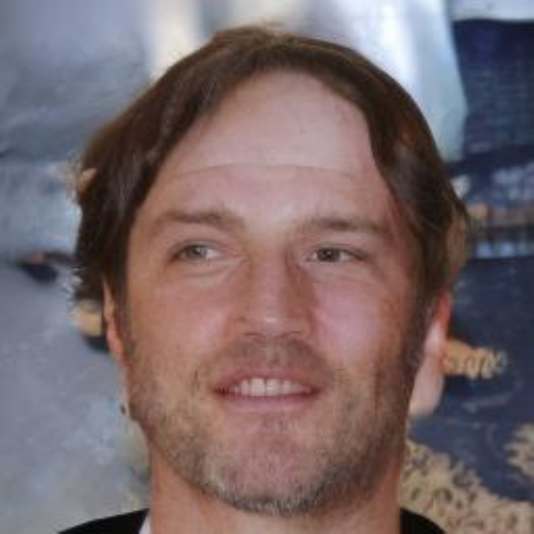}&
    \includegraphics[width=0.2\linewidth]{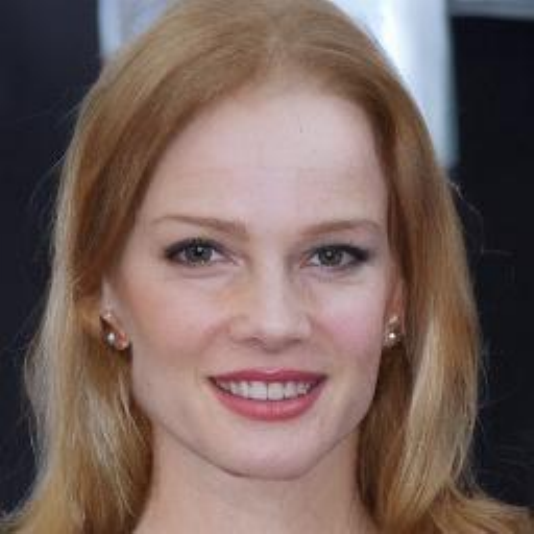}&
    \includegraphics[width=0.2\linewidth]{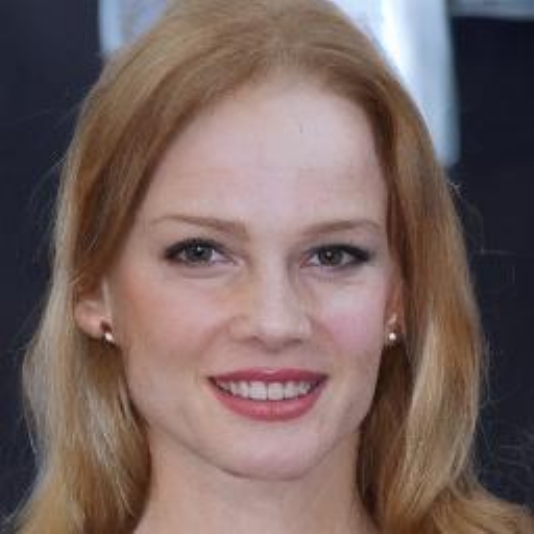}
    \\[-1mm]
    \includegraphics[width=0.2\linewidth]{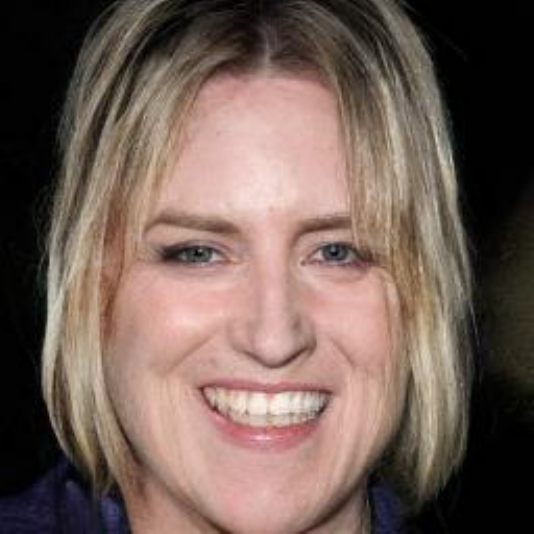}&
    \includegraphics[width=0.2\linewidth]{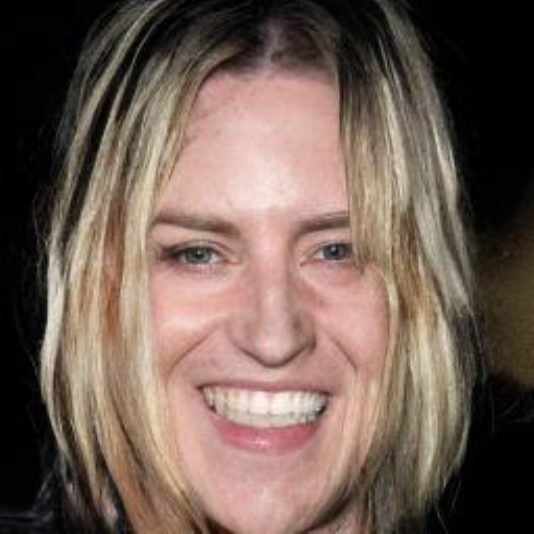}&
    \includegraphics[width=0.2\linewidth]{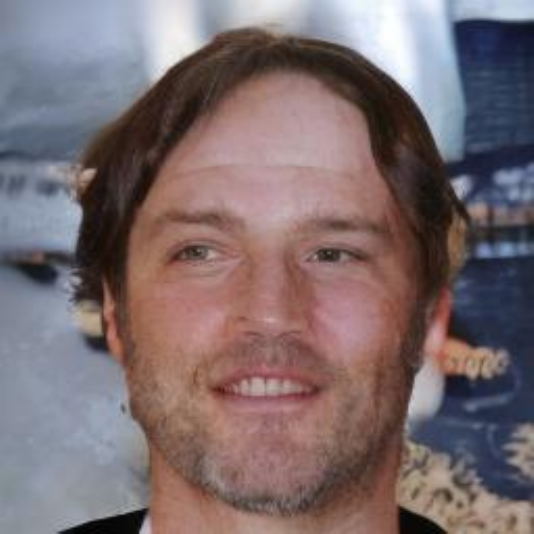}&
    \includegraphics[width=0.2\linewidth]{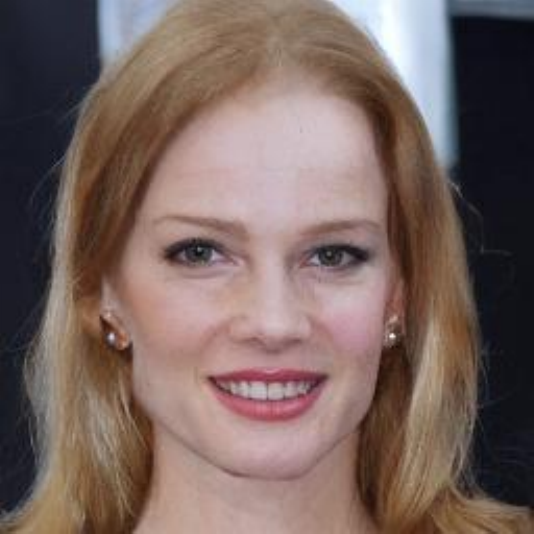}&
    \includegraphics[width=0.2\linewidth]{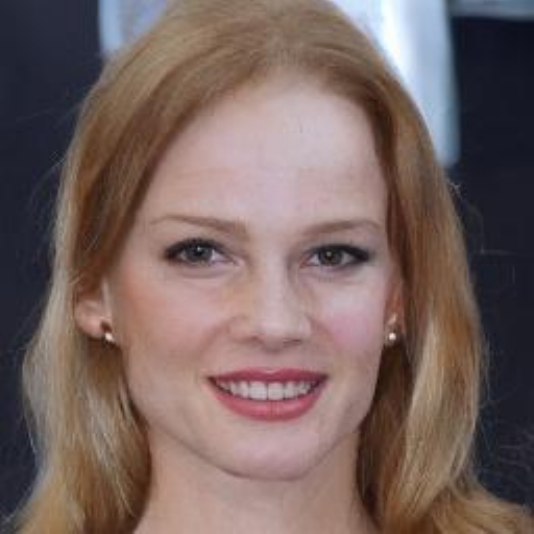}
    \\
    (a) RF & (b) OT-CFM & (c) HRF2 & (d) HRF2-D & (e) HRF2-D\&V \\
    \end{tabular}}
    \caption{Examples of the generated images for CelebA-HQ starting from the same noise for all models. The 5 rows from top to bottom correspond to total NFEs = 1, 5, 10, 50, 500. }
    \label{fig:app:nfe:image}
\end{figure*}

We present more results on image data: \cref{tab:fid:mnist} for MNIST, \cref{tab:fid:cifar10} for CIFAR-10, and \cref{tab:fid:celeba,fig:app:nfe:image} for CelebA-HQ 256. Again, we consistently observe that data coupling enhances sampling quality for both low and high total NFEs, but collapses when total NFE is reduced to 1, while velocity coupling produces high-quality samples even under this extreme case. 

\begin{table}[t]
\centering
\resizebox{1.0\columnwidth}{!}{
\begin{tabular}{cccccc}
\toprule
\textbf{Total NFEs} & \textbf{RF (1.08M)} & \textbf{OT-CFM  (1.08M)} & \textbf{HRF2 (1.07M)} & \textbf{HRF2-D (1.07M)} & \textbf{HRF2-D\&V (1.07M)} \\
\midrule
5   & 19.187 ± 0.188 & 13.977 ± 0.166 & 15.798 ± 0.151 
    & 10.167 ± 0.136 & \textbf{5.519 ± 0.112} \\
10  & 7.974 ± 0.119  & 4.477 ± 0.099  & 6.644 ± 0.076  
    & \textbf{3.823 ± 0.038}  & 3.861 ± 0.089 \\
20  & 6.151 ± 0.090  & 2.763 ± 0.036  & 3.408 ± 0.076  
    & \textbf{2.318 ± 0.053}  & 3.720 ± 0.045 \\
50  & 5.605 ± 0.057  & 2.321 ± 0.038  & 2.664 ± 0.058  
    & \textbf{1.929 ± 0.031}  & 3.604 ± 0.016 \\
100 & 5.563 ± 0.049  & 2.346 ± 0.023  & 2.588 ± 0.075  
    & \textbf{1.847 ± 0.011}  & 3.423 ± 0.003 \\
500 & 5.453 ± 0.047  & 2.296 ± 0.007  & 2.574 ± 0.121  
    & \textbf{1.913 ± 0.043}  & 3.546 ± 0.107 \\
\bottomrule
\end{tabular}}
\vspace{5pt}
\caption{FID performance on MNIST under different total NFE settings. \textbf{Bold} for the best. }
\label{tab:fid:mnist}
\end{table}

\begin{table}[t]
\centering
\resizebox{1.0\columnwidth}{!}{
\begin{tabular}{cccccc}
\toprule
\textbf{Total NFEs} & \textbf{RF (35.75M)} & \textbf{OT-CFM  (35.75M)} & \textbf{HRF2 (44.81M)} & \textbf{HRF2-D (44.81M)} & \textbf{HRF2-D\&V (44.81M)} \\
\midrule
5   & 36.209 ± 0.142 & 23.111 ± 0.010 & 30.884 ± 0.104 
    & 22.817 ± 0.072 & \textbf{6.315 ± 0.057} \\
10  & 14.113 ± 0.092 & 12.564 ± 0.016 & 12.065 ± 0.024 
    & 10.969 ± 0.025 & \textbf{5.739 ± 0.017} \\
20  & 8.355 ± 0.065  & 8.553 ± 0.002  & 7.129 ± 0.027  
    & 6.860 ± 0.022  & \textbf{5.332 ± 0.009} \\
50  & 5.514 ± 0.034  & 5.911 ± 0.005  & 4.847 ± 0.028  
    & \textbf{4.739 ± 0.006}  & 5.142 ± 0.024 \\
100 & 4.588 ± 0.013  & 4.952 ± 0.012  & 4.334 ± 0.054  
    & \textbf{4.301 ± 0.022}  & 5.078 ± 0.044 \\
500 & 3.887 ± 0.035  & 4.184 ± 0.086  & 3.706 ± 0.043  
    & \textbf{3.578 ± 0.028}  & 5.095 ± 0.032 \\
\bottomrule
\end{tabular}}
\vspace{5pt}
\caption{FID performance on CIFAR-10 under different total NFE settings. \textbf{Bold} for the best. }
\label{tab:fid:cifar10}
\end{table}

\begin{table}[t]
\centering
\resizebox{1.0\columnwidth}{!}{
\begin{tabular}{cccccc}
\toprule
\textbf{Total NFEs} & \textbf{RF (457.06M)} & \textbf{OT-CFM  (457.06M)} & \textbf{HRF2 (616.20M)} & \textbf{HRF2-D (616.20M)} & \textbf{HRF2-D\&V (616.20M)} \\
\midrule
5   & 38.641 ± 0.126 & 29.646 ± 0.093 & 34.246 ± 0.107 
    & 32.918 ± 0.085 & \textbf{7.612 ± 0.015} \\
10  & 16.876 ± 0.088 & 12.879 ± 0.083 & 15.391 ± 0.074 
    & 13.424 ± 0.022 & \textbf{6.931 ± 0.038} \\
20  & 10.027 ± 0.060 & 7.426 ± 0.042  & 9.291 ± 0.042  
    & 7.048 ± 0.033  & \textbf{6.560 ± 0.039} \\
50  & 7.395 ± 0.021  & 5.545 ± 0.023  & 6.927 ± 0.041  
    & \textbf{5.529 ± 0.021}  & 6.330 ± 0.022 \\
100 & 6.850 ± 0.064  & 5.236 ± 0.034  & 6.450 ± 0.062  
    & \textbf{4.961 ± 0.019}  & 6.248 ± 0.023 \\
500 & 6.418 ± 0.026  & 5.094 ± 0.019  & 6.188 ± 0.056  
    & \textbf{4.624 ± 0.029}  & 6.225 ± 0.015 \\
\bottomrule
\end{tabular}}
\vspace{5pt}
\caption{FID performance on CelebA-HQ 256 under different total NFE settings. \textbf{Bold} for the best. }
\label{tab:fid:celeba}
\end{table}

%% file: 12_appendix_imp.tex
\section{Implementation Details}
\label{app:imp}

\subsection{Synthetic Data}
\label{app:imp:syn}
For synthetic data experiments, we employ a neural network architecture with two distinct stages. The first stage separately encodes spatial and temporal inputs with linear layers and Sinusoidal Positional Embeddings. The second stage concatenates the processed features and refines them through multiple linear layers to produce the final output. The model consists of 304,513 parameters, totaling 0.30M in size.

For data coupling, we train the model from scratch following \cref{alg:data_coup} strictly. A key consideration is the choice of batch sizes, as two different batch sizes are involved -- one for batch OT and another for training. In 1D and 2D experiments, a large batch size is necessary for stable training, but using an excessively large batch size for batch OT is computationally inefficient. To address this, we set the batch size for batch OT to 100 while using a batch size of 1,000 for gradient computation. This means that in each training iteration, we perform batch OT on 100 data points 10 times to accumulate a full batch for gradient updates.

For velocity coupling, we use the HRF2-D model from the previous step as the base model to generate $(v_0, v_1)$ pairs at a fixed space-time location $(x_t, t)$, following \cref{alg:hrf_v_coup}. During training, we observed that the performance depends on the quality of the base model. To mitigate this, we save multiple checkpoints of HRF2-D and select the best-performing checkpoint via a validation dataset as the base model for velocity coupling.

Computational requirements during training are shown in \cref{tab:compute:lowdim}. In the low-dimensional setting, batch OT becomes more time-consuming than the training itself. As a result, HRF2-D trains significantly slower than HRF2. In contrast, HRF2-D\&V uses precomputed velocity pairs and therefore does not require batch OT during training. Moreover, it operates with a smaller batch size (1000) than HRF2 (5000), resulting in lower memory usage and faster training. 

During the evaluation, we select the best checkpoint from a validation set for all models (HRF2, HRF2-D, HRF2-D\&V). For each seed, we conduct the experiment three times, yielding three best models per seed. Each model is then evaluated three times, resulting in nine experimental results per seed. Finally, we report the mean and standard deviation over three different seeds, totaling 27 experimental results. 

\begin{table}[t]
\centering
\resizebox{1.0\columnwidth}{!}{
\begin{tabular}{ccccccc}
\toprule
\textbf{Training} 
& \multicolumn{3}{c}{\textbf{1D data}} 
& \multicolumn{3}{c}{\textbf{2D data}} \\
\cmidrule(r){2-4} \cmidrule(r){5-7}
& \textbf{HRF2 (0.30M)} & \textbf{HRF2-D (0.30M)} & \textbf{HRF2-D\&V (0.30M)} & \textbf{HRF2 (0.32M)} & \textbf{HRF2-D (0.32M)} & \textbf{HRF2-D\&V (0.32M)} \\
\midrule
Time (s/iter) & 0.0028 & 0.0581 & 0.0025 & 0.0029 & 0.0588 & 0.0027 \\
Memory (MB) & 658 & 658 & 566 & 660 & 660 & 568 \\
Param.\ Counts & 304,513 & 304,513 & 304,513 & 321,154 & 321,154 & 321,154 \\
\bottomrule
\end{tabular}
}
\vspace{5pt}
\caption{Computational requirements during training on synthetic datasets. }
\label{tab:compute:lowdim}
\end{table}

\subsection{Image Data}
\label{app:imp:img}

We adapt and modify the model architectures based on \citet{zhang2025towards} for MNIST and CIFAR-10 datasets and \citet{dao2023flowmatchinglatentspace} for the CelebA-HQ dataset. 

\noindent\textbf{MNIST.} For MNIST, we use the standard UNet. The ResNet blocks in the UNet function similarly to the model used for synthetic data. They process spatial and temporal inputs separately using convolutional and linear layers, respectively. The processed features are then concatenated and passed through a series of linear layers to capture space-time dependencies. The model consists of 1.07M parameters. 

\noindent\textbf{CIFAR-10.} For CIFAR-10, the model consists of two UNets: a large UNet for processing $v_\tau$ and $\tau$, and a smaller UNet (one-fourth the size) for processing $x_t$ and $t$. The outputs of each ResNet block in the smaller UNet are input to the corresponding ResNet blocks in the larger UNet, facilitating information exchange between different scales. The model consists of 44.81M parameters. 

\noindent\textbf{CelebA-HQ.} For CelebA-HQ, we first encode images into a latent space using the pretrained VAE encoder from Stable Diffusion \citep{rombach2022high}. We then use DiT \citep{peebles2023scalablediffusionmodelstransformers} as the backbone to process $v_\tau$ in this latent space. To condition the velocity prediction on $x_t$, we inject $x_t$ into each DiT block via cross-attention layers, while keeping the main DiT architecture unchanged. The time embedding is also modified by replacing $\text{embedding}(t)$ with $\text{embedding}(t) + \text{embedding}(\tau)$ to incorporate time information in both time axes. 

For training RF and OT-CFM on MNIST and CIFAR-10, we follow the procedures and hyperparameter settings from \citet{tong2023improving} and \citet{LipmanICLR2023}. For HRF2 on the same datasets, we adopt the training setup from \citet{zhang2025towards}. For all models on CelebA-HQ, we follow the procedures and hyperparameters from \citet{dao2023flowmatchinglatentspace}. 

For data coupling, we train the model from scratch following \cref{alg:data_coup}. Both the batch OT and training batch sizes are set to 128 for MNIST and CIFAR-10 and 256 for CelebA-HQ. 

For velocity coupling, we start from the HRF2-D model obtained in the previous step. Following the synthetic data experiments, we select the best-performing model on the validation dataset to ensure training quality. The training speed for velocity coupling is primarily limited by the velocity sample generation. %
Therefore, we generate velocity pairs before training and perform the training offline. 

We train the UNet for MNIST and CIFAR-10 on 1 NVIDIA L40S GPU and the DiT for CelebA-HQ on 8 NVIDIA L40S GPUs. Computational requirements, including training time and memory usage, are shown in \cref{tab:compute:mnist,tab:compute:cifar10,tab:compute:celeba}. 

\begin{table}[t]
\centering
\resizebox{1.0\columnwidth}{!}{
\begin{tabular}{lccccc}
\toprule
\textbf{MNIST} & \textbf{RF (1.08M)} & \textbf{OT-CFM (1.08M)} & \textbf{HRF2 (1.07M)} & \textbf{HRF2-D (1.07M)} & \textbf{HRF2-D\&V (1.07M)} \\
\midrule
Time (s/iter) & 0.045 & 0.046 & 0.046 & 0.046 & 0.046 \\
Memory (MB) & 2546 & 2546 & 2546 & 2546 & 2546 \\
Param.\ Counts & 1,075,361 & 1,075,361 & 1,065,698 & 1,065,698 & 1,065,698 \\
\bottomrule
\end{tabular}
}
\vspace{5pt}
\caption{Computational cost and model size for different methods on MNIST. }
\label{tab:compute:mnist}
\end{table}

\begin{table}[t]
\centering
\resizebox{1.0\columnwidth}{!}{
\begin{tabular}{lccccc}
\toprule
\textbf{CIFAR-10} & \textbf{RF (35.75M)} & \textbf{OT-CFM (35.75M)} & \textbf{HRF2 (44.81M)} & \textbf{HRF2-D (44.81M)} & \textbf{HRF2-D\&V (44.81M)} \\
\midrule
Time (s/iter) & 0.166 & 0.169 & 0.196 & 0.202 & 0.200 \\
Memory (MB) & 7480 & 7480 & 9220 & 9220 & 9220 \\
Param.\ Counts & 35,746,307 & 35,746,307 & 44,807,843 & 44,807,843 & 44,807,843 \\
\bottomrule
\end{tabular}
}
\vspace{5pt}
\caption{Computational cost and model size for different methods on CIFAR-10. }
\label{tab:compute:cifar10}
\end{table}

\begin{table}[h]
\centering
\resizebox{1.0\columnwidth}{!}{
\begin{tabular}{lccccc}
\toprule
\textbf{CelebA-HQ 256} & \textbf{RF (457.06M)} & \textbf{OT-CFM (457.06M)} & \textbf{HRF2 (616.20M)} & \textbf{HRF2-D (616.20M)} & \textbf{HRF2-D\&V (616.20M)} \\
\midrule
Time (s/iter) & 0.418 & 0.420 & 0.688 & 0.688 & 0.688 \\
Memory per GPU (GB) & 26.65 & 26.65 & 39.89 & 39.89 & 36.79 \\
Param.\ Counts & 457,062,416 & 457,062,416 & 616,197,186 & 616,197,186 & 616,197,186 \\
\bottomrule
\end{tabular}
}
\vspace{5pt}
\caption{Computational cost and model size for different methods on CelebA-HQ. }
\label{tab:compute:celeba}
\end{table}

For each model, %
we conduct five evaluation runs, and report the means and standard deviations.

%% file: neurips_2025.bbl
\begin{thebibliography}{34}
\providecommand{\natexlab}[1]{#1}
\providecommand{\url}[1]{\texttt{#1}}
\expandafter\ifx\csname urlstyle\endcsname\relax
  \providecommand{\doi}[1]{doi: #1}\else
  \providecommand{\doi}{doi: \begingroup \urlstyle{rm}\Url}\fi

\bibitem[Albergo and Vanden-Eijnden(2023)]{albergo2023building}
M.~Albergo and E.~Vanden-Eijnden.
\newblock Building normalizing flows with stochastic interpolants.
\newblock In \emph{Proc. ICLR}, 2023.

\bibitem[Albergo et~al.(2023)Albergo, Boffi, and Vanden-Eijnden]{albergo2023stochastic}
M.~Albergo, N.~Boffi, and E.~Vanden-Eijnden.
\newblock Stochastic interpolants: A unifying framework for flows and diffusions.
\newblock \emph{arXiv preprint arXiv:2303.08797}, 2023.

\bibitem[Chen and Lipman(2024)]{chen2024flow}
R.~T. Chen and Y.~Lipman.
\newblock Flow matching on general geometries.
\newblock In \emph{ICLR}, 2024.

\bibitem[Chen et~al.(2024)Chen, Goldstein, Hua, Albergo, Boffi, and Vanden-Eijnden]{chen2024probabilistic}
Y.~Chen, M.~Goldstein, M.~Hua, M.~S. Albergo, N.~M. Boffi, and E.~Vanden-Eijnden.
\newblock Probabilistic forecasting with stochastic interpolants and f$\backslash$" ollmer processes.
\newblock In \emph{ICML}, 2024.

\bibitem[Cheng and Schwing(2025)]{ChengICCV2025}
H.~K. Cheng and A.~Schwing.
\newblock {The Curse of Conditions: Analyzing and Improving Optimal Transport for Conditional Flow-Based Generation}.
\newblock In \emph{Proc. ICCV}, 2025.

\bibitem[Dao et~al.(2023)Dao, Phung, Nguyen, and Tran]{dao2023flowmatchinglatentspace}
Q.~Dao, H.~Phung, B.~Nguyen, and A.~Tran.
\newblock Flow matching in latent space, 2023.
\newblock URL \url{https://arxiv.org/abs/2307.08698}.

\bibitem[Deshpande et~al.(2018)Deshpande, Zhang, and Schwing]{DeshpandeCVPR2018}
I.~Deshpande, Z.~Zhang, and A.~G. Schwing.
\newblock {Generative Modeling using the Sliced Wasserstein Distance}.
\newblock In \emph{Proc. CVPR}, 2018.

\bibitem[Deshpande et~al.(2019)Deshpande, Hu, Sun, Pyrros, Siddiqui, Koyejo, Zhao, Forsyth, and Schwing]{IDeshpandeCVPR2019}
I.~Deshpande, Y.-T. Hu, R.~Sun, A.~Pyrros, N.~Siddiqui, S.~Koyejo, Z.~Zhao, D.~Forsyth, and A.~G. Schwing.
\newblock {Max-Sliced Wasserstein Distance and its use for GANs}.
\newblock In \emph{Proc. CVPR}, 2019.

\bibitem[Esser et~al.(2024)Esser, Kulal, Blattmann, Entezari, M{\"u}ller, Saini, Levi, Lorenz, Sauer, Boesel, et~al.]{esser2024scaling}
P.~Esser, S.~Kulal, A.~Blattmann, R.~Entezari, J.~M{\"u}ller, H.~Saini, Y.~Levi, D.~Lorenz, A.~Sauer, F.~Boesel, et~al.
\newblock Scaling rectified flow transformers for high-resolution image synthesis.
\newblock In \emph{Forty-first International Conference on Machine Learning}, 2024.

\bibitem[Fatras et~al.(2020)Fatras, Zine, Flamary, Gribonval, and Courty]{fatras2020learning}
K.~Fatras, Y.~Zine, R.~Flamary, R.~Gribonval, and N.~Courty.
\newblock Learning with minibatch wasserstein: asymptotic and gradient properties.
\newblock In \emph{AISTATS 2020-23nd International Conference on Artificial Intelligence and Statistics}, volume 108, pages 1--20, 2020.

\bibitem[Fatras et~al.(2021)Fatras, Zine, Majewski, Flamary, Gribonval, and Courty]{fatras2021minibatch}
K.~Fatras, Y.~Zine, S.~Majewski, R.~Flamary, R.~Gribonval, and N.~Courty.
\newblock Minibatch optimal transport distances; analysis and applications.
\newblock \emph{arXiv preprint arXiv:2101.01792}, 2021.

\bibitem[Flamary et~al.(2021)Flamary, Courty, Gramfort, Alaya, Boisbunon, Chambon, Chapel, Corenflos, Fatras, Fournier, et~al.]{flamary2021pot}
R.~Flamary, N.~Courty, A.~Gramfort, M.~Z. Alaya, A.~Boisbunon, S.~Chambon, L.~Chapel, A.~Corenflos, K.~Fatras, N.~Fournier, et~al.
\newblock {POT}: Python optimal transport.
\newblock \emph{Journal of Machine Learning Research}, 22\penalty0 (78):\penalty0 1--8, 2021.

\bibitem[Gat et~al.(2024)Gat, Remez, Shaul, Kreuk, Chen, Synnaeve, Adi, and Lipman]{gat2024discrete}
I.~Gat, T.~Remez, N.~Shaul, F.~Kreuk, R.~T. Chen, G.~Synnaeve, Y.~Adi, and Y.~Lipman.
\newblock Discrete flow matching.
\newblock In \emph{NeurIPS}, 2024.

\bibitem[Guo and Schwing(2025)]{schwing2025variational}
P.~Guo and A.~G. Schwing.
\newblock Variational rectified flow matching.
\newblock \emph{arXiv preprint arXiv:2502.09616}, 2025.

\bibitem[Jing et~al.(2024)Jing, Berger, and Jaakkola]{jing2024alphafold}
B.~Jing, B.~Berger, and T.~Jaakkola.
\newblock Alphafold meets flow matching for generating protein ensembles.
\newblock \emph{arXiv preprint arXiv:2402.04845}, 2024.

\bibitem[Karras et~al.(2018)Karras, Aila, Laine, and Lehtinen]{karras2018progressive}
T.~Karras, T.~Aila, S.~Laine, and J.~Lehtinen.
\newblock Progressive growing of {GAN}s for improved quality, stability, and variation.
\newblock In \emph{International Conference on Learning Representations}, 2018.
\newblock URL \url{https://openreview.net/forum?id=Hk99zCeAb}.

\bibitem[Krizhevsky(2009)]{krizhevsky2009learning}
A.~Krizhevsky.
\newblock Learning multiple layers of features from tiny images.
\newblock Technical report, University of Toronto, 2009.

\bibitem[LeCun et~al.(1998)LeCun, Bottou, Bengio, and Haffner]{lecun1998gradient}
Y.~LeCun, L.~Bottou, Y.~Bengio, and P.~Haffner.
\newblock Gradient-based learning applied to document recognition.
\newblock \emph{Proceedings of the IEEE}, 1998.

\bibitem[Lipman et~al.(2023)Lipman, Chen, Ben-Hamu, Nickel, and Le]{LipmanICLR2023}
Y.~Lipman, R.~Chen, H.~Ben-Hamu, M.~Nickel, and M.~Le.
\newblock {Flow Matching for Generative Modeling}.
\newblock In \emph{Proc. ICLR}, 2023.

\bibitem[Liu(2022)]{liu2022rectified}
Q.~Liu.
\newblock Rectified flow: A marginal preserving approach to optimal transport.
\newblock \emph{arXiv preprint arXiv:2209.14577}, 2022.

\bibitem[Liu et~al.(2023{\natexlab{a}})Liu, Gong, and Liu]{liu2023flow}
X.~Liu, C.~Gong, and Q.~Liu.
\newblock Flow straight and fast: Learning to generate and transfer data with rectified flow.
\newblock In \emph{Proc. ICLR}, 2023{\natexlab{a}}.

\bibitem[Liu et~al.(2023{\natexlab{b}})Liu, Zhang, Ma, and Peng]{liu2023instaflow}
X.~Liu, X.~Zhang, J.~Ma, and J.~Peng.
\newblock Instaflow: One step is enough for high-quality diffusion-based text-to-image generation.
\newblock In \emph{ICLR}, 2023{\natexlab{b}}.

\bibitem[Park et~al.(2024)Park, Lee, Kim, Lee, Hong, and Kim]{park2024constant}
D.~Park, S.~Lee, S.~Kim, T.~Lee, Y.~Hong, and H.~J. Kim.
\newblock Constant acceleration flow.
\newblock In \emph{NeurIPS}, 2024.

\bibitem[Peebles and Xie(2023)]{peebles2023scalablediffusionmodelstransformers}
W.~Peebles and S.~Xie.
\newblock Scalable diffusion models with transformers, 2023.
\newblock URL \url{https://arxiv.org/abs/2212.09748}.

\bibitem[Peyr{\'e} et~al.(2019)Peyr{\'e}, Cuturi, et~al.]{peyre2019computational}
G.~Peyr{\'e}, M.~Cuturi, et~al.
\newblock Computational optimal transport: With applications to data science.
\newblock \emph{Foundations and Trends{\textregistered} in Machine Learning}, 11\penalty0 (5-6):\penalty0 355--607, 2019.

\bibitem[Pooladian et~al.(2023)Pooladian, Ben-Hamu, Domingo-Enrich, Amos, Lipman, and Chen]{pooladian2023multisample}
A.-A. Pooladian, H.~Ben-Hamu, C.~Domingo-Enrich, B.~Amos, Y.~Lipman, and R.~Chen.
\newblock Multisample flow matching: Straightening flows with minibatch couplings.
\newblock In \emph{Proc. ICML}, 2023.

\bibitem[Rombach et~al.(2022)Rombach, Blattmann, Lorenz, Esser, and Ommer]{rombach2022high}
R.~Rombach, A.~Blattmann, D.~Lorenz, P.~Esser, and B.~Ommer.
\newblock High-resolution image synthesis with latent diffusion models.
\newblock In \emph{Proc. CVPR}, 2022.

\bibitem[Stark et~al.(2024)Stark, Jing, Wang, Corso, Berger, Barzilay, and Jaakkola]{stark2024dirichlet}
H.~Stark, B.~Jing, C.~Wang, G.~Corso, B.~Berger, R.~Barzilay, and T.~Jaakkola.
\newblock Dirichlet flow matching with applications to dna sequence design.
\newblock In \emph{ICML}, 2024.

\bibitem[Tong et~al.(2024)Tong, Malkin, Huguet, Zhang, Rector-Brooks, Fatras, Wolf, and Bengio]{tong2023improving}
A.~Tong, N.~Malkin, G.~Huguet, Y.~Zhang, J.~Rector-Brooks, K.~Fatras, G.~Wolf, and Y.~Bengio.
\newblock Improving and generalizing flow-based generative models with minibatch optimal transport.
\newblock \emph{TMLR}, 2024.

\bibitem[Villani(2009)]{villani2009optimal}
C.~Villani.
\newblock \emph{Optimal transport: old and new}, volume 338.
\newblock Springer, 2009.

\bibitem[Yim et~al.(2023)Yim, Campbell, Foong, Gastegger, Jim{\'e}nez-Luna, Lewis, Satorras, Veeling, Barzilay, Jaakkola, and Noe]{yim2023fast}
J.~Yim, A.~Campbell, A.~Y. Foong, M.~Gastegger, J.~Jim{\'e}nez-Luna, S.~Lewis, V.~G. Satorras, B.~S. Veeling, R.~Barzilay, T.~Jaakkola, and F.~Noe.
\newblock Fast protein backbone generation with {SE}(3) flow matching.
\newblock In \emph{NeurIPS Workshop: Machine Learning in Structural Biology}, 2023.

\bibitem[Zhang and Gienger(2024)]{zhang2024affordance}
F.~Zhang and M.~Gienger.
\newblock Affordance-based robot manipulation with flow matching.
\newblock \emph{arXiv preprint arXiv:2409.01083}, 2024.

\bibitem[Zhang et~al.(2024)Zhang, Pu, Kawamura, Loza, Bengio, Shung, and Tong]{zhang2024trajectory}
X.~Zhang, Y.~Pu, Y.~Kawamura, A.~Loza, Y.~Bengio, D.~L. Shung, and A.~Tong.
\newblock Trajectory flow matching with applications to clinical time series modeling.
\newblock In \emph{NeurIPS}, 2024.

\bibitem[Zhang et~al.(2025)Zhang, Yan, Schwing, and Zhao]{zhang2025towards}
Y.~Zhang, Y.~Yan, A.~Schwing, and Z.~Zhao.
\newblock Towards hierarchical rectified flow.
\newblock In \emph{The Thirteenth International Conference on Learning Representations}, 2025.

\end{thebibliography}
